%% file: iclr2027_arxiv_261120.tex
\documentclass{article}
\usepackage{iclr2027_conference,times}

\input{math_commands.tex}

\usepackage{amsmath}
\usepackage{amssymb}
\usepackage{amsthm}
\usepackage{array}
\usepackage{booktabs}
\usepackage{caption}
\newcolumntype{S}[1]{w{l}{#1}}
\newcolumntype{N}[1]{w{r}{#1}}
\usepackage{enumitem}
\usepackage{graphicx}
\usepackage{float}
\usepackage[colorlinks=true, citecolor=blue, urlcolor=blue]{hyperref}
\usepackage{microtype}
\usepackage{url}
\usepackage{wrapfig}
\usepackage{xcolor}
\usepackage{colortbl}
\definecolor{zebraTint}{HTML}{F4F4F4}

\usepackage{booktabs}
\usepackage{array}
\usepackage{tabularx}
\usepackage{threeparttable}
\usepackage{multirow}
\usepackage{multicol}
\usepackage{makecell}
\usepackage{longtable}
\usepackage{ragged2e}
\usepackage{adjustbox}
\usepackage{caption}
\usepackage{calc}
\usepackage{tikz}
\usepackage{algorithm}

\definecolor{gainteal}{HTML}{138A7E}
\definecolor{lossred}{HTML}{A33A3A}
\newcommand{\dbar}[2]{%
  \makebox[28pt][l]{%
    \makebox[6pt][r]{\ifnum#2<0 \textcolor{lossred!65}{\rule[-0.6pt]{\dimexpr#1pt*75/100\relax}{4.4pt}}\fi}%
    \makebox[22pt][l]{\textcolor{black!35}{\rule[-1.0pt]{0.5pt}{5.2pt}}%
      \ifnum#2>0 \textcolor{gainteal!65}{\rule[-0.6pt]{\dimexpr#1pt*75/100\relax}{4.4pt}}\fi}%
  }%
}
\newcommand{\dcell}[3]{\dbar{#1}{#2}\hspace{1pt}\makebox[15pt][r]{#3}}

\newtheoremstyle{opdplain}{6pt}{6pt}{\normalfont}{}{\bfseries}{.}{0.5em}{}
\theoremstyle{opdplain}
\newtheorem{lemma}{Lemma}
\newtheorem{proposition}{Proposition}

\definecolor{questionborder}{HTML}{A33A3A}
\definecolor{findingborder}{HTML}{2E6F50}
\definecolor{promptblue}{HTML}{4F94CD}
\definecolor{promptbg}{HTML}{EEF5FC}

\newcommand{\prompttemplate}[2]{%
    \par\medskip\noindent\begingroup
    \setlength{\parskip}{0pt}%
    \setlength{\fboxsep}{6pt}%
    \colorbox{promptblue}{%
        \parbox{\dimexpr\linewidth-2\fboxsep\relax}{%
            \color{white}\bfseries\strut #1}}%
    \par\nointerlineskip
    \setlength{\fboxsep}{10pt}%
    \fcolorbox{promptblue}{promptbg}{%
        \parbox{\dimexpr\linewidth-2\fboxsep-2\fboxrule\relax}{#2}}%
    \endgroup\par\medskip
}

\title{Towards Understanding On-Policy Distillation through the Lens of Test-Time Scaling}

\author{
\normalsize
\textbf{Xinmu Ge}$^{1,2,3,\dagger}$ \hspace{0.30em}
\textbf{Zizhuo Zhang}$^{4,\dagger}$ \hspace{0.30em}
\textbf{Yu Huang}$^{1,3}$ \hspace{0.30em}
\textbf{Jianing Zhu}$^{4}$ \hspace{0.30em}
\textbf{Lin Yuan}$^{3}$ \hspace{0.30em} 
\textbf{Wanli Gu}$^{3}$ \\
\textbf{Weichang Wu}$^{3}$ \hspace{0.5em}
\textbf{Weiran Huang}$^{1,2}$ \hspace{0.5em} 
\textbf{Bo Han}$^{4,*}$ \hspace{0.5em} 
\textbf{Xiaolu Zhang}$^{3,*}$  \hspace{0.5em} 
\textbf{Jiangchao Yao}$^{1,*}$ \\
\\[-3pt]
\normalfont\normalsize
$^{1}$Shanghai Jiao Tong University \hspace{0.50em}
$^{2}$Shanghai Innovation Institute \hspace{0.50em}
$^{3}$Ant Group \hspace{0.50em} \\
$^{4}$Hong Kong Baptist University \hspace{0.50em}
$^{\dagger}$Equal contribution \hspace{0.50em}
$^{*}$Corresponding authors \\[4pt]
\multicolumn{1}{c}{\normalfont\small
\hspace*{-5pt}%
\href{https://huggingface.co/collections/Geraldxm/opd-test-time-scaling-math-code-and-fact-checkpoints-6aba42275d3362d882cfc472}{\raisebox{-1.5pt}{\includegraphics[height=11pt]{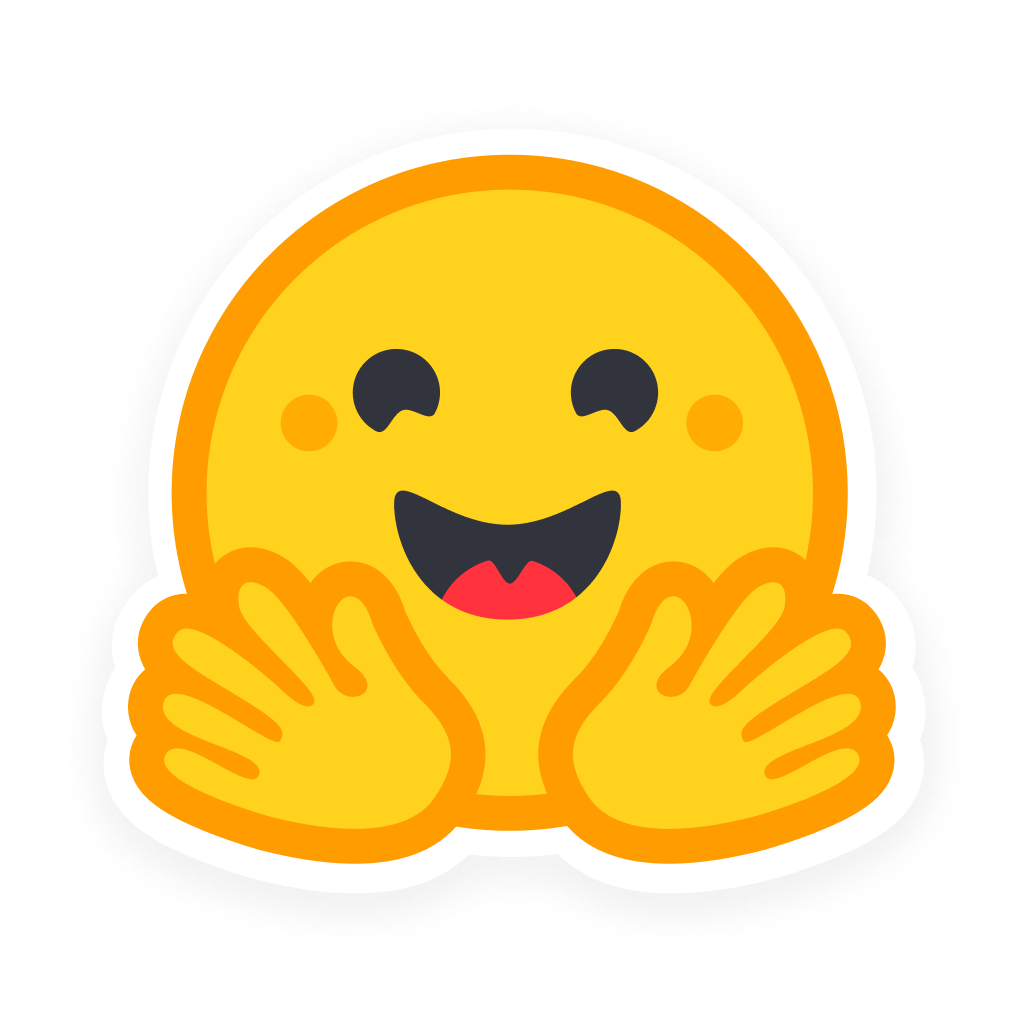}}\hspace{2pt}Hugging Face} \hspace{1.5em}
\href{https://github.com/Geraldxm/opd-test-time-scaling}{\raisebox{-1.5pt}{\includegraphics[height=9pt]{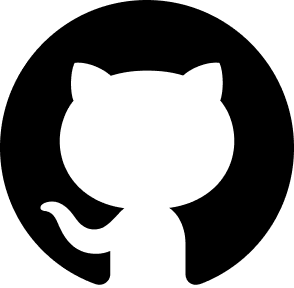}}\hspace{2pt}GitHub}\hspace*{5pt}}
}

\iclrfinalcopy

\newcommand{\rev}[1]{#1}
\newlength{\crossrowheight}
\newenvironment{revisionblock}{\begingroup}{\par\endgroup}
\begin{document}
\raggedbottom
\maketitle
\pagestyle{plain}
\vspace{-15pt}

\begin{abstract}
On-policy distillation (OPD) has emerged as a promising post-training technique for enhancing LLM reasoning. Under the reverse KL objective, the idealized optimum of OPD aligns the student distribution with that of the teacher. When the teacher consistently outperforms the student, this naturally suggests that OPD should yield broad improvements over the pre-OPD student. However, do such improvements extend across the entire range of test-time sampling budgets? In this work, we revisit this expectation through the lens of test-time scaling by varying the sampling budget $K$ and evaluating performance with pass@$K$. Across multiple settings, we observe two distinct patterns: OPD can improve pass@$K$ at both small and large sampling budgets, but it can also improve small-budget performance while reducing large-budget pass@$K$. We show one condition that guarantees such a reversal and an idealized reverse KL counterexample where it occurs even when the teacher has higher accuracy on every problem. To choose between two candidate teachers at a target sampling budget, we propose the \textit{Teacher Advantage Score at $K$} (TAS@$K$), which can be computed before OPD training to predict which teacher will lead to a larger improvement in pass@$K$. Across three domains and thirteen benchmarks, the ordering predicted by TAS@$K$ agrees with the observed pass@$K$ improvements of the resulting OPD models in 83.6\% of experiments, providing a useful signal for teacher selection at the target pass@$K$.

\end{abstract}

\vspace{-8pt}
\begin{figure*}[h!]
    \centering
    \includegraphics[width=0.95\textwidth]{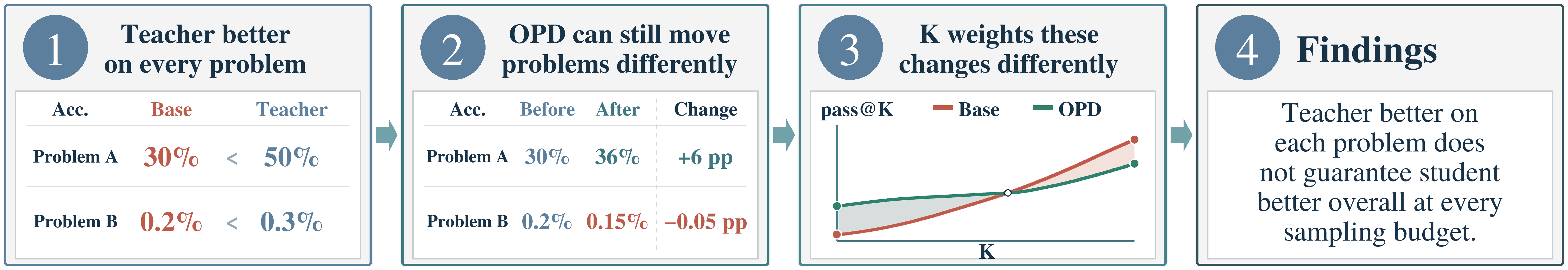}
    \vspace{-7.7pt}
    \captionsetup{width=0.95\textwidth}
    \caption{\textbf{Overview of the pass@$K$ reversal in OPD.} Even when the teacher is better on every problem,  OPD updates may improve the accuracy of some problems while slightly hurting others. Such uneven changes can increase pass@$K$ at small $K$ while decreasing it as $K$ grows. }
    \label{fig:opd-sampling-budget-teaser}
\end{figure*}

\renewcommand{\topfraction}{0.50}
\renewcommand{\dbltopfraction}{0.50}
\renewcommand{\textfraction}{0.45}
\renewcommand{\floatpagefraction}{0.70}
\setcounter{topnumber}{2}
\setcounter{dbltopnumber}{2}
\setcounter{bottomnumber}{0}
\setcounter{totalnumber}{2}

\section{Introduction}
\label{sec:introduction}

On-policy distillation (OPD) has emerged as a promising approach for improving the reasoning performance of large language models (LLMs)~\citep{agarwal2024onpolicy,song2026survey}, with its practical effectiveness demonstrated by recent industry-scale efforts, including DeepSeek-V4~\citep{xu2026deepseekv4}, Qwen3~\citep{yang2025qwen3}, and Nemotron-Cascade 2~\citep{yang2026nemotroncascade2}. In OPD, the student generates on-policy rollouts, while a stronger teacher provides token-level guidance through a reverse KL, encouraging the student distribution to move closer to the teacher~\citep{li2026rethinkingopd}. Current evaluations primarily focus on the performance gains of OPD under relatively small test-time sampling budgets (e.g., pass@1)~\citep{li2026rethinkingopd,jin2026entropyopd,yang2026exopd}. This raises a natural question: as a distillation paradigm that explicitly aligns the student with a stronger teacher, does OPD yield consistent gains as the test-time sampling budget scales from small to large $K$?

\par
Across multiple configurations in Table~\ref{Tble:OPD_Setting}, we observe two qualitatively different patterns. As shown in Table~\ref{tab:pass1-overview}, in some cases, OPD improves pass@$K$ across the full range of sampling budgets. In others, it improves pass@$K$ at small $K$ while degrading performance at large $K$ relative to the pre-OPD student. We further observe this latter pattern across multiple OPD variants, including EOPD~\citep{jin2026entropyopd}, ExOPD~\citep{yang2026exopd}, and Direct-OPD~\citep{feng2026directopd}, suggesting that the phenomenon is not specific to a particular OPD formulation. We give a sufficient condition under which pass@$1$ rises while pass@$K$ falls at a larger budget. Our theoretical analysis in Section~\ref{sec:opd-learning-path} shows that this can happen even when the teacher has higher accuracy on every problem.

\par
Motivated by these observations, we further ask whether one can predict, before full OPD training, which of two teachers will produce the student with higher pass@$K$ at a target sampling budget. To this end, we propose the \textit{Teacher Advantage Score at $K$} (TAS@$K$), which compares the relative likelihood preference of the two candidate teachers for correct over incorrect responses generated by the pre-OPD base model, and weights each problem for the target budget. The sign of TAS@$K$ provides a useful signal for selecting the teacher expected to yield the larger gain in target pass@$K$ after OPD training. Among all 177 comparisons, TAS@$K$ selects the teacher whose OPD-trained model achieves higher observed pass@$K$ in 148 cases, achieving an agreement rate of 83.6\%.

\par
Our contributions are summarized as follows:
\begin{itemize}[leftmargin=*,itemindent=0pt]
\item We systematically study OPD through the lens of test-time scaling, showing that its gains can exhibit qualitatively different patterns across sampling budgets (Section~\ref{sec:empirical-results}).
\item We theoretically provide a sufficient condition for a budget-dependent pass@$K$ reversal, showing it can occur even when the teacher has higher accuracy on every problem (Section~\ref{sec:opd-theory}).
\item We propose TAS@$K$, a teacher-selection score that can be computed before OPD training to predict which candidate teacher will yield a larger pass@$K$ improvement after OPD (Section~\ref{sec:teacher-selection}).
\end{itemize}

\begin{table}[t]
{
\centering
    \caption{\textbf{OPD settings.} Student-teacher pairs, benchmarks, and maximum sampling budgets.}
    \label{Tble:OPD_Setting}
    \vspace{-3mm}
    \resizebox{\textwidth}{!}{
    \setlength{\tabcolsep}{1.2pt}
    \begin{tabular}{l|ll@{\hspace{0.5pt}}c@{\hspace{0.5pt}}cc}
    \toprule[1.6pt]
    \textbf{Domain} & \textbf{Student} & \textbf{Teacher} & \textbf{Max $K$} & \textbf{Benchmarks} \\
    \midrule
    \multirow{4}{*}{\textbf{Math}} & Qwen3-1.7B-Base~\citep{yang2025qwen3} & Qwen3-4B-Base-GRPO~\citep{li2026rethinkingopd} & 1,024 & \multirow{4}{*}{\makecell{AMC23~\citep{mathai2023amc23},\\AIME24~\citep{aime24},\\AIME25~\citep{aime25},\\AIME26~\citep{aime26}}} \\
    ~ & DS-Distill-Qwen-1.5B~\citep{deepseekai2025deepseekr1} & Skywork-OR1-Math-7B~\citep{he2025skyworkor1} & 1,024 & ~ \\
    ~ & DS-Distill-Qwen-1.5B~\citep{deepseekai2025deepseekr1} & JustRL-DS-1.5B~\citep{he2025justrl} & 1,024 & ~ \\
    ~ & Qwen3-1.7B~\citep{yang2025qwen3} & Qwen3-4B-RL-Math~\citep{yang2026exopd} & 1,024 & ~ \\
    \midrule
    \multirow{2}{*}{\textbf{Code}} & Qwen3-1.7B~\citep{yang2025qwen3} & Qwen3-8B~\citep{yang2025qwen3} & \multirow{2}{*}{\makecell{512\\128}} & \multirow{2}{*}{\makecell{HumanEval+, MBPP+~\citep{liu2023humaneval},\\LiveCodeBench~\citep{jain2025livecodebench}}} \\
    ~ & Qwen3-1.7B~\citep{yang2025qwen3} & Qwen3-4B-RL-Code~\citep{yang2026exopd} & ~ & ~ \\
    \midrule
    \textbf{Fact} & \makecell[l]{Qwen3-1.7B~\citep{yang2025qwen3}\\Qwen3-1.7B~\citep{yang2025qwen3}} & \makecell[l]{Qwen3-8B-RL-RAR~\citep{chen2025train}\\SESA-8B-search~\citep{fu2026selfplaymeetsskillevolution}} & \makecell{1,024\\1,024\\1,024\\1,024} & \makecell{HotpotQA~\citep{yang2018hotpotqa},\\TriviaQA~\citep{joshi2017triviaqa},\\PopQA~\citep{mallen2023PopQA},\\2WikiMultiHopQA~\citep{ho20202wikiQA}} \\
    \bottomrule[1.6pt]
    \end{tabular}}
\par
}
\end{table}

\section{Preliminaries}
\label{sec:preliminaries}

\subsection{On-Policy Distillation}
Let $\pi_\theta$ and $\pi_T$ denote the student and teacher models, respectively.
Given a problem $x$, the student samples a reasoning trajectory $y$ with length
$L$.
At the $t$-th decoding step, the student reaches the prefix state $(x,y_{<t})$, where the teacher provides a dense target distribution over the next token.
The standard OPD objective minimizes the reverse KL divergence between the student and teacher distributions over these student-visited states~\citep{li2026rethinkingopd,gu2024minillm}:
\begin{equation}
    \mathcal{L}_{\mathrm{OPD}}(\theta)
    =
    \mathbb{E}_{x\sim\mathcal{D},\,
    y\sim\pi_\theta(\cdot\mid x)}
    \left[
    \sum_{t=1}^{L}
    D_{\mathrm{KL}}\!\left(
    \pi_\theta(\cdot\mid x,y_{<t})
    \,\middle\|\,
    \pi_T(\cdot\mid x,y_{<t})
    \right)
    \right].
\end{equation}
Here, $D_{\mathrm{KL}}(\cdot\|\cdot)$ denotes the KL divergence over the next-token vocabulary; the student distribution is its first argument and the teacher distribution is the second, yielding the reverse KL direction.
In practice, OPD can use full-vocabulary KL, top-$k$ token KL, or sampled-token estimators.
Our settings include both top-$k$ and sampled-token implementations.
The defining distinction from off-policy distillation is the source of training trajectories: OPD samples them from the student policy being updated, so teacher feedback is based on the student's current behavior~\citep{agarwal2024onpolicy}.

\subsection{Sampling-Budget Evaluation}
Test-time scaling allocates additional computation by sampling multiple responses for the same problem~\citep{brown2024large}.
Given a model $\pi_\theta$ and $K$ independently sampled responses $\{y_i\}_{i=1}^{K}$ for a problem $x$, pass@$K$ measures whether at least one response is correct~\citep{chen2021codex}:
\begin{equation}
    \mathrm{pass}@K
    =
    \mathbb{E}_{
    x\sim\mathcal D,\,
    \{y_i\}_{i=1}^{K}\sim\pi_\theta(\cdot\mid x)
    }
    \left[
    \mathbb I\!\left(
    \exists i\in\{1,\ldots,K\},
    \text{ s.t. }y_i\text{ is correct}
    \right)
    \right].
\end{equation}
Pass@$K$ reflects the model's ability to discover at least one successful reasoning trajectory with increased sampling opportunities.
Small $K$ measures how readily a model produces a correct response under a limited budget; large $K$ indicates the capability coverage when more responses are sampled.
More details of finite-sample estimator used in our experiments are provided in Appendix~\ref{app:evaluation-details}.

\section{Theoretical Analysis of \texorpdfstring{Pass@$K$}{Pass@K}}
\label{sec:opd-theory}
To understand how OPD gains can persist or reverse as the sampling budget grows, we first examine how problem-level accuracy changes contribute to pass@$K$. We then ask whether a teacher that is more accurate on every problem prevents the student's accuracy from decreasing on any problem during OPD training. Let $p_\theta(x)$ denote the probability of a correct response to the problem $x$ under the model $\pi_\theta$, which we call its problem-level accuracy. For $K$ independent responses sampled from the same model, we define the problem-level and dataset-level pass@$K$ values as follows:
\[
 \mathrm{pass}@K(x;\theta)=1-(1-p_\theta(x))^K,\qquad
 \mathrm{pass}@K(\theta)=\mathbb E_x[\mathrm{pass}@K(x;\theta)].
\]
\subsection{Problem-Level Accuracy Changes Across Sampling Budgets}
\label{sec:opd-pointwise}

Let $\theta_0$ and $\theta_T$ denote the student parameters before and after OPD. For a problem $x$, write $\Delta p(x)=p_{\theta_T}(x)-p_{\theta_0}(x)$ and $\Delta\mathrm{pass}@K(x)=\mathrm{pass}@K(x;\theta_T)-\mathrm{pass}@K(x;\theta_0)$.

\begin{lemma}[Problem-Level Pass@K Preserves the Direction of Accuracy Changes]
\label{lem:opd-pointwise}
For every finite positive integer $K$, $\Delta\mathrm{pass}@K(x)$ and $\Delta p(x)$ have the same sign. For fixed $K$, as $\Delta p(x)\to0$,
\begin{equation}
\label{eq:opd-pointwise-change}
 \Delta\mathrm{pass}@K(x)
 =K(1-p_{\theta_0}(x))^{K-1}\Delta p(x)+O(\Delta p(x)^2).
\end{equation}
\end{lemma}

The direction of the change is exact for arbitrary finite changes in $p$, whereas Eq.~(\ref{eq:opd-pointwise-change}) only characterizes its local magnitude. In this local approximation, $K(1-p_{\theta_0}(x))^{K-1}$ serves as a budget-dependent sensitivity factor that scales each problem's accuracy change. Thus, the same problem cannot improve at small $K$ and degrade at large $K$, while for fixed $K>1$, changes on lower-accuracy problems have a larger effect on pass@$K$. The exact finite-change identity and proof are provided in Appendix~\ref{app:opd-metric-proof}.

\subsection{Dataset-Level Gains and Losses}
\label{sec:opd-metrics}

Since the problem-level accuracy changes $\Delta p(x)$ are fixed across sampling budgets, while the factor $K(1-p_{\theta_0}(x))^{K-1}$ that scales their contributions varies with $K$, the aggregate effect can change across budgets. At the dataset level, $\Delta\mathrm{pass}@K=\mathbb E_x[\Delta\mathrm{pass}@K(x)]$. Therefore, changes on different problems can be reweighted as $K$ varies, allowing the overall gain to reverse sign. The following proposition gives a sufficient condition for such a reversal.

\begin{proposition}[A Sufficient Condition for Higher Pass@1 but Lower Pass@K]
\label{prop:opd-budget-divergence}
Let $G=\mathbb E_x[\max\{\Delta p(x),0\}]$ and $L=\mathbb E_x[\max\{-\Delta p(x),0\}]>0$. Suppose every improved problem has initial accuracy at least $p_{+}$, and every degraded problem has initial accuracy at most $p_{-}$, where $0<p_{-}<p_{+}<1$. For any fixed $K>1$, the following inequality is sufficient for a reversal:
\begin{equation}
\label{eq:opd-budget-condition}
 1<\frac{G}{L}<\left(\frac{1-p_{-}}{1-p_{+}}\right)^{K-1}
 \quad\Longrightarrow\quad
 \Delta\mathrm{pass}@1>0,\quad\Delta\mathrm{pass}@K<0.
\end{equation}
\end{proposition}

Here, $p_{+}$ is a lower bound on the initial accuracy of improved problems, while $p_{-}$ is an upper bound for degraded problems. At $K=1$, $G>L$ raises pass@$1$; at the target $K$, the losses outweigh the gains after budget-dependent weighting. This condition is sufficient, not necessary. The proof, intermediate bound, and a numerical example are provided in Appendix~\ref{app:opd-metric-proof}.

\subsection{A Stronger Teacher Does Not Guarantee Problem-Level Improvement}
\label{sec:opd-learning-path}

We model small-step training by a differentiable path $\{\theta_\tau\}_{\tau\in[0,T]}$. For a student-visited prefix $s=(x',y_{<t})\sim\rho_{\theta_\tau}$, consider full-vocabulary reverse KL as an idealized OPD update. Under a continuous-time approximation of the expected update, with sampled prefixes held fixed when computing gradients, let $b_x(\tau)=\frac{d}{d\tau}p_{\theta_\tau}(x)$ denote the rate at which accuracy changes on $x$:
\begin{equation}
\label{eq:opd-rate-identity}
 b_x(\tau)=\mathbb E_{\substack{s\sim\rho_{\theta_\tau}\\a\sim\pi_{\theta_\tau}(\cdot\mid s)}}
 \!\left[\log\frac{\pi_T(a\mid s)}{\pi_{\theta_\tau}(a\mid s)}
 \left\langle\nabla_\theta p_{\theta_\tau}(x),\nabla_\theta\log\pi_{\theta_\tau}(a\mid s)\right\rangle\right].
\end{equation}
Teacher preferences guide token updates at student-visited prefixes, but those updates can raise or lower problem-level accuracy. Update conventions and derivations are given in Appendices~\ref{app:opd-normalization}-\ref{app:opd-reverse-kl}.

\begin{proposition}[A Stronger Teacher Does Not Prevent a Local Budget Reversal]
\label{prop:opd-teacher-nonmonotone}
For any sampling budget $J$, let $\Delta_h\mathrm{pass}@J=\mathrm{pass}@J(\theta_h)-\mathrm{pass}@J(\theta_0)$. For any fixed $K>1$, suppose the initial OPD-induced accuracy rates on a finite empirical dataset satisfy
\begin{equation}
\label{eq:opd-local-reversal}
 \mathbb E_x[b_x(0)]>0,
 \qquad
 \mathbb E_x\!\left[K(1-p_{\theta_0}(x))^{K-1}b_x(0)\right]<0.
\end{equation}
Then every sufficiently small $h>0$ yields $\Delta_h\mathrm{pass}@1>0$ and $\Delta_h\mathrm{pass}@K<0$. At $K=1024$, these inequalities are realizable under full-vocabulary reverse KL distillation even when the teacher is more accurate on every problem; problem-level teacher and student accuracies alone therefore cannot tell whether this OPD update raises or lowers accuracy on a given problem.
\end{proposition}

{Appendix~\ref{app:opd-counterexample} gives the two-problem construction, and Appendix~\ref{app:opd-metric-proof} gives the finite-path identity. Exactly matching a teacher that is more accurate on every evaluation problem would improve all budgets, but does not imply monotonic improvement during finite training. The construction shows how finite OPD training can yield small-budget gains while reducing large-budget pass@$K$.}

\begin{table}[t]
{
\centering
\setlength{\tabcolsep}{0.85pt}
\renewcommand{\arraystretch}{1}

\caption{\rev{\textbf{Pass@$1$ and full-budget pass@$K$ (\%) with Dataset~1.} Table~\ref{tab:pass1-overview-dataset2} in Appendix~\ref{app:training-data-results} reports the corresponding Dataset~2 results for the same eight settings.}}
\label{tab:pass1-overview}
\vspace{-8.5pt}
\fontsize{6.93}{7.7}\selectfont
\begin{tabular}{@{}l|ll|cc|cc@{}}
\toprule[1.01pt]
\textbf{Domain} & \textbf{Student} & \textbf{Teacher} & \textbf{Pass@$1$ (Base $\to$ OPD)} & $\boldsymbol{\Delta}$ \textbf{(pp)} & \textbf{Pass@$K$ (Base $\to$ OPD)} & $\boldsymbol{\Delta}$ \textbf{(pp)} \\
\midrule[0.32pt]
\multirow{4}{*}{\textbf{Math}} & Qwen3-1.7B-Base & Qwen3-4B-Base-GRPO & 10.6 $\longrightarrow$ 18.4 & \dcell{7.8}{1}{\textcolor{gainteal}{\textbf{+7.8}}} & 73.3 $\longrightarrow$ 65.4 & \dcell{7.9}{-1}{\textcolor{lossred}{\textbf{\textminus 7.9}}} \\
~ & DS-Distill-Qwen-1.5B & Skywork-OR1-Math-7B & 36.5 $\longrightarrow$ 41.7 & \dcell{5.1}{1}{\textcolor{gainteal}{\textbf{+5.1}}} & 87.5 $\longrightarrow$ 83.5 & \dcell{4.0}{-1}{\textcolor{lossred}{\textbf{\textminus 4.0}}} \\
~ & DS-Distill-Qwen-1.5B & JustRL-DS-1.5B & 36.5 $\longrightarrow$ 51.8 & \dcell{15.3}{1}{\textcolor{gainteal}{\textbf{+15.3}}} & 87.5 $\longrightarrow$ 85.0 & \dcell{2.5}{-1}{\textcolor{lossred}{\textbf{\textminus 2.5}}} \\
~ & Qwen3-1.7B & Qwen3-4B-RL-Math & 19.7 $\longrightarrow$ 47.3 & \dcell{27.6}{1}{\textcolor{gainteal}{\textbf{+27.6}}} & 79.2 $\longrightarrow$ 83.5 & \dcell{4.3}{1}{\textcolor{gainteal}{\textbf{+4.3}}} \\
\midrule[0.32pt]
\multirow{2}{*}{\textbf{Code}} & Qwen3-1.7B & Qwen3-8B & 36.6 $\longrightarrow$ 38.8 & \dcell{2.3}{1}{\textcolor{gainteal}{\textbf{+2.3}}} & 52.3 $\longrightarrow$ 59.5 & \dcell{7.2}{1}{\textcolor{gainteal}{\textbf{+7.2}}} \\
~ & Qwen3-1.7B & Qwen3-4B-RL-Code & 36.6 $\longrightarrow$ 42.7 & \dcell{6.1}{1}{\textcolor{gainteal}{\textbf{+6.1}}} & 52.3 $\longrightarrow$ 70.1 & \dcell{17.8}{1}{\textcolor{gainteal}{\textbf{+17.8}}} \\
\midrule[0.32pt]
\multirow{2}{*}{\textbf{Fact}} & Qwen3-1.7B & Qwen3-8B-RL-RAR & 9.5 $\longrightarrow$ 10.5 & \dcell{1.0}{1}{\textcolor{gainteal}{\textbf{+1.0}}} & 29.1 $\longrightarrow$ 38.0 & \dcell{8.9}{1}{\textcolor{gainteal}{\textbf{+8.9}}} \\
~ & Qwen3-1.7B & SESA-8B-search & 9.5 $\longrightarrow$ 10.7 & \dcell{1.2}{1}{\textcolor{gainteal}{\textbf{+1.2}}} & 29.1 $\longrightarrow$ 37.6 & \dcell{8.5}{1}{\textcolor{gainteal}{\textbf{+8.5}}} \\
\bottomrule[1.01pt]
\end{tabular}
\par
}
\end{table}

\section{Empirical Results under Test-Time Scaling}
\label{sec:empirical-results}

\subsection{Experimental Settings}
\label{sec:experimental-settings}

\begin{revisionblock}
\noindent\textbf{OPD settings.} We train each of the eight student-teacher configurations in Table~\ref{Tble:OPD_Setting} on two datasets, denoted Dataset~1 and Dataset~2. For Math, these are DAPO-Math-17K~\citep{yu2025dapo} and Hendrycks MATH~\citep{hendrycks2021math}. For Code, they are CodeR1-12K~\citep{ganler2025coder1} and APPS~\citep{hendrycks2021apps}. For Fact, they are NQ~\citep{kwiatkowski2019naturalquestions}/HotpotQA~\citep{yang2018hotpotqa} and TriviaQA~\citep{joshi2017triviaqa}. Training details are in Appendix~\ref{app:main-training}.
\end{revisionblock}

\noindent\textbf{Evaluation.} For Math, we evaluate on AMC2023~\citep{mathai2023amc23}, AIME2024~\citep{aime24}, AIME2025~\citep{aime25}, and AIME2026~\citep{aime26}.
For Code, we evaluate on HumanEval+ and MBPP+~\citep{liu2023humaneval} and LiveCodeBench v5 and v6~\citep{jain2025livecodebench}.
For Fact, we evaluate on HotpotQA, TriviaQA~\citep{joshi2017triviaqa}, PopQA~\citep{mallen2023PopQA}, and 2WikiMultiHopQA~\citep{ho20202wikiQA}, using 200 randomly sampled problems from each benchmark.
We report pass@$K$ across sampling budgets. More evaluation details are provided in Appendix~\ref{app:evaluation-details}.

\subsection{Main Results}
\begin{figure*}[!t]
\centering
    \includegraphics[width=0.95\textwidth]{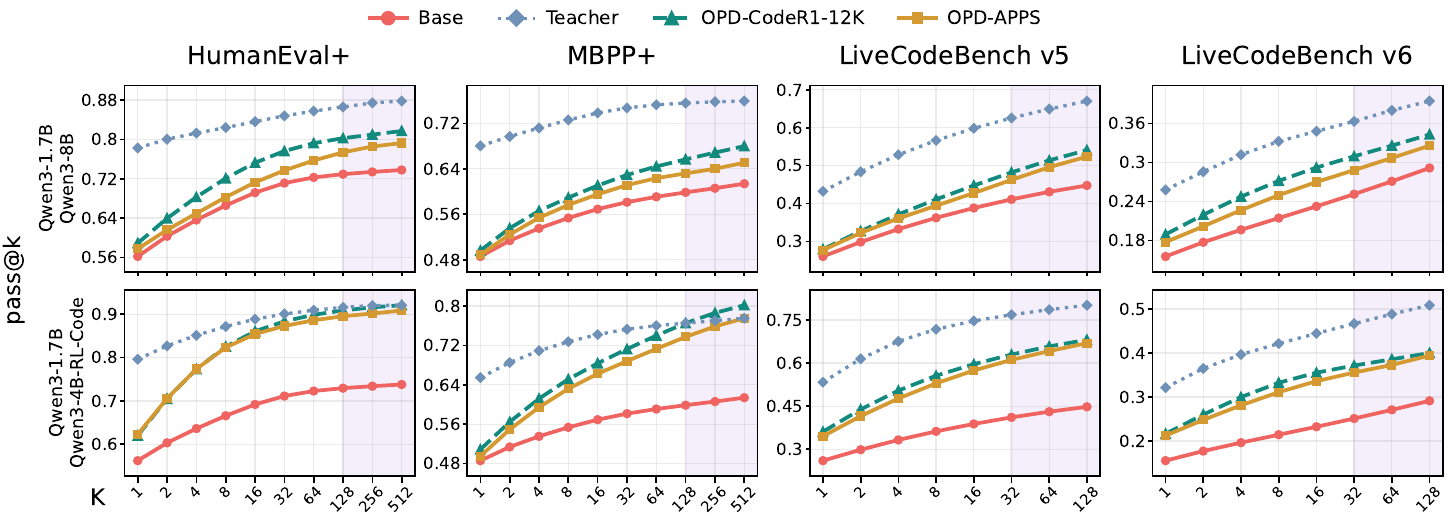}
    \vspace{-7.7pt}
    \captionsetup{width=0.95\textwidth}
    
    \caption{\rev{\textbf{Code pass@$K$ with two training datasets.} Both CodeR1-12K and APPS training yield gains over the pre-OPD base model at large budgets in both student-teacher configurations.}}
    \label{fig:code-settings}
    \par\vspace{3.5pt}
\centering
    \includegraphics[width=0.95\textwidth]{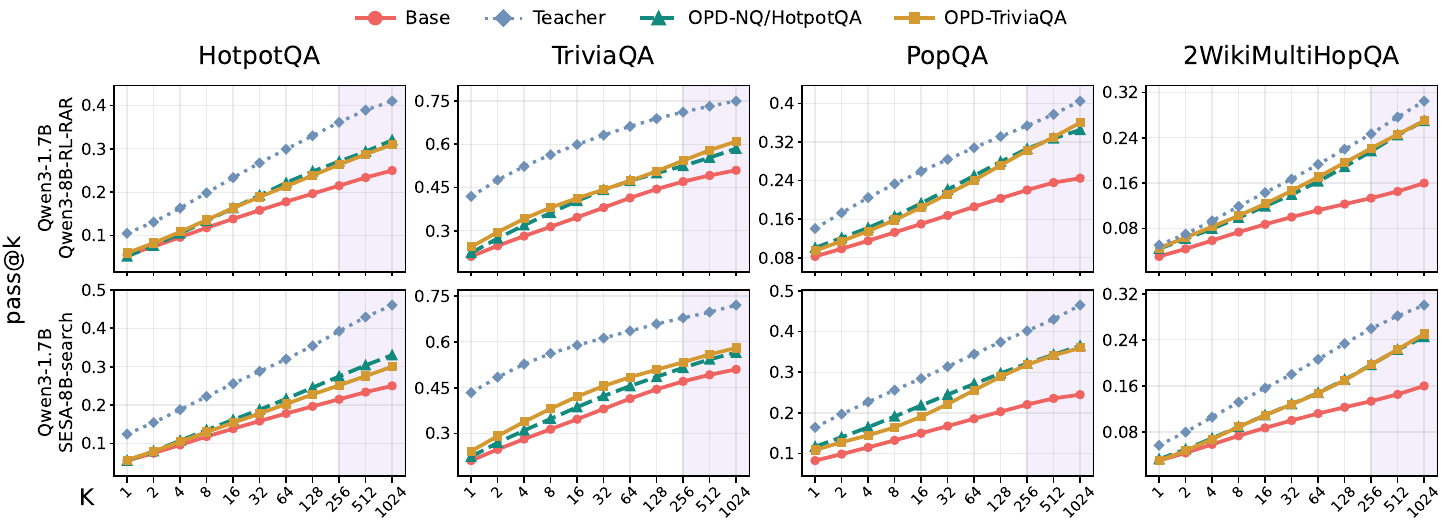}
    \vspace{-7.7pt}
    \captionsetup{width=0.95\textwidth}
    
    \caption{\rev{\textbf{Fact pass@$K$ with two training datasets.} Both NQ/HotpotQA and TriviaQA training yield gains over the pre-OPD base model on all four benchmarks at $K=1024$.}}
    \label{fig:fact-settings}
\end{figure*}

\noindent\textbf{Pass@$K$ gains persist or widen from small to large $K$.}
\rev{With Dataset~1, OPD improves pass@$1$ in all eight pairs} (Table~\ref{tab:pass1-overview}), and in several settings these gains persist or even widen as the sampling budget increases. In both Code settings, OPD stays ahead of the pre-OPD base model at every evaluated budget (Fig.~\ref{fig:code-settings}). For Qwen3-1.7B~$\leftarrow$~Qwen3-4B-RL-Code (student $\leftarrow$ teacher), the MBPP+ gain grows from 2.3 percentage points at $K=1$ to 18.8 percentage points at $K=512$. Similarly, with the same student and either Qwen3-8B-RL-RAR or SESA-8B-search as the teacher, both Fact settings remain ahead of the pre-OPD base model \rev{at $K=1024$ on all four benchmarks (Fig.~\ref{fig:fact-settings}).}

\begin{figure*}[t]
    \centering
    \includegraphics[width=0.95\textwidth]{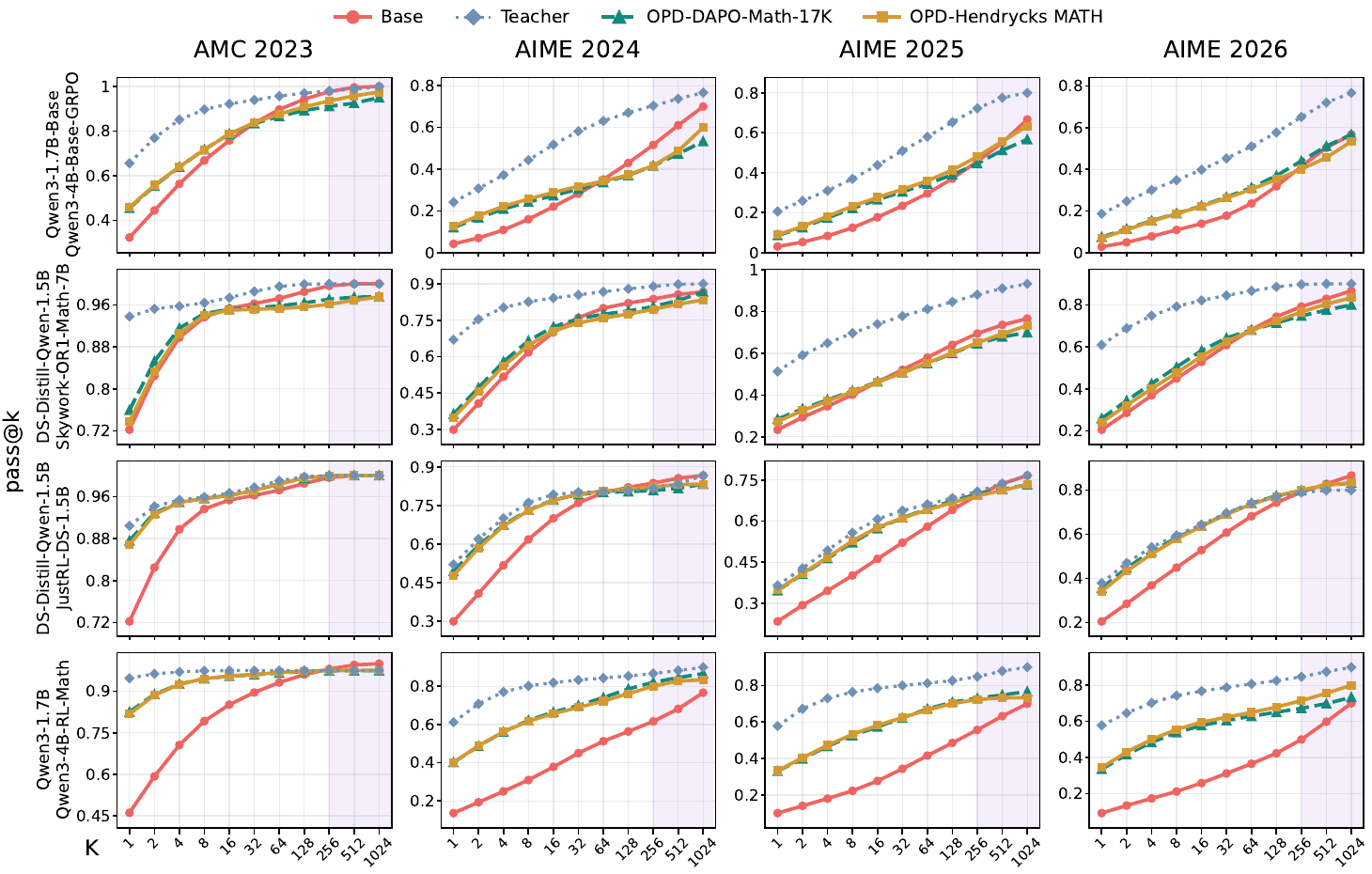}
    \vspace{-7.7pt}
    \captionsetup{width=0.95\textwidth}
    
    \caption{\rev{\textbf{Math pass@$K$ with two training datasets.} Both DAPO-Math-17K and Hendrycks MATH training yield large-budget reversals across multiple student-teacher configurations.}}
    \label{fig:math-settings}
\end{figure*}

\noindent\textbf{Pass@$K$ gains may decrease or even reverse as $K$ grows.}
In several Math settings, small-budget gains diminish as $K$ increases and can eventually reverse. In the Qwen3-1.7B-Base $\leftarrow$ Qwen3-4B-Base-GRPO setting and both DS-Distill-Qwen-1.5B settings using Skywork-OR1-Math-7B or JustRL-DS-1.5B as the teacher, several benchmark curves fall below the pre-OPD base model at large $K$ (Fig.~\ref{fig:math-settings}). For the Qwen3-1.7B-Base pair on AIME2024, OPD raises pass@$1$ by 7.8 percentage points but lowers pass@$1024$ by 16.7 points. Even for Qwen3-1.7B $\leftarrow$ Qwen3-4B-RL-Math, where the average gain remains positive at the largest sampling budget, gains on some benchmarks diminish or reverse. This reflects the budget-dependent aggregation analyzed in Section~\ref{sec:opd-theory}.

\begin{revisionblock}
\noindent\textbf{The scaling patterns persist across training datasets.} Math exhibits large-budget reversals in 10 of 16 teacher-benchmark combinations with DAPO-Math-17K and 12 of 16 with Hendrycks MATH (Table~\ref{tab:scaling-training-datasets}). Both training datasets produce reversals with all four Math teachers. Code and Fact retain large-budget gains in all eight combinations per domain under both training datasets. These results extend both scaling patterns across the two training datasets in each domain. Tables~\ref{tab:pass1-overview} and~\ref{tab:pass1-overview-dataset2} summarize the two datasets. Appendix~\ref{app:training-data-results} reports the Dataset~2 numerical results.
\end{revisionblock}
\begin{table}[t]
\centering
\renewcommand{\arraystretch}{1}
\setlength{\aboverulesep}{1pt}\setlength{\belowrulesep}{1pt}
\small\setlength{\tabcolsep}{5pt}

\caption{Reversals across training datasets: teacher-benchmark combinations with higher pass@$1$ and lower pass@$K_{\max}$ than the pre-OPD base model. $K_{\max}$ is the largest evaluated budget.}
\label{tab:scaling-training-datasets}
\vspace{-8.5pt}
\begin{tabular}{@{}llclc@{}}\toprule[1.01pt]
& \multicolumn{2}{c}{\textbf{Dataset 1}} & \multicolumn{2}{c}{\textbf{Dataset 2}} \\
\cmidrule(lr){2-3}\cmidrule(lr){4-5}
\textbf{Domain} & \textbf{Training data} & \textbf{Reversals} & \textbf{Training data} & \textbf{Reversals} \\ \midrule[0.32pt]
Math & DAPO-Math-17K & 10/16 & Hendrycks MATH & 12/16 \\
Code & CodeR1-12K & 0/8 & APPS & 0/8 \\
Fact & NQ/HotpotQA & 0/8 & TriviaQA & 0/8 \\
\bottomrule[1.01pt]\end{tabular}
\end{table}

\begin{figure*}[t]
    \centering
    \includegraphics[width=0.930\textwidth]
        {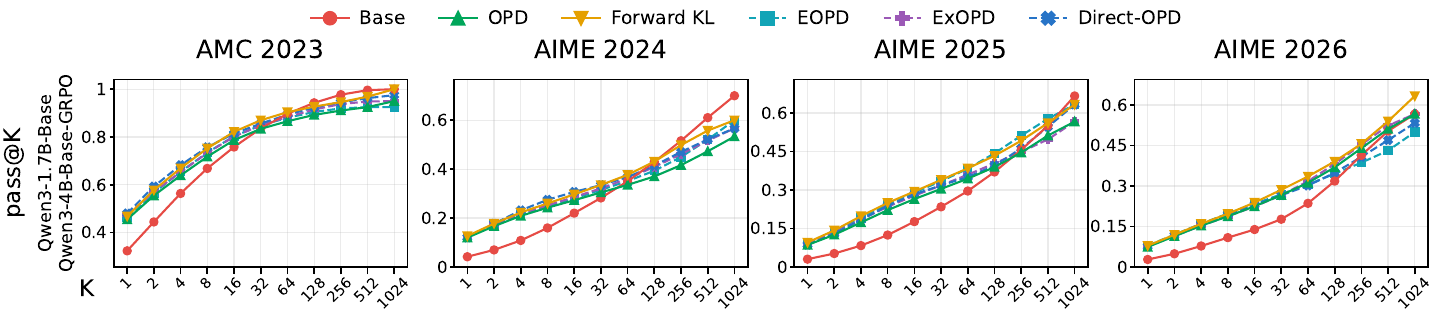}
    \vspace{-7.7pt}
    \captionsetup{width=0.930\textwidth}
    \caption{\textbf{Pass@$K$ for other OPD methods.} Several show large-budget reversals.}
    \label{fig:qwen3-advanced-opd}
\end{figure*}

\noindent\textbf{Large-budget reversals also occur with other OPD methods.}
In this Qwen3 Math setting, all methods use Qwen3-1.7B-Base as the student and Qwen3-4B-Base-GRPO as the teacher; Direct-OPD also uses the pre-RL Qwen3-4B-Base checkpoint.
We compare standard OPD with EOPD~\citep{jin2026entropyopd}, which applies entropy-gated forward KL to regularize the student, and a pure forward KL variant without entropy gating. We also evaluate ExOPD~\citep{yang2026exopd}, which extrapolates teacher reward signal, and Direct-OPD~\citep{feng2026directopd}, which uses the pre/post-RL teacher log-ratio as its training signal.
Their training configurations are reported in Table~\ref{tab:training-hyperparameters}.
All five improve pass@$1$ on each of the four benchmarks, yet several fall below the pre-OPD base model at $K=1024$ (Fig.~\ref{fig:qwen3-advanced-opd}), showing that large-budget degradation can also occur across alternative OPD formulations.

\subsection{Why Can \texorpdfstring{Pass@$K$}{Pass@K} Gains Persist or Reverse?}
\label{sec:passk-decline-analysis}

\begin{figure*}[t]
    \centering
    \includegraphics[width=0.95\textwidth]
        {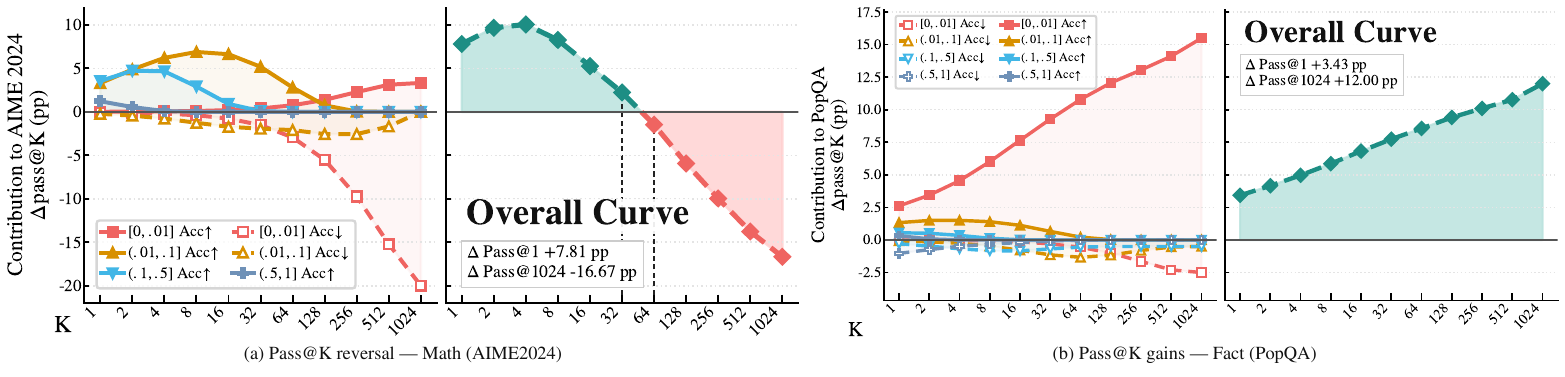}
    \vspace{-7.7pt}
    \captionsetup{width=0.95\textwidth}
    \caption{\textbf{Problem-level decomposition of pass@$K$ changes.} Left panels group contributions by estimated pre-OPD accuracy and the direction of the accuracy change; right panels sum them, showing (a) a reversal on AIME2024 and (b) widening gains on PopQA.}
    \label{fig:math-fact-passk-decomposition}
\end{figure*}

\noindent\textbf{Accuracy-grouped contributions to dataset-level $\Delta$pass@$K$ preserve their signs across $K$.} As shown in the left panels of Fig.~\ref{fig:math-fact-passk-decomposition}, we group problems by their pre-OPD accuracy and the direction of their accuracy change after OPD. Each group contributes either positively or negatively to the dataset-level $\Delta$pass@$K$, and the sign of its contribution remains unchanged as $K$ increases, although its magnitude varies with $K$. This observation is consistent with our analysis in Lemma~\ref{lem:opd-pointwise}.

\noindent\textbf{Aggregating these group contributions can lead to improvements or degradations in pass@$K$ at large $K$.} The right panels of Fig.~\ref{fig:math-fact-passk-decomposition} show their aggregate effect. On AIME2024 (Fig.~\ref{fig:math-fact-passk-decomposition}a), positive contributions dominate at small $K$, but negative contributions grow to dominate at large $K$, resulting in a reversal. On PopQA (Fig.~\ref{fig:math-fact-passk-decomposition}b), positive contributions remain dominant and the overall gain widens with $K$. These contrasting outcomes illustrate how the budget-dependent weighting in Section~\ref{sec:opd-metrics} shapes dataset-level pass@$K$ changes across different sampling budgets and benchmarks.

\begin{figure*}[!t]
    \centering
    \includegraphics[width=0.95\textwidth]{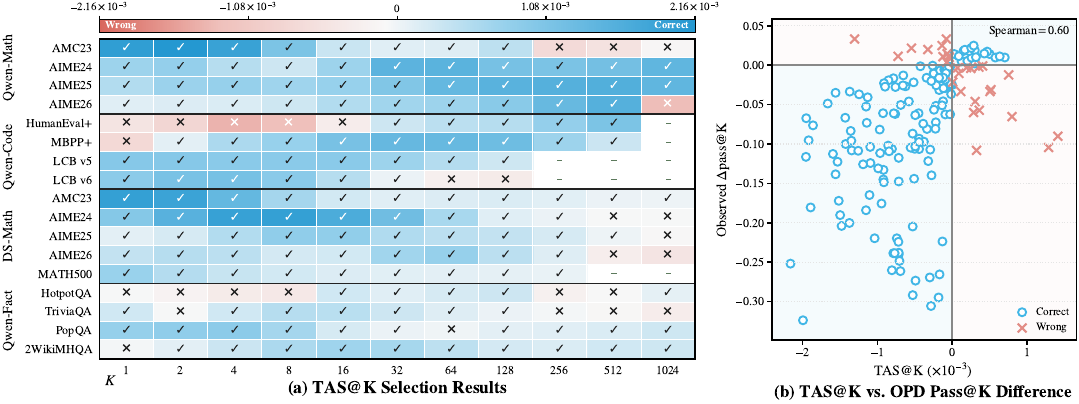}
    \vspace{-7.7pt}
    \captionsetup{width=0.95\textwidth}
    \caption{\textbf{Teacher selection with TAS@$K$.} Left: selection outcomes across settings and budgets. Blue/check and red/cross cells indicate correct and incorrect selections, respectively. Color intensity shows the absolute score in units of $10^{-3}$; dashes mark unavailable budgets. Right: signed TAS@$K$ versus the student pass@$K$ difference between teachers $A$ and $B$.}
    \label{fig:tas-teacher-selection}
\end{figure*}

\section{Budget-Aware Pairwise Teacher Selection}
\label{sec:teacher-selection}

\noindent\textbf{We aim to select the teacher that leads to higher student pass@$K$ at a target sampling budget.} To this end, we introduce the \textit{Teacher Advantage Score at $K$} (TAS@$K$), which ranks two candidate teachers for the same pre-OPD base model and training setup, at a chosen target budget $K$.

\subsection{From Teacher Preference to a Budget-Aware Score}
\noindent\textbf{TAS@$K$ computation.}
For each problem $x$ in the target dataset $\mathcal D$, we sample $N$ responses $\{y_i\}_{i=1}^{N}$ from the pre-OPD base model and estimate its accuracy $\widehat p_x=\frac{1}{N}\sum_{i=1}^{N}r_i$, where $r_i\in\{0,1\}$ indicates whether response $y_i$ is correct. Given two candidate teachers $A$ and $B$, we compute $d_i=\ell_A(y_i\mid x)-\ell_B(y_i\mid x)$, where $\ell_A$ and $\ell_B$ denote the mean token log likelihood of response $y_i$ under teachers $A$ and $B$, respectively. The TAS@$K$ score is computed as follows:
\begin{equation}
\mathrm{TAS}@K(A,B)
=\frac{1}{|\mathcal D|}\sum_{x\in\mathcal D}
K\widehat p_x(1-\widehat p_x)^K\widehat A_x(A,B).
\label{eq:tas}
\end{equation}
For problems with at least one correct and one incorrect sampled response, $\widehat A_x(A,B)$ is defined as:
\begin{equation}
\widehat A_x(A,B)
=\frac{\sum_{i:r_i=1}d_i}{\sum_{i=1}^{N}r_i}
-\frac{\sum_{i:r_i=0}d_i}{\sum_{i=1}^{N}(1-r_i)}.
\end{equation}
If all sampled responses are correct or all are incorrect, we set $\widehat A_x(A,B)=0$.
With the definition, a positive TAS@$K$ favors teacher $A$ and a negative score favors
teacher $B$ for the target budget $K$. Notably, computing TAS@$K$ does not
involve any OPD-trained model; it only uses responses sampled from
the pre-OPD base model and their likelihoods under the candidate
teachers. Therefore, TAS@$K$ can be computed before OPD training.
Detailed derivations are provided in Appendix~\ref{app:pairwise-teacher-proof}.

\begin{figure*}[!t]
\centering
    \includegraphics[width=0.95\textwidth]
        {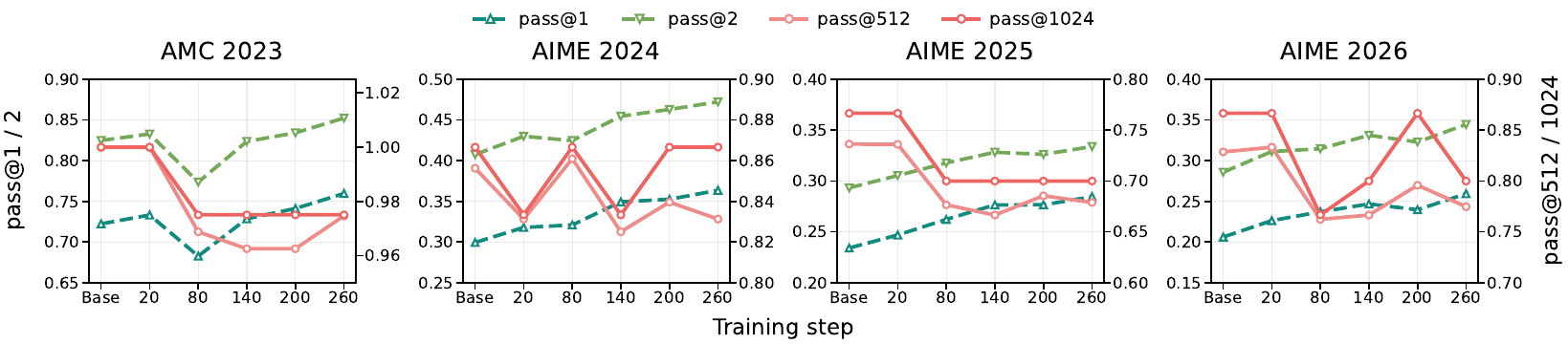}
    \vspace{-7.7pt}
    \captionsetup{width=0.95\textwidth}
    \caption{\textbf{Math training dynamics: small-budget gains, large-budget declines.} At~Step~260, pass@$1$ exceeds Base on four benchmarks, but pass@$1024$ trails it on three.}
    \label{fig:opd-training-dynamics}
    \par\vspace{1pt}
\centering
    \includegraphics[width=0.95\textwidth]
        {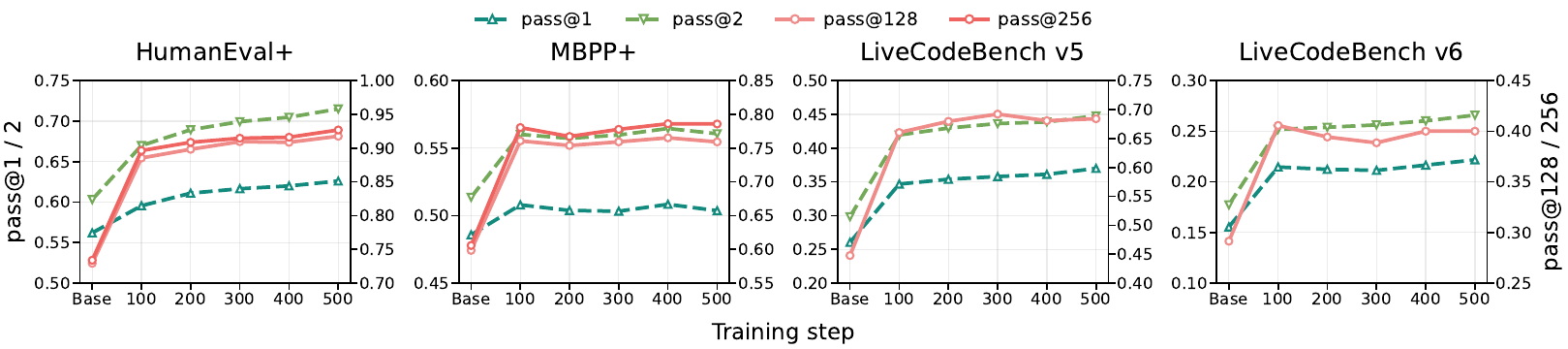}
    \vspace{-7.7pt}
    \captionsetup{width=0.95\textwidth}
    \caption{\textbf{Code training dynamics: gains appear early at small and large budgets.} From~Step~100 onward, the pass@$1$ and largest-budget curves stay above Base in every panel.}
    \label{fig:opd-training-dynamics-code}
    \vspace{-8pt}
\end{figure*}

\begin{revisionblock}
\subsection{Experimental Results of TAS@$K$}
\noindent\textbf{Settings.} We evaluate four teacher pairs ($A/B$):
(i) Qwen-Math uses Qwen3-8B / Qwen3-4B-RL-Math.
(ii) Qwen-Code uses Qwen3-8B / Qwen3-4B-RL-Code.
(iii) DeepSeek-Math uses Skywork-OR1-Math-7B / JustRL-DS-1.5B.
(iv) Qwen-Fact uses Qwen3-8B-RL-RAR / SESA-8B-search.
We use Dataset~1-trained students on the benchmarks in Section~\ref{sec:experimental-settings}, adding MATH500~\citep{lightman2023let} for DeepSeek-Math. Baselines select randomly, by teacher pass@$1$ or pass@$K$, or by UTAS@$K$, which removes budget weights from the same per-problem preferences. Agreement measures whether the chosen teacher yields higher student pass@$K$. Score ties use equal selection probabilities, yielding expected agreement and loss. Mean loss averages $L(K)=\max\{S_A(K),S_B(K)\}-S_{\mathrm{selected}}(K)$, where $S_A$ and $S_B$ are the two OPD-trained models' pass@$K$.

{\noindent\textbf{TAS@$K$ achieves 83.6\% agreement and the lowest mean selection loss.} Across 177 comparisons, TAS@$K$ achieves 83.6\% agreement and 0.498 pp mean selection loss, compared with 59.6\% and 1.632 pp for teacher pass@$K$ (Table~\ref{tab:selection-baselines}). Its mean selection loss is 69.5\% lower than teacher pass@$K$ and 28.9\% lower than UTAS@$K$. The value of student-specific teacher selection is particularly clear in DeepSeek-Math: the teacher-performance baselines favor Skywork, while the JustRL-distilled model performs better at most budgets. TAS@$K$ achieves 90.6\% agreement in this pair and the highest agreement in Qwen-Fact. Appendices~\ref{app:selection-uncertainty} and~\ref{app:budget-switches} report confidence intervals and budget-dependent teacher preferences across the evaluated benchmarks and sampling budgets.}
\end{revisionblock}
\begin{table}[t]
\centering
\renewcommand{\arraystretch}{1}
\setlength{\aboverulesep}{1pt}
\setlength{\belowrulesep}{1pt}

\footnotesize
\setlength{\tabcolsep}{3pt}
\captionsetup{justification=centering}
\caption{Selection agreement (\%) with mean loss (pp) in parentheses. Best values are bold and second-best underlined, separately for agreement and loss. Ties share markers.}
\label{tab:selection-baselines}
\vspace{-8.5pt}
\begin{tabular}{@{}lccccc@{}}\toprule[1.01pt]
\textbf{Teacher pair} & \textbf{Random} & \textbf{Teacher pass@$1$} & \textbf{Teacher pass@$K$} & \textbf{UTAS@$K$} & \textbf{TAS@$K$} \\ \midrule[0.32pt]
Qwen-Math & 50.0 (8.654) & \underline{90.9} (\underline{0.208}) & \textbf{95.5} (\textbf{0.079}) & \underline{90.9} (\underline{0.208}) & \underline{90.9} (\underline{0.208}) \\
Qwen-Code & 50.0 (4.379) & 72.2 (2.081) & \textbf{91.7} (\textbf{0.242}) & 72.2 (2.579) & \underline{77.8} (\underline{1.468}) \\
DeepSeek-Math & 50.0 (2.511) & 3.8 (4.932) & 4.7 (4.908) & \textbf{96.2} (\textbf{0.090}) & \underline{90.6} (\underline{0.303}) \\
Qwen-Fact & 50.0 (0.652) & 43.2 (0.715) & \underline{63.6} (\underline{0.377}) & 61.4 (0.389) & \textbf{72.7} (\textbf{0.230}) \\
\midrule[0.32pt] Overall & 50.0 (3.956) & 49.2 (2.130) & 59.6 (1.632) & \underline{81.4} (\underline{0.700}) & \textbf{83.6} (\textbf{0.498}) \\
\bottomrule[1.01pt]\end{tabular}
\end{table}

\noindent\textbf{TAS@$K$ exhibits a positive correlation with the observed pass@$K$ difference.}
{Signed TAS@$K$ correlates with the signed student pass@$K$ difference (Pearson $r=0.46$, Spearman $\rho=0.60$), linking pre-OPD teacher preferences to subsequent student pass@$K$ outcomes (Fig.~\ref{fig:tas-teacher-selection}, right).}

\begin{revisionblock}
\subsection{Teacher Selection Across Benchmarks}
\label{sec:cross-benchmark-selection}
For leave-one-benchmark-out evaluation, scores averaged over three benchmarks select a teacher for the fourth within each pair (164 comparisons). Separately, scores from AMC2023 and AIME2024/2025/2026 select between Skywork and JustRL. We evaluate these choices on BRUMO2025, CMIMC2025, and HMMT-Feb2025, on 100 disjoint test problems.

TAS@$K$ achieves 81.7\% leave-one-benchmark-out agreement and 81.8\% on the three test benchmarks (Table~\ref{tab:cross-benchmark-selection}). These results support teacher selection beyond the benchmarks used to compute the scores. Appendix~\ref{app:cross-benchmark-protocol} details both protocols and their paired problem-bootstrap confidence intervals.
\end{revisionblock}
\begin{table}[t]
\centering
\renewcommand{\arraystretch}{1}
\setlength{\aboverulesep}{1pt}
\setlength{\belowrulesep}{1pt}
\footnotesize

\caption{Teacher selection across benchmarks. TAS@$K$ entries are agreement (\%) [95\% CI]. The left and right protocols use 164 and 33 comparisons, respectively.}
\label{tab:cross-benchmark-selection}
\vspace{-8.5pt}
\setlength{\crossrowheight}{\baselineskip}
\begin{minipage}[t]{0.495\textwidth}\centering
\textbf{Leave-one-benchmark-out}\\[3pt]
\setlength{\tabcolsep}{2pt}\begin{tabular}{@{}lrc@{}}\toprule[1.01pt]
\textbf{Pair} & \textbf{Hits} & \textbf{TAS@$K$} \\ \midrule[0.32pt]
Qwen-Math & 40/44 & 90.9 [81.8, 97.7] \\
Qwen-Code & 30/32 & 93.8 [90.6, 93.8] \\
DeepSeek-Math & 39/44 & 88.6 [70.5, 90.9] \\
Qwen-Fact & 25/44 & 56.8 [36.4, 61.4] \\
\textbf{Overall} & \textbf{134/164} & \textbf{81.7 [73.2, 82.3]} \\\bottomrule[1.01pt]\end{tabular}\end{minipage}\hfill
\begin{minipage}[t]{0.495\textwidth}\centering
\textbf{Additional test benchmarks}\\[3pt]
\setlength{\tabcolsep}{2pt}\begin{tabular}{@{}lrc@{}}\toprule[1.01pt]
\textbf{Benchmark} & \textbf{Hits} & \textbf{TAS@$K$} \\ \midrule[0.32pt]
\rule[-0.375\crossrowheight]{0pt}{1.25\crossrowheight}BRUMO2025 & 9/11 & 81.8 [54.5, 81.8] \\
\rule[-0.375\crossrowheight]{0pt}{1.25\crossrowheight}CMIMC2025 & 9/11 & 81.8 [81.8, 90.9] \\
\rule[-0.375\crossrowheight]{0pt}{1.25\crossrowheight}HMMT-Feb2025 & 9/11 & 81.8 [45.5, 90.9] \\
\rule[-0.375\crossrowheight]{0pt}{1.25\crossrowheight}\textbf{Overall} & \textbf{27/33} & \textbf{81.8 [69.7, 84.8]} \\\bottomrule[1.01pt]\end{tabular}\end{minipage}
\end{table}

\section{Further Analysis}
\label{sec:further-analysis}

\noindent\textbf{Large-budget gains and reversals can emerge early during OPD training.} As shown in Fig.~\ref{fig:opd-training-dynamics} and Fig.~\ref{fig:opd-training-dynamics-code}, we compare intermediate checkpoints with the pre-OPD base model for DS-Distill-Qwen-1.5B~$\leftarrow$~Skywork-OR1-Math-7B (the Math pair) and Qwen3-1.7B~$\leftarrow$~Qwen3-4B-RL-Code (the Code pair). For the Math pair, pass@$1$ improves while pass@$K$ shows an overall downward trend as $K$ increases; by Step~80 on AIME2026, pass@$1$ is 3.1 percentage points above the pre-OPD base model while pass@$1024$ is 10.0 percentage points below it. For the Code pair, pass@$1$ improves together with an overall upward trend in pass@$K$; Step~100 already improves pass@$1$ and each benchmark's largest evaluated pass@$K$ on all four code benchmarks. These results show that the large-$K$ behavior observed after OPD training can already emerge at these early stages of intermediate training steps.

\begin{wrapfigure}{r}{0.46\textwidth}
\centering
\vspace*{-13.2pt}
    \captionsetup{type=figure,width=0.913\linewidth}
    \includegraphics[width=0.843\linewidth]
        {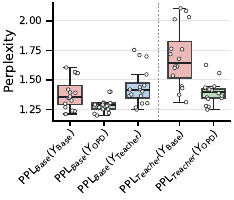}
    \vspace{-7.7pt}
    \caption{\textbf{Math trajectory perplexity.} $Y_{\mathrm{Base}}$, $Y_{\mathrm{OPD}}$, and $Y_{\mathrm{Teacher}}$ denote trajectory sources; $\mathrm{PPL}_{\mathrm{Base}}$ and $\mathrm{PPL}_{\mathrm{Teacher}}$ denote the respective scoring models.}
    \label{fig:skywork-retained-alignment}
    \vspace{-22pt}
\end{wrapfigure}

\noindent\textbf{OPD trajectories are favored by the teacher while remaining likely under the pre-OPD base model.} For the Math pair, Fig.~\ref{fig:skywork-retained-alignment} compares the perplexity of trajectories sampled from the pre-OPD base, OPD-trained, and teacher models, scored by both the pre-OPD base model and the teacher. OPD-generated trajectories receive lower perplexity than pre-OPD base trajectories under the teacher scorer, while remaining even more likely than both pre-OPD base and teacher trajectories under the pre-OPD base scorer. This suggests that OPD shifts generation toward trajectories better aligned with the teacher while remaining within high-likelihood regions of the pre-OPD base distribution. More details on the sampling and scoring setup are in Appendix~\ref{app:random-trajectory-ppl}.

\begin{figure}[t]
\centering
\setlength{\fboxsep}{4pt}
\begin{minipage}{\textwidth}
\centering
\fcolorbox{questionborder}{black!3}{\parbox{\dimexpr\linewidth-2\fboxsep-2\fboxrule\relax}{%
  \centering\footnotesize\textbf{AMC2023 12A, Problem 4:} How many digits are in the base-ten representation of
  \(8^5\cdot5^{10}\cdot15^5\)?}}

\vspace{2pt}
\begin{minipage}[t]{\dimexpr(\linewidth-5pt)/2\relax}
\fcolorbox{black!40}{black!3}{\parbox[t][21mm][t]{\dimexpr\linewidth-2\fboxsep-2\fboxrule\relax}{%
  \raggedright\fontsize{7.5}{10.5}\selectfont\setlength{\baselineskip}{10.8pt}\setlength{\lineskiplimit}{-\maxdimen}\setlength{\fboxsep}{2pt} \textbf{Base trajectory (221/1024).} \(8=2^3\), \(15=3\cdot5\), so \(N=8^5\,5^{10}\,15^5=2^{15}\,3^5\,5^{15}\). Then \(\log_{10}N=15\log_{10}2+5\log_{10}3+15\log_{10}5\approx17.3855\), so \(N\) has \(\lfloor17.3855\rfloor+1=18\) digits: \(\boxed{18}\).\par
  \vfill
  {\scriptsize\textcolor{findingborder}{\textbf{Reading note.} Completes the factorization and digit-count route.}}}}
\end{minipage}\hspace{4pt}%
\begin{minipage}[t]{\dimexpr(\linewidth-5pt)/2\relax}
\fcolorbox{black!40}{black!3}{\parbox[t][21mm][t]{\dimexpr\linewidth-2\fboxsep-2\fboxrule\relax}{%
  \raggedright\fontsize{7.5}{10.5}\selectfont\setlength{\baselineskip}{10.8pt}\setlength{\lineskiplimit}{-\maxdimen}\setlength{\fboxsep}{2pt} \textbf{OPD trajectory (893/1024).} \textbf{Step 1:} \(8^5\,5^{10}\,15^5=2^{15}\,3^5\,5^{15}\). \textbf{Step 2:} \(2^{15}\,3^5\,5^{15}=10^{15}\cdot243\). \textbf{Step 3:} \(10^{15}\) appends 15 zeros to \(243\), so \(15+3=18\) digits: \(\boxed{18}\).\par
  \vfill
  {\scriptsize\textcolor{findingborder}{\textbf{Reading note.} Follows the same route, then uses \(10^{15}\cdot243\).}}}}
\end{minipage}

\vspace{-7.7pt}
\captionsetup{width=\textwidth}
\caption{\textbf{Case study of a problem both models solve.} At $K=1024$, the pre-OPD base model achieves an accuracy of 221/1,024, compared with 893/1,024 for the OPD-trained model. Both models generate similar reasoning patterns, such as the same factorization route.}
\label{fig:qwen-trajectory-cases}
\end{minipage}
\vspace*{-8pt}
\end{figure}

\noindent\textbf{An illustrative case study.}
Fig.~\ref{fig:qwen-trajectory-cases} compares two correct responses to AMC2023 12A Problem~4, where the OPD-trained model is substantially more accurate. Both use the same factorization but finish differently: the pre-OPD base trajectory uses logarithms, whereas the OPD trajectory rewrites it with a power of ten. This shows that both models can share the same solution structure while taking different, equally correct trajectories.

\begin{revisionblock}
\noindent\textbf{Response diversity.} We also analyze lexical diversity using Self-BLEU. Diversity decreases in three Math cases with large-budget reversals and increases in eight Code and Fact cases with large-budget gains. Appendix~\ref{app:self-bleu} reports the Self-BLEU computation and the per-setting diversity results.
\end{revisionblock}

\section{Related Work}
\label{sec:related-work}

\noindent\textbf{On-policy distillation.} Classical and sequence-level distillation transfer teacher predictions or generated sequences~\citep{hinton2015distilling,kim-rush-2016-sequence}; GKD~\citep{agarwal2024onpolicy} and MiniLLM~\citep{gu2024minillm} instead use student-generated trajectories for teacher feedback. Recent studies identify prefix mismatch and estimator bias~\citep{fu2026revisitingopd,zhu2026facesonpolicydistillationpitfalls}, while others examine early parameter updates and training efficiency~\citep{cai2026foresee}. REOPOLD~\citep{ko2026reopold}, entropy-aware forward KL~\citep{jin2026entropyopd}, and trust-region OPD~\citep{xing2026tropd} seek more stable supervision or better preservation of diversity; Direct-OPD~\citep{feng2026directopd} transfers RL-induced policy shifts, whereas ExOPD~\citep{yang2026exopd} extrapolates them. Alongside these studies of training signals and objectives, we examine how the resulting student's pass@$K$ changes as the test-time sampling budget increases, including whether gains persist or reverse.

\noindent\textbf{Sampling efficiency and coverage.} Studies of RLVR report low-/high-budget crossovers and persistent gains~\citep{yue2025rlcapacity,wu2025invisibleleash,liu2025prorl}; MATH-Beyond examines expansion beyond the base model's finite-budget solved set~\citep{mayilvahanan2026mathbeyond}. For OPD, one study reports gains converging at large $K$ and examines the quality of teacher feedback~\citep{wang2026demystifyingopd}. Sampled-demonstration self-distillation can flatten pass@$K$ curves~\citep{nicolicioiu2026}. An OPD pass@$1$/pass@$16$ reversal on MBPP+ appears in the evaluation of influence-directed distillation~\citep{yang2026influencedirected}. We examine both reversals and persistent gains across Math, Code, and Fact. We explain how problem-level accuracy changes receive different weights as $K$ grows and, through an idealized analysis of OPD's full-vocabulary reverse KL update, show why some problems can worsen during finite training even when the teacher is more accurate on every problem.

\noindent\textbf{Teacher suitability and selection.} Teacher-student compatibility matters for OPD~\citep{li2026rethinkingopd}. RSR~\citep{yang2026informativealignment} assesses the suitability of teacher-generated trajectories for supervised fine-tuning and uses this signal for teacher selection, whereas TrustMOPD~\citep{sun2026trustmopd} allocates supervision among teachers at each student-generated prefix. We instead use a score computed before OPD training to select the teacher expected to yield higher student pass@$K$ at a target budget, given two candidates for the same pre-OPD base model under the same training setup.

\section{Conclusion}
\label{sec:conclusion}
In this work, we study on-policy distillation through the lens of test-time scaling. Across the evaluated settings, we observe two distinct patterns: pass@$K$ gains can persist or widen from small to large sampling budgets, or improve at small budgets but reverse at large budgets. Our theoretical analysis derives a sufficient condition for such large-budget degradation based on how problem-level accuracy changes contribute to pass@$K$ across sampling budgets. Additionally, we introduce TAS@$K$ to guide teacher selection at a target sampling budget, with its rankings matching the observed student pass@$K$ ordering in 83.6\% of comparisons. Overall, we hope these findings encourage future OPD studies to account for test-time sampling budgets in both evaluation and teacher selection in practice.

\renewcommand{\topfraction}{0.70}
\renewcommand{\dbltopfraction}{0.70}
\renewcommand{\textfraction}{0.20}
\renewcommand{\floatpagefraction}{0.50}
\setcounter{bottomnumber}{1}
\setcounter{totalnumber}{3}
\bibliography{iclr2027_conference}
\bibliographystyle{iclr2027_conference}

\appendix

\section{Proofs and Scope of the Analysis}
\label{app:opd-theory}

\subsection{Problem-Level Accuracy and Dataset-Level Changes}
\label{app:opd-metric-proof}

Following the main text, the problem-level metric is
\[
 \mathrm{pass}@K(x;\theta)=1-(1-p_\theta(x))^K,
\]
and the dataset-level metric is $\mathrm{pass}@K(\theta)=\mathbb E_x[\mathrm{pass}@K(x;\theta)]$. Conditional independence across the $K$ samples gives the following problem-level decomposition of the dataset-level changes
\[
 \Delta\mathrm{pass}@1=\mathbb E_x[\Delta p(x)],\qquad
 \Delta\mathrm{pass}@K=\mathbb E_x[\Delta\mathrm{pass}@K(x)].
\]
Throughout, $\Delta$ denotes the post-training value minus the pre-training value. The expression $1-(1-p)^K$ assumes that the $K$ responses are independently sampled from the same model distribution.

The exact finite change is
\[
 \Delta\mathrm{pass}@K(x)
 =(1-p_{\theta_0}(x))^K-(1-p_{\theta_T}(x))^K
 =\int_{p_{\theta_0}(x)}^{p_{\theta_T}(x)}K(1-r)^{K-1}\,dr.
\]

\begin{proof}[Proof of Lemma~\ref{lem:opd-pointwise}]
The function $1-(1-p)^K$ is strictly increasing on $[0,1]$; for $K>1$, its derivative vanishes at $p=1$, which does not affect strict monotonicity. Its integral form implies that $\Delta\mathrm{pass}@K(x)$ has the same sign as $\Delta p(x)$. For fixed $K$, Taylor expansion at $p_{\theta_0}(x)$ yields
\[
 \Delta\mathrm{pass}@K(x)
 =K(1-p_{\theta_0}(x))^{K-1}\Delta p(x)+O(\Delta p(x)^2).
\]
\end{proof}
For $K>1$, local sensitivity to an absolute accuracy change is greater at lower initial accuracy. For the same problem, the direction of the pass@$K$ change is consistent across sampling budgets.

\begin{proof}[Proof of Proposition~\ref{prop:opd-budget-divergence}]
Let $G=\mathbb E_x\max\{\Delta p(x),0\}$ and $L=\mathbb E_x\max\{-\Delta p(x),0\}>0$. The left inequality in \eqref{eq:opd-budget-condition} gives $\Delta\mathrm{pass}@1=G-L>0$. For gain problems, the integral kernel is at most $K(1-p_{+})^{K-1}$; for loss problems, it is at least $K(1-p_{-})^{K-1}$. Therefore,
\begin{equation}
\label{eq:opd-budget-bound}
 \Delta\mathrm{pass}@K
 \le K(1-p_{+})^{K-1}G-K(1-p_{-})^{K-1}L<0.
\end{equation}
The final strict inequality is exactly the ratio condition in the proposition. Its right-hand side is a conservative lower bound on relative sensitivity, so the condition is sufficient but not necessary.
\end{proof}

For example, under the proposition's probability-region separation assumption, let $K=10$, $p_{+}=.2$, $p_{-}=.02$, and $G=3L$. The sensitivity bounds for gains and losses are approximately $1.342$ and $8.337$, respectively; the loss-to-gain ratio is approximately $6.212>G/L=3$. Thus, $\Delta\mathrm{pass}@1=2L>0$, whereas the loss bound for the same example at the larger budget $K=10$ gives
\[
 \Delta\mathrm{pass}@10\le[30(.8)^9-10(.98)^9]L
 \simeq-4.311L<0.
\]

If $p_{\theta_\tau}(x)$ is absolutely continuous along a finite training path and $\int_0^T\mathbb E_x[|b_x(\tau)|]d\tau<\infty$, the chain rule and interchange of integration then give the following integral expression.
\begin{equation}
\label{eq:opd-metric-path}
 \Delta\mathrm{pass}@K=\int_0^T\mathbb E_x\!\left[
 K(1-p_{\theta_\tau}(x))^{K-1}b_x(\tau)\right]d\tau,
\end{equation}
\subsection{OPD Updates and Prefix Weighting}
\label{app:opd-normalization}

Let $x'\sim\mathcal D$ and $y\sim\pi_{\theta_{\rm old}}(\cdot\mid x')$ denote a student rollout drawn with fixed sampling parameters $\theta_{\rm old}$, and let $s=(x',y_{<t})$. In this subsection, $L$ denotes trajectory length, distinct from the preceding loss quantity, and we assume $0<\mathbb E[L]<\infty$. The effective token distribution induced by the trajectory-sum objective satisfies
\[
 \mathbb E_{s\sim\rho_{\theta_{\rm old}}}[g(s)]
 =\frac{\mathbb E_{x',y}[\sum_{t=1}^{L}g(x',y_{<t})]}{\mathbb E_{x',y}[L]}.
\]
The normalized local loss is therefore
\[
 \mathcal L(\theta;\theta_{\rm old})=
 \mathbb E_{s\sim\rho_{\theta_{\rm old}}}[D(\theta;s)].
\]
It equals the trajectory-sum loss under fixed rollouts divided by $\mathbb E[L]$; matching one update between the two objectives requires $\eta_{\rm normalized}=\mathbb E[L]\eta_{\rm sum}$. Different token or microbatch weights require a corresponding redefinition of $\rho$ and of the length normalization $\mathbb E[L]$ that it induces.

We differentiate $\mathcal L$ with respect to its first argument and stop gradients through the rollout. The continuously resampled limit is
\[
 \dot\theta_\tau=-\mathbb E_{s\sim\rho_{\theta_\tau}}
 [\nabla_\theta D(\theta_\tau;s)],\qquad
 b_x(s,\tau):=\langle\nabla_\theta p_{\theta_\tau}(x),
 -\nabla_\theta D(\theta_\tau;s)\rangle.
\]
Here, $b_x(s,\tau)$ is a single-prefix contribution, whereas $b_x(\tau)=\mathbb E_{s\sim\rho_{\theta_\tau}}[b_x(s,\tau)]$ is the problem-level average rate.
Under differentiability, integrable domination, and interchangeability of differentiation and expectation, $\frac{d}{d\tau}p_{\theta_\tau}(x)=b_x(\tau)$. This is neither the full gradient that differentiates through the on-policy sampling distribution nor the exact dynamics of Adam, finite batches, or top-$k$ distillation. The first-order response of an actual discrete update must use its realized parameter displacement; when top-$p$ or top-$k$ changes the support, global smoothness cannot be assumed across a switching point. The continuous progress variable $\tau$ can absorb proportional rescalings of loss normalization and step size; it is not directly an optimization step count or measured wall-clock time.

\subsection{Accuracy Dynamics under Full-Vocabulary Reverse KL}
\label{app:opd-reverse-kl}

We first consider the full-vocabulary idealization $D(\theta;s)=D_{\rm KL}(\pi_\theta(\cdot\mid s)\|\pi_T(\cdot\mid s))$ with full support for both distributions. This is a sufficient setting for the following calculation, not the top-$k$ token KL used in the main experiments:
\[
 \nabla_\theta D(\theta;s)=\sum_a\pi_\theta(a\mid s)
 \log\frac{\pi_\theta(a\mid s)}{\pi_T(a\mid s)}
 \nabla_\theta\log\pi_\theta(a\mid s),
\]
where the $+1$ term vanishes by the zero-mean score identity. Define the gradient alignment
\[
 h_{x,a}(s,\tau)=\langle\nabla_\theta p_{\theta_\tau}(x),
 \nabla_\theta\log\pi_{\theta_\tau}(a\mid s)\rangle.
\]
The contribution of the individual prefix state $s$ to the problem-level average rate is then
\[
 b_x(s,\tau)=\sum_a\pi_{\theta_\tau}(a\mid s)
 \log\frac{\pi_T(a\mid s)}{\pi_{\theta_\tau}(a\mid s)}h_{x,a}(s,\tau).
\]
Thus, reverse KL divergence alone does not determine the sign of $b_x$.

For the full-response distribution induced by a fixed evaluation decoding rule, let $R_x(y)\in\{0,1\}$ denote correctness and assume $p_{\theta_\tau}(x)\in(0,1)$ along the path. All conditional expectations below are taken under $y\sim\pi_{\theta_\tau}(\cdot\mid x)$. Define
\[
 a_x(\tau)=\left\langle
 \mathbb E[\nabla_\theta\log\pi_{\theta_\tau}(y\mid x)\mid R_x=1]-
 \mathbb E[\nabla_\theta\log\pi_{\theta_\tau}(y\mid x)\mid R_x=0],
 \dot\theta_\tau\right\rangle.
\]
The score identity gives $\frac{d}{d\tau}p_{\theta_\tau}(x)=p_{\theta_\tau}(x)(1-p_{\theta_\tau}(x))a_x(\tau)$. If $a_x$ is integrable, then
\[
 \frac{d}{d\tau}\operatorname{logit}p_{\theta_\tau}(x)=a_x(\tau),\qquad
 p_{\theta_T}(x)=\sigma\!\left(\operatorname{logit}p_{\theta_0}(x)+\int_0^T a_x(\tau)d\tau\right).
\]
Here, $\operatorname{logit}p=\log[p/(1-p)]$ and $\sigma(z)=1/(1+e^{-z})$. This identity applies to any differentiable evaluation-policy path satisfying these conditions; it is not specific to full-vocabulary KL. The evaluation decoding rule may differ from the training sampling rule, but both $p$ and the scores above must use the same evaluation distribution. Full-vocabulary KL only specifies the update direction above, not the magnitude or sign of the induced change in problem-level accuracy.

\subsection{A Two-Problem Counterexample}
\label{app:opd-counterexample}

\begin{proof}[Proof of Proposition~\ref{prop:opd-teacher-nonmonotone}]
For $K=1$, the initial dataset-level rate is $\mathbb E_x[b_x(0)]$. For any fixed $K>1$, the chain rule gives the following expression for the initial rate of the dataset-level metric
\[
 \left.\frac{d}{d\tau}\mathrm{pass}@K(\theta_\tau)\right|_{\tau=0}
 =\mathbb E_x\!\left[K(1-p_{\theta_0}(x))^{K-1}b_x(0)\right].
\]
The two strict inequalities in \eqref{eq:opd-local-reversal}, differentiability, and a first-order expansion therefore give the claimed signs for every sufficiently small positive finite update.

It remains to show that the criterion can hold despite a teacher that is more accurate on every problem. Consider two equally weighted, independent three-class softmax problems, with the first class correct:
\[
 P_1=(.3,.35,.35),\ Q_1=(.5,.25,.25),\qquad
 P_2=(.001,.899,.1),\ Q_2=(.002,.997,.001).
\]
The teacher assigns a higher probability to the correct class on both problems. Apply standard gradient descent to the average KL loss $(D_1+D_2)/2$, where $D_j=D_{\rm KL}(P_j\|Q_j)$. For problem $j\in\{1,2\}$ and class $i\in\{1,2,3\}$, let $r_{ji}=\log(Q_{ji}/P_{ji})$ and $\bar r_j=\sum_iP_{ji}r_{ji}$. The negative logit gradient of the single-problem KL is $v_{ji}=P_{ji}(r_{ji}-\bar r_j)$; hence, the actual direction for the average loss is $v_{ji}/2$, and
\[
 b_j=\frac{P_{j1}}{2}\left(v_{j1}-\sum_iP_{ji}v_{ji}\right).
\]
Substitution gives
\[
 b_1\simeq .02802438>0,\qquad
 b_2\simeq-.00016832<0.
\]
Therefore,
\[
 \left.\frac{d}{d\tau}\mathrm{pass}@1(\theta_\tau)\right|_{\tau=0}
 =(b_1+b_2)/2\simeq .01392803>0,
\]
\[
 \left.\frac{d}{d\tau}\mathrm{pass}@1024(\theta_\tau)\right|_{\tau=0}
 =512[(.7)^{1023}b_1+(.999)^{1023}b_2]
 \simeq-.03096647<0.
\]
For example, the expressions above verify $0.02802<b_1<0.02803$ and $-0.0001684<b_2<-0.0001683$; together with $0.7^{1023}<10^{-158}$ and $0.999^{1023}>1/3$, these bounds ensure that the two derivatives have strictly opposite signs. Smoothness further gives
\[
 \Delta_h\mathrm{pass}@K
 =h\left.\frac{d}{d\tau}\mathrm{pass}@K(\theta_\tau)\right|_{\tau=0}+O(h^2).
\]
Hence, for sufficiently small $h>0$, the changes have different directions at different sampling budgets. With step size $h=.2$ for the average loss (equivalently, $.1$ for the summed KL), updating the logits and reapplying softmax gives
\[
 p_1:.3\to.30563462,\quad p_2:.001\to.00096659,
\]
\[
 \mathrm{pass}@1:.1505\to.15330060,\qquad
 \mathrm{pass}@1024:.82051426\to.81426152,
\]
and $D_1:.08228288\to.07758400$ and $D_2:.36680638\to.33247070$; thus, both problem-level KL divergences decrease over this finite step, as expected under a reverse-KL update.

Now replace the teacher for the second problem with
\[
 Q'_2=(.002,.998\!\cdot\!.899/.999,.998\!\cdot\!.1/.999).
\]
This preserves the teacher's accuracy on both problems, while making the teacher-to-student probability ratios identical across the incorrect classes. Let $p=P_{21}=.001$ and $q=Q'_{21}=.002$. With $c=\operatorname{logit}q-\operatorname{logit}p>0$, direct simplification gives
\[
 b'_2=\frac{p^2c}{2}\left[(1-p)^2+P_{22}^2+P_{23}^2\right]
 \simeq6.3035711\times10^{-7}>0.
\]
Thus, the same student and identical problem-level teacher accuracies can produce opposite signs of the accuracy change on the second problem. Accuracy vectors alone are insufficient to determine this sign under a full-vocabulary reverse KL update.
\end{proof}
{This construction realizes a budget-dependent reversal under reverse KL divergence.}

\subsection{Scope of the Finite-Training Analysis}

If $p_{\theta_\tau}(x)$ is absolutely continuous, then $p_{\theta_T}(x)-p_{\theta_0}(x)=\int_0^T b_x(\tau)d\tau$. This describes only a finite path: local positive or negative rates cannot be extrapolated to a convergence claim. If, at the end of training, $\pi_{\theta_T}(\cdot\mid x)=\pi_T(\cdot\mid x)$ holds on every evaluation problem under the same decoding rule, and the teacher is no worse than the pre-OPD student on every problem, then $p_{\theta_T}(x)\ge p_{\theta_0}(x)$, so neither problem-level nor dataset-level pass@$K$ degrades for any $K$. Lowering KL only on training prefix states does not ensure this evaluation-distribution match; teacher realizability, state coverage, and optimization convergence are likewise not implied by the local analysis above.

\subsection{Derivation of TAS@$K$}
\label{app:pairwise-teacher-proof}

Fix a problem $x$ and suppress its subscript.
Let $S$ be the pre-OPD base-model response distribution, $T$ a teacher distribution, and $r(y)\in\{0,1\}$ the correctness indicator. Assume $p,q\in(0,1)$, $D_{\mathrm{KL}}(s^+\|t^+)<\infty$, $D_{\mathrm{KL}}(s^-\|t^-)<\infty$, and that the displayed log-ratio expectations are finite.
Write
\[
    S=p\,s^+ +(1-p)s^-,\qquad
    T=q\,t^+ +(1-q)t^-,
\]
where $s^+$ and $s^-$ are the base-model distributions conditional on correct and incorrect responses, respectively, and $t^+,t^-$ are the analogous teacher conditionals.
Consider the restricted student family
\[
    \pi_z=z\,s^+ +(1-z)s^- ,
\]
which changes only the problem-level accuracy while preserving the two within-group distributions.
This restriction is an analytic model, not an assumption that a real LLM update leaves within-group composition unchanged; it isolates the effect of problem-level accuracy alone in the analysis.

Let $d^+=D_{\mathrm{KL}}(s^+\|t^+)$ and $d^-=D_{\mathrm{KL}}(s^-\|t^-)$. Because the supports of $s^+$ and $s^-$ are disjoint,
\begin{equation}
    D_{\mathrm{KL}}(\pi_z\|T)
    =D_{\mathrm{KL}}(\mathrm{Bern}(z)\|\mathrm{Bern}(q))
     +z d^+ +(1-z)d^-.
    \label{eq:restricted-kl-decomposition}
\end{equation}
Let $z^*$ denote the student accuracy that minimizes $D_{\mathrm{KL}}(\pi_z\|T)$. Differentiating in $z$ yields
\[
    \operatorname{logit}z^*
    =\operatorname{logit}q-(d^+-d^-).
\]

Define the teacher's sequence-level likelihood advantage over the pre-OPD base model as
\[
    a_T(y)=\log T(y\mid x)-\log S(y\mid x),
\]
and its correct-versus-incorrect preference gap as
\[
    M_x(T)=\mathbb E_{s^+}[a_T(y)]-\mathbb E_{s^-}[a_T(y)].
\]
Direct substitution gives
\[
    M_x(T)=\operatorname{logit}q-\operatorname{logit}p-(d^+-d^-),
\]
and hence, at the restricted optimum, the student accuracy satisfies, on the logit scale,
\begin{equation}
    \operatorname{logit}z_x^*(T)=\operatorname{logit}p_x+M_x(T).
    \label{eq:restricted-optimum-m}
\end{equation}
For two teachers $A$ and $B$, define their sequence-level pairwise advantage as
\begin{equation}
 A_x(A,B)=
 \mathbb E_{y\sim s_x^+}\!\left[\log\frac{T_A(y\mid x)}{T_B(y\mid x)}\right]
 -\mathbb E_{y\sim s_x^-}\!\left[\log\frac{T_A(y\mid x)}{T_B(y\mid x)}\right].
 \label{eq:teacher-advantage-exact}
\end{equation}
Subtracting Eq.~(\ref{eq:restricted-optimum-m}) for the two teachers then gives
\begin{equation}
    \operatorname{logit}z^*_{A,x}-\operatorname{logit}z^*_{B,x}
    =M_x(T_A)-M_x(T_B)\equiv A_x(A,B).
    \label{eq:tas-exact-ordering}
\end{equation}
Here, logit is the logarithm of the ratio of correct to incorrect response probability. The identity states that the difference between the two optimal student accuracies on this scale equals the teachers' relative preference gap. Since logit is strictly increasing, the teacher whose sequence-level preferences more strongly favor correct responses yields the higher optimal accuracy on this problem. This conclusion assumes exact reverse KL minimization within the restricted student family.

To motivate the operational score, suppose both restricted optima remain near the base accuracy $p_x$ and their pairwise preference difference is small. Expanding the inverse logit around $p_x$ gives
\[
    z^*_{A,x}-z^*_{B,x}
    \approx p_x(1-p_x)A_x(A,B).
\]
Combining this with the pass@$K$ derivative $K(1-p_x)^{K-1}$ gives the first-order contribution
\[
    Kp_x(1-p_x)^K A_x(A,B),
\]
which motivates the budget weight used by TAS@$K$ in Eq.~(\ref{eq:tas}) at a target budget $K$.

Empirically, both teachers score the same responses $y_i\sim S$. With sequence-sum log likelihood and the same within-group weights, the $-\log S(y_i\mid x)$ term cancels teacher by teacher, yielding an estimator of $A_x(A,B)$. Our operational score instead uses the length-normalized mean token log likelihood of each response when scoring the two teachers, rather than the sequence-sum form
\[
    \bar{\ell}_{T,i}=\frac{1}{L_i}\sum_{t=1}^{L_i}\log T(y_{i,t}\mid x,y_{i,<t}),
    \qquad d_i=\bar{\ell}_{T_A,i}-\bar{\ell}_{T_B,i}.
\]
For correctness labels $r_i\in\{0,1\}$, the empirical problem-level advantage is
\begin{equation}
    \widehat A_x(A,B)
    =\begin{cases}
    \mathbb E[d_i\mid r_i=1]-\mathbb E[d_i\mid r_i=0],
      &0<\sum_i r_i<n_x,\\
    0, &\text{otherwise}.
    \end{cases}
    \label{eq:teacher-advantage}
\end{equation}
Thus, the operational $\widehat A_x$ is a length-normalized empirical surrogate rather than the exact sequence-level quantity in Eq.~(\ref{eq:teacher-advantage-exact}). Problems without both a correct and an incorrect pre-OPD response are retained in the full dataset denominator; zero here records unavailable preference information, not teacher equivalence, so these problems carry no directional evidence in this comparison.

\begin{revisionblock}
\paragraph{Observable and unobservable regimes.}
The preference gap is estimable only when the pre-OPD base model produces both correct and incorrect sampled responses for a problem. We call these \emph{valid problems}. If all sampled responses are correct or all are incorrect, we set that problem's preference gap to zero and retain the problem in the dataset average's denominator. Table~\ref{tab:valid-coverage} reports coverage. Zero-correct problems may become solvable after OPD and affect the final student ordering, even though they supply no direct preference signal. Cross-problem parameter sharing and changes in within-group response composition also fall outside the restricted derivation.

\paragraph{Selection on valid problems.}
{We recompute the two OPD-trained models' pass@$K$ on valid problems and evaluate the existing teacher choices against these rankings (Table~\ref{tab:valid-selection}). TAS@$K$ achieves 82.5\% overall agreement, compared with 77.4\% for UTAS@$K$, 58.5\% for teacher pass@$K$, and 48.0\% for teacher pass@$1$. TAS@$K$ leads all baselines on this valid subset.}
\end{revisionblock}
\begin{table}[!htbp]
\centering
\renewcommand{\arraystretch}{1}
\setlength{\aboverulesep}{1pt}
\setlength{\belowrulesep}{1pt}

\small
\setlength{\tabcolsep}{4pt}

\caption{Problems with both correct and incorrect sampled pre-OPD base responses.}
\label{tab:valid-coverage}
\vspace{-8.5pt}
\begin{tabular}{@{}lrr@{}}\toprule[1.01pt]
\textbf{Teacher pair} & \textbf{Valid / total problems} & \textbf{Valid (\%)} \\ \midrule[0.32pt]
Qwen-Math & 103/130 & 79.2 \\
Qwen-Code & 461/1597 & 28.9 \\
DeepSeek-Math & 447/630 & 71.0 \\
Qwen-Fact & 213/800 & 26.6 \\
\bottomrule[1.01pt]\end{tabular}
\end{table}
\begin{table}[!htbp]
\centering
\renewcommand{\arraystretch}{1}
\setlength{\aboverulesep}{1pt}
\setlength{\belowrulesep}{1pt}

\footnotesize
\setlength{\tabcolsep}{4pt}

\caption{Selection agreement (\%) with student rankings computed on valid problems. Parentheses give the change (pp) from agreement computed on all benchmark problems.}
\label{tab:valid-selection}
\vspace{-8.5pt}
\begin{tabular}{@{}lrrrr@{}}\toprule[1.01pt]
\textbf{Teacher pair} & \textbf{Teacher pass@$1$} & \textbf{Teacher pass@$K$} & \textbf{UTAS@$K$} & \textbf{TAS@$K$} \\ \midrule[0.32pt]
Qwen-Math & 86.4 (-4.5) & 90.9 (-4.5) & 86.4 (-4.5) & 86.4 (-4.5) \\
Qwen-Code & 69.4 (-2.8) & 88.9 (-2.8) & 63.9 (-8.3) & 86.1 (+8.3) \\
DeepSeek-Math & 3.8 (0.0) & 4.7 (0.0) & 94.3 (-1.9) & 90.6 (0.0) \\
Qwen-Fact & 45.5 (+2.3) & 65.9 (+2.3) & 59.1 (-2.3) & 65.9 (-6.8) \\
Overall & 48.0 (-1.1) & 58.5 (-1.1) & 77.4 (-4.0) & 82.5 (-1.1) \\
\bottomrule[1.01pt]\end{tabular}
\end{table}

\paragraph{Top-$k$ local proxy.}
We additionally evaluate a training-local proxy $G$ that replaces the sampled-token log likelihood at each response position with the corresponding Base-weighted teacher log likelihood over the Base model's top-16 next-token candidates, then applies the same response-length normalization.
Across the four matched pairs in Fig.~\ref{fig:tas-teacher-selection}, it obtains 39/44 Qwen-Math, 30/36 Qwen-Code, 48/53 DeepSeek-Math, and 29/44 Qwen-Fact point-direction agreements.
Because it depends on a local truncation of the vocabulary and is not the realized finite update, we treat $G$ as a sensitivity analysis and use the sampled-response token-average TAS@$K$ in the main text.

\section{Evaluation Details}
\label{app:evaluation-details}

\paragraph{Finite-sample pass@$K$ estimator.}
For each problem, we draw $n$ responses and let $c$ be the number judged correct.
For any $K\le n$, we use the standard unbiased estimator~\citep{chen2021codex}
\begin{equation}
    \widehat{\mathrm{pass}@K}
    =1-\frac{\binom{n-c}{K}}{\binom{n}{K}},
\end{equation}
with the ratio set to zero when $n-c<K$.
Dataset-level pass@$K$ averages this quantity over the problems in the evaluation set, yielding a single scalar per benchmark and sampling depth.

\begin{revisionblock}
\paragraph{Response pools and budget reuse.}
\label{app:tas-sampling}
For the original teacher-selection experiments, $N=\max K$ for each benchmark (Table~\ref{tab:tas-response-counts}). TAS@$K$ and pre-OPD pass@$K$ reuse the same $N$ sampled base responses. The two teachers score this same response pool. OPD-trained models are evaluated using their own responses. The per-problem correctness estimate and teacher preference gap are shared across budgets, while TAS@$K$ recomputes the budget weight for each target $K$. Thus the score is computed separately for each $K$, but comparisons on the same benchmark are correlated.

\paragraph{Sensitivity to the number of base responses.}
We recompute TAS@$K$ using $N/2$, $N/4$, or $N/8$ base responses sampled without replacement, and full-$N$ pools sampled with replacement, with 1,000 repetitions per design. Both teachers use the same sampled responses. Target budgets and student evaluation results remain fixed. Smaller pools estimate the correctness probability and preference gap in the TAS formula even when the target $K$ exceeds the reduced pool size. This is a plug-in score, not an unbiased pass@$K$ estimate from fewer than $K$ responses. Table~\ref{tab:tas-response-sensitivity} reports mean agreement and the central 95\% resampling interval. Agreement decreases as fewer responses are used. Since the full pools are available, we use all $N$ responses for the reported teacher choices.
\end{revisionblock}
\begin{table}[!htbp]
\centering
\renewcommand{\arraystretch}{1}
\setlength{\aboverulesep}{1pt}
\setlength{\belowrulesep}{1pt}

\footnotesize
\setlength{\tabcolsep}{4pt}

\caption{Response counts for the original teacher-selection experiments.}
\label{tab:tas-response-counts}
\vspace{-8.5pt}
\begin{tabular}{@{}llrr@{}}\toprule[1.01pt]
\textbf{Domain} & \textbf{Benchmark} & \textbf{$N$} & \textbf{$\max K$} \\ \midrule[0.32pt]
Math & AMC2023, AIME2024/2025/2026 & 1024 & 1024 \\
Math & MATH500 & 256 & 256 \\
Code & HumanEval+, MBPP+ & 512 & 512 \\
Code & LiveCodeBench v5/v6 & 128 & 128 \\
Fact & HotpotQA, TriviaQA, PopQA, 2WikiMultiHopQA & 1024 & 1024 \\
\bottomrule[1.01pt]\end{tabular}
\end{table}
\begin{table}[!htbp]
\centering
\renewcommand{\arraystretch}{1}
\setlength{\aboverulesep}{1pt}
\setlength{\belowrulesep}{1pt}

\small
\setlength{\tabcolsep}{4pt}

\caption{Sensitivity of TAS@$K$ selection to base response sampling over all 177 comparisons.}
\label{tab:tas-response-sensitivity}
\vspace{-8.5pt}
\begin{tabular}{@{}lrr@{}}\toprule[1.01pt]
\textbf{Sampling design} & \textbf{Agreement (\%) [95\% interval]} & \textbf{Change} \\ \midrule[0.32pt]
Original full $N$ & 83.6 & 0.0 pp \\
$N/2$ & 80.8 [75.1, 85.9] & -2.8 pp \\
$N/4$ & 78.8 [71.2, 85.9] & -4.8 pp \\
$N/8$ & 76.9 [68.1, 84.7] & -6.7 pp \\
Full-$N$ bootstrap & 81.1 [75.7, 85.9] & -2.5 pp \\
\bottomrule[1.01pt]\end{tabular}
\end{table}

\paragraph{Math.}
We use the curated benchmark snapshots from \texttt{math-vault}~\citep{ge_math_vault_2026} and the
\texttt{math-eval} pipeline~\citep{ge_math_eval_2026} for inference, replay, answer extraction, and scoring.
Official metrics compare the last complete boxed answer
\(\boxed{\cdot}\)
with the canonical answer.
Unless otherwise noted, evaluation uses temperature \(0.7\), top-\(p\) \(0.95\), seed \(0\), a 32,768-token context window, a 1,024-token prompt limit, and a 31,744-token output limit.
Each main Math curve uses 1,024 responses per problem on AMC2023, AIME2024, AIME2025, and AIME2026, as in the main text.

\paragraph{Math prompt format.}
For completion checkpoints, we pass the following string as a completion; for chat checkpoints, we place the identical string in the user message after the problem:

\begin{center}
\fbox{\begin{minipage}{0.88\linewidth}
\centering\ttfamily
\{\{problem\}\} Please reason step by step, and put your final answer within
\textbackslash boxed\{\}.
\end{minipage}}
\end{center}

\paragraph{Code.}
HumanEval+ and MBPP+ are evaluated with the EvalPlus~\citep{liu2023humaneval} task snapshots and unit-test harness\footnote{\url{https://github.com/evalplus/evalplus}}, requiring a completion to pass the benchmark's base and additional tests.
Both benchmarks use 512 responses per problem.
LiveCodeBench v5 (880 problems) and v6 (175 new problems) use the official code-generation evaluator~\citep{jain2025livecodebench} with 128 responses per problem.
All evaluations use the non-thinking Qwen3 chat serialization and seed 0. HumanEval+ and MBPP+ use temperature 0.7 and top-$p$ 0.95; LiveCodeBench uses temperature 1.0 and top-$p$ 0.8.
Pass@$K$ is computed from the per-problem correct-completion counts using the estimator above.

\paragraph{Code prompt formats.}
HumanEval+ and MBPP+ share the following user message, where \texttt{\{\{problem\}\}} is the benchmark-provided programming task. We supply no separate system message for these two benchmarks; the user message alone specifies the task and provides the problem.
\prompttemplate{HumanEval+ / MBPP+: user message}{\small\ttfamily\raggedright
Solve the following programming problem. Return only the complete Python solution, without Markdown fences. \{\{problem\}\}
}

LiveCodeBench v5 and v6 use the following system message. The user message depends on whether the task provides starter code; \texttt{\{\{question\}\}} denotes the problem statement.
\prompttemplate{LiveCodeBench: system message}{\small\ttfamily\raggedright
You are an expert Python programmer. You will be given a question (problem specification) and will generate a correct Python program that matches the specification and passes all tests.
}
\prompttemplate{LiveCodeBench: user message without starter code}{\small\ttfamily\raggedright
\#\#\# Question:\\
\{\{question\}\}\par\medskip
\#\#\# Format: Read the inputs from stdin solve the problem and write the answer to stdout (do not directly test on the sample inputs). Enclose your code within delimiters as follows. Ensure that when the python program runs, it reads the inputs, runs the algorithm and writes output to STDOUT.\\
\textasciigrave{}\textasciigrave{}\textasciigrave{}python\\
\# YOUR CODE HERE\\
\textasciigrave{}\textasciigrave{}\textasciigrave{}\par\medskip
\#\#\# Answer: (use the provided format with backticks)
}
\prompttemplate{LiveCodeBench: user message with starter code}{\small\ttfamily\raggedright
\#\#\# Question:\\
\{\{question\}\}\par\medskip
\#\#\# Format: You will use the following starter code to write the solution to the problem and enclose your code within delimiters.\\
\textasciigrave{}\textasciigrave{}\textasciigrave{}python\\
\{\{starter\_code\}\}\\
\textasciigrave{}\textasciigrave{}\textasciigrave{}\par\medskip
\#\#\# Answer: (use the provided format with backticks)
}

\paragraph{Fact.}
The closed-book Fact evaluation contains HotpotQA, TriviaQA, PopQA, and 2WikiMultiHopQA, with 200 problems per benchmark.
The Fact training pool includes 90,447 HotpotQA train examples; for evaluation, we sampled 200 questions from the HotpotQA dev split after screening for exact question-text overlap with the full Fact training pool under NFKC normalization, case folding, and whitespace collapse; no matches were found.
The SESA teacher is the released SESA-8B-search checkpoint~\citep{fu2026selfplaymeetsskillevolution}; its model page is \url{https://huggingface.co/kuailexuexi/SESA-8B-search}. This closed-book evaluation does not use the search tool from the original agentic setup.
It uses the frozen non-thinking chat prompt, temperature 0.7, top-$p$ 0.95, seed 0, a 4,096-token response cap, and an 8,192-token context cap.
Responses are scored by exact match after answer extraction and normalization against the benchmark's accepted aliases; failed extractions count as incorrect.
The shared pre-OPD base model, both OPD-trained models, and both teachers use 1,024 responses per problem on all four benchmarks, with a shared evaluation budget.

\paragraph{Fact prompt format.}
All four Fact benchmarks use the following user message without a separate system message. Only the question is substituted for \texttt{\{\{problem\}\}}; no retrieved context or gold answer is included.
\prompttemplate{Fact: user message}{\small\ttfamily\raggedright
\{\{problem\}\}\\
Please reason step by step, then give a concise final answer in \textless{}answer\textgreater{}\textless{}/answer\textgreater{} tags. Put the tagged answer on a new final line.
}

\section{Training Configurations}
\label{app:main-training}

This appendix reports training configurations for the main experiments and OPD variants.

\subsection{Main Experiments}

Table~\ref{tab:main-training-configurations} summarizes the configurations of our locally trained OPD models by domain. The Qwen3-1.7B-Base $\leftarrow$ Qwen3-4B-Base-GRPO setting uses the released \texttt{lllyx/Qwen3-1.7B-Base-OPD} checkpoint~\citep{li2026rethinkingopd}\footnote{\url{https://huggingface.co/Thinking-Space/Qwen3-1.7B-Base-OPD}}.

\begin{table}[!htbp]
\centering
\setlength{\tabcolsep}{2pt}
\renewcommand{\arraystretch}{1}
\caption{Training configurations by domain. Batch gives the global prompt batch size; length gives the prompt / response token limits; LR / schedule gives the learning rate and its schedule.}
\label{tab:main-training-configurations}
\vspace{-8.5pt}
\fontsize{6.93}{7.7}\selectfont
\begin{tabular}{@{}l|lccc@{}}
\toprule[1.6pt]
\textbf{Domain} & \textbf{Training set} & \textbf{Batch} & \textbf{LR / schedule} & \textbf{Length} \\
\midrule
\textbf{Math} & DAPO-Math-17K~\citep{yu2025dapo} & 64 & $10^{-6}$ / constant & 1024 / 7168 \\
\textbf{Code} & CodeR1-12K~\citep{ganler2025coder1} & 128 & $5{\times}10^{-7}$ / cosine & 2048 / 4096 \\
\textbf{Fact} & NQ~\citep{kwiatkowski2019naturalquestions} + HotpotQA & 64 & $10^{-6}$ / constant & 1024 / 3072 \\
\bottomrule[1.6pt]
\end{tabular}
\end{table}

For reference, the model names in Table~\ref{Tble:OPD_Setting} are shorthands for the released checkpoints: Qwen3-4B-Base-GRPO denotes \texttt{Thinking-Space/Qwen3-4B-Base-GRPO}, Qwen3-4B-RL-Math denotes \texttt{Keven16/Qwen3-4B-Non-Thinking-RL-Math-Step500}, Qwen3-4B-RL-Code denotes \texttt{Keven16/Qwen3-4B-Non-Thinking-RL-Code-Step300}, Qwen3-8B-RL-RAR denotes \texttt{chentong00/Qwen3-8B-GRPO-Binary-RAR}, DS-Distill-Qwen-1.5B denotes \texttt{deepseek-ai/DeepSeek-R1-Distill-Qwen-1.5B}, JustRL-DS-1.5B denotes \texttt{hbx/JustRL-DeepSeek-1.5B} at global Step~260, and Skywork-OR1-Math-7B is used at Step~260 of \texttt{Skywork/Skywork-OR1-Math-7B}. The remaining names (Qwen3-1.7B-Base, Qwen3-1.7B, and Qwen3-8B) match their released checkpoints under the same names.

\subsection{OPD Variants}

OPD, Direct-OPD, EOPD, pure forward KL, and ExOPD are all trained using temperature 1, top-\(p\) 1, top-\(k\) disabled, one PPO epoch per batch, zero learning-rate warmup, and weight decay 0.01. Table~\ref{tab:training-hyperparameters} reports the configurations for this comparison, including per-variant batch sizes and lengths.

\begin{table}[!htbp]
\centering
\setlength{\tabcolsep}{2pt}
\renewcommand{\arraystretch}{1}
\caption{Training configurations for OPD variants. Batch gives global / PPO
mini-batch size; lengths give the maximum prompt and response lengths. Rollout sampling
top-$k$ is disabled for all locally executed rows. Each variant follows the
default configuration of its respective repository.}
\label{tab:training-hyperparameters}
\vspace{-8.5pt}
\fontsize{6.93}{7.7}\selectfont
\begin{tabular}{@{}l|ccc@{}}
\toprule[1.6pt]
\textbf{Setting} & \textbf{Batch} & \textbf{Lengths} & \textbf{LR / schedule} \\
\midrule
OPD \citep{li2026rethinkingopd}
& 64 / 64
& 1024 / 7168
& \(1{\times}10^{-6}\) / cosine \\
Direct-OPD \citep{feng2026directopd}
& 128 / 32
& 1024 / 4096
& \(3{\times}10^{-6}\) / cosine \\
EOPD \citep{jin2026entropyopd}
& 128 / 32
& 1024 / 4096
& \(3{\times}10^{-6}\) / cosine \\
FKL
& 128 / 32
& 1024 / 4096
& \(3{\times}10^{-6}\) / cosine \\
ExOPD \citep{yang2026exopd}
& 1024 / 1024
& 2048 / 16384
& \(1{\times}10^{-5}\) / constant \\
\bottomrule[1.6pt]
\end{tabular}
\end{table}

\begin{revisionblock}
\subsection{Dataset 2 Training Configurations}
\label{app:additional-training-datasets}
Dataset~2 uses the answered training split of Hendrycks MATH (7,498 source rows) for Math, APPS for Code, and TriviaQA for Fact. The pre-OPD students and teachers are retained. Math runs use 260 updates, batch size 64, four responses per prompt, a constant learning rate of $10^{-6}$, and prompt/response caps of 1024/7168 tokens. Code runs retain the CodeR1 optimization settings in Table~\ref{tab:main-training-configurations} and use the final APPS checkpoint after 312 updates. Fact runs use 2,650 updates with the Dataset~1 optimizer and decoding settings, extending the epoch cap to reach that update count. The TriviaQA training data contain 87,621 rows after filtering, including removal of one exact prompt overlap with the evaluated PopQA subset. Exact normalized prompt checks found no overlap between the Hendrycks training prompts and the evaluated Math benchmarks.

The two training datasets test whether scaling patterns persist across data sources. For Qwen3-1.7B-Base, the Dataset~2 checkpoint uses a chat template, whereas the Dataset~1 base evaluation uses completion prompts. This comparison therefore does not isolate training data as the sole changed factor. The maximum sampling budgets are listed in Appendix~\ref{app:evaluation-details}.
\end{revisionblock}

\section{Numerical Results for Test-Time Scaling Curves}
\begin{revisionblock}
The first three tables report Dataset~1: DAPO-Math-17K, CodeR1-12K, and NQ/HotpotQA. Appendix~\ref{app:training-data-results} reports Dataset~2 results with Hendrycks MATH, APPS, and TriviaQA.
\end{revisionblock}

Tables~\ref{tab:primary-pass-numerical}, \ref{tab:code-pass-numerical}, and~\ref{tab:fact-pass-numerical} report benchmark-level pass@$K$ (\%) for the Math, Code, and Fact settings. Each setting identifies the student and teacher as student $\leftarrow$ teacher; Base, OPD, and Teacher occupy separate blocks, with one row per benchmark. Columns give the sampling budget $K$.

Table~\ref{tab:primary-pass-numerical} lists the four Math settings on AMC2023 and AIME2024--2026. The benchmark-level values show where gains at pass@$1$ coexist with lower pass@$1024$ than the pre-OPD base model.

\begin{table}[!htbp]
\centering
\caption{Math pass@$K$ (\%) for four student $\leftarrow$ teacher settings on AMC2023 and AIME2024--2026. Rows compare Base, OPD, and Teacher; columns give the sampling budget $K$ in each case.}
\label{tab:primary-pass-numerical}
\vspace{-8.5pt}
\fontsize{7.245}{8.05}\selectfont
\setlength{\tabcolsep}{(\textwidth-319.0pt)/26}
\renewcommand{\arraystretch}{0.88}
\setlength{\aboverulesep}{1pt}
\setlength{\belowrulesep}{1pt}
\begin{tabular}{@{}S{75.0pt}|S{35.9pt}S{56.9pt}*{8}{N{12.7pt}}*{3}{N{16.4pt}}@{}}
\toprule[1.6pt]
\textbf{Setting} & \textbf{Role} & \textbf{Dataset} & \textbf{1} & \textbf{2} & \textbf{4} & \textbf{8} & \textbf{16} & \textbf{32} & \textbf{64} & \textbf{128} & \textbf{256} & \textbf{512} & \textbf{1024} \\
\midrule

\multirow{12}{*}{\shortstack[l]{Qwen3-1.7B-Base\,$\leftarrow$\\Qwen3-4B-Base-GRPO}} & \multirow{4}{*}{Base} & AMC2023 & 32.3 & 44.4 & 56.3 & 66.8 & 75.8 & 83.5 & 89.7 & 94.3 & 97.7 & 99.6 & 100.0 \\
 &  &\cellcolor{zebraTint} AIME2024 &\cellcolor{zebraTint} 4.2 &\cellcolor{zebraTint} 7.0 &\cellcolor{zebraTint} 10.9 &\cellcolor{zebraTint} 16.0 &\cellcolor{zebraTint} 22.0 &\cellcolor{zebraTint} 28.2 &\cellcolor{zebraTint} 35.0 &\cellcolor{zebraTint} 42.9 &\cellcolor{zebraTint} 51.6 &\cellcolor{zebraTint} 61.0 &\cellcolor{zebraTint} 70.0 \\
 &  & AIME2025 & 3.1 & 5.3 & 8.3 & 12.4 & 17.7 & 23.5 & 29.6 & 37.0 & 45.7 & 55.1 & 66.7 \\
 &  &\cellcolor{zebraTint} AIME2026 &\cellcolor{zebraTint} 2.8 &\cellcolor{zebraTint} 4.9 &\cellcolor{zebraTint} 7.8 &\cellcolor{zebraTint} 10.9 &\cellcolor{zebraTint} 13.9 &\cellcolor{zebraTint} 17.7 &\cellcolor{zebraTint} 23.6 &\cellcolor{zebraTint} 31.8 &\cellcolor{zebraTint} 41.4 &\cellcolor{zebraTint} 50.4 &\cellcolor{zebraTint} 56.7 \\
\cmidrule(l){2-14}
 & \multirow{4}{*}{OPD} & AMC2023 & 45.4 & 55.4 & 63.9 & 71.7 & 78.7 & 83.4 & 86.6 & 89.2 & 91.0 & 92.5 & 95.0 \\
 &  &\cellcolor{zebraTint} AIME2024 &\cellcolor{zebraTint} 12.0 &\cellcolor{zebraTint} 16.7 &\cellcolor{zebraTint} 20.9 &\cellcolor{zebraTint} 24.2 &\cellcolor{zebraTint} 27.3 &\cellcolor{zebraTint} 30.5 &\cellcolor{zebraTint} 33.6 &\cellcolor{zebraTint} 37.0 &\cellcolor{zebraTint} 41.6 &\cellcolor{zebraTint} 47.3 &\cellcolor{zebraTint} 53.3 \\
 &  & AIME2025 & 8.6 & 12.6 & 17.4 & 22.2 & 26.5 & 30.4 & 34.4 & 39.0 & 44.6 & 51.2 & 56.7 \\
 &  &\cellcolor{zebraTint} AIME2026 &\cellcolor{zebraTint} 7.4 &\cellcolor{zebraTint} 11.4 &\cellcolor{zebraTint} 15.3 &\cellcolor{zebraTint} 18.8 &\cellcolor{zebraTint} 22.5 &\cellcolor{zebraTint} 26.6 &\cellcolor{zebraTint} 31.2 &\cellcolor{zebraTint} 36.9 &\cellcolor{zebraTint} 44.0 &\cellcolor{zebraTint} 51.1 &\cellcolor{zebraTint} 56.7 \\
\cmidrule(l){2-14}
 & \multirow{4}{*}{Teacher} & AMC2023 & 65.6 & 76.9 & 85.1 & 89.7 & 92.2 & 94.0 & 95.7 & 97.0 & 97.9 & 98.7 & 100.0 \\
 &  &\cellcolor{zebraTint} AIME2024 &\cellcolor{zebraTint} 24.2 &\cellcolor{zebraTint} 30.7 &\cellcolor{zebraTint} 37.2 &\cellcolor{zebraTint} 44.4 &\cellcolor{zebraTint} 51.8 &\cellcolor{zebraTint} 58.2 &\cellcolor{zebraTint} 63.1 &\cellcolor{zebraTint} 67.1 &\cellcolor{zebraTint} 70.4 &\cellcolor{zebraTint} 73.7 &\cellcolor{zebraTint} 76.7 \\
 &  & AIME2025 & 20.6 & 26.0 & 31.1 & 36.9 & 43.8 & 50.9 & 58.0 & 65.3 & 72.2 & 77.6 & 80.0 \\
 &  &\cellcolor{zebraTint} AIME2026 &\cellcolor{zebraTint} 18.7 &\cellcolor{zebraTint} 24.6 &\cellcolor{zebraTint} 30.1 &\cellcolor{zebraTint} 34.8 &\cellcolor{zebraTint} 39.8 &\cellcolor{zebraTint} 45.2 &\cellcolor{zebraTint} 51.0 &\cellcolor{zebraTint} 57.7 &\cellcolor{zebraTint} 65.1 &\cellcolor{zebraTint} 72.0 &\cellcolor{zebraTint} 76.7 \\
\midrule
\multirow{12}{*}{\shortstack[l]{DS-Distill-Qwen-1.5B\,$\leftarrow$\\Skywork-OR1-Math-7B}} & \multirow{4}{*}{Base} & AMC2023 & 72.2 & 82.5 & 89.8 & 93.6 & 95.3 & 96.2 & 97.2 & 98.5 & 99.6 & 100.0 & 100.0 \\
 &  &\cellcolor{zebraTint} AIME2024 &\cellcolor{zebraTint} 29.9 &\cellcolor{zebraTint} 40.8 &\cellcolor{zebraTint} 51.7 &\cellcolor{zebraTint} 61.8 &\cellcolor{zebraTint} 70.0 &\cellcolor{zebraTint} 76.0 &\cellcolor{zebraTint} 80.0 &\cellcolor{zebraTint} 82.1 &\cellcolor{zebraTint} 83.7 &\cellcolor{zebraTint} 85.6 &\cellcolor{zebraTint} 86.7 \\
 &  & AIME2025 & 23.4 & 29.3 & 34.6 & 40.2 & 46.2 & 52.1 & 58.0 & 64.1 & 69.5 & 73.6 & 76.7 \\
 &  &\cellcolor{zebraTint} AIME2026 &\cellcolor{zebraTint} 20.6 &\cellcolor{zebraTint} 28.6 &\cellcolor{zebraTint} 36.9 &\cellcolor{zebraTint} 45.0 &\cellcolor{zebraTint} 52.9 &\cellcolor{zebraTint} 60.8 &\cellcolor{zebraTint} 68.2 &\cellcolor{zebraTint} 74.4 &\cellcolor{zebraTint} 79.2 &\cellcolor{zebraTint} 82.9 &\cellcolor{zebraTint} 86.7 \\
\cmidrule(l){2-14}
 & \multirow{4}{*}{OPD} & AMC2023 & 76.0 & 85.2 & 91.6 & 94.3 & 95.1 & 95.4 & 95.8 & 96.4 & 97.1 & 97.5 & 97.5 \\
 &  &\cellcolor{zebraTint} AIME2024 &\cellcolor{zebraTint} 36.4 &\cellcolor{zebraTint} 47.2 &\cellcolor{zebraTint} 57.9 &\cellcolor{zebraTint} 66.6 &\cellcolor{zebraTint} 72.3 &\cellcolor{zebraTint} 75.6 &\cellcolor{zebraTint} 77.5 &\cellcolor{zebraTint} 78.8 &\cellcolor{zebraTint} 80.6 &\cellcolor{zebraTint} 83.1 &\cellcolor{zebraTint} 86.7 \\
 &  & AIME2025 & 28.5 & 33.4 & 37.7 & 42.0 & 46.4 & 50.8 & 55.1 & 60.0 & 64.7 & 67.9 & 70.0 \\
 &  &\cellcolor{zebraTint} AIME2026 &\cellcolor{zebraTint} 25.9 &\cellcolor{zebraTint} 34.5 &\cellcolor{zebraTint} 42.5 &\cellcolor{zebraTint} 50.4 &\cellcolor{zebraTint} 58.2 &\cellcolor{zebraTint} 64.3 &\cellcolor{zebraTint} 68.0 &\cellcolor{zebraTint} 71.4 &\cellcolor{zebraTint} 74.8 &\cellcolor{zebraTint} 77.5 &\cellcolor{zebraTint} 80.0 \\
\cmidrule(l){2-14}
 & \multirow{4}{*}{Teacher} & AMC2023 & 93.8 & 95.2 & 95.7 & 96.4 & 97.3 & 98.5 & 99.5 & 99.9 & 100.0 & 100.0 & 100.0 \\
 &  &\cellcolor{zebraTint} AIME2024 &\cellcolor{zebraTint} 67.0 &\cellcolor{zebraTint} 75.4 &\cellcolor{zebraTint} 80.2 &\cellcolor{zebraTint} 82.6 &\cellcolor{zebraTint} 84.1 &\cellcolor{zebraTint} 85.5 &\cellcolor{zebraTint} 86.7 &\cellcolor{zebraTint} 87.9 &\cellcolor{zebraTint} 88.9 &\cellcolor{zebraTint} 89.8 &\cellcolor{zebraTint} 90.0 \\
 &  & AIME2025 & 51.4 & 59.1 & 64.9 & 69.7 & 74.0 & 77.9 & 81.3 & 84.7 & 88.2 & 91.1 & 93.3 \\
 &  &\cellcolor{zebraTint} AIME2026 &\cellcolor{zebraTint} 60.9 &\cellcolor{zebraTint} 68.9 &\cellcolor{zebraTint} 74.9 &\cellcolor{zebraTint} 79.2 &\cellcolor{zebraTint} 82.1 &\cellcolor{zebraTint} 84.5 &\cellcolor{zebraTint} 86.6 &\cellcolor{zebraTint} 88.5 &\cellcolor{zebraTint} 89.7 &\cellcolor{zebraTint} 90.0 &\cellcolor{zebraTint} 90.0 \\
\midrule
\multirow{12}{*}{\shortstack[l]{DS-Distill-Qwen-1.5B\,$\leftarrow$\\JustRL-DS-1.5B}} & \multirow{4}{*}{Base} & AMC2023 & 72.2 & 82.5 & 89.8 & 93.6 & 95.3 & 96.2 & 97.2 & 98.5 & 99.6 & 100.0 & 100.0 \\
 &  &\cellcolor{zebraTint} AIME2024 &\cellcolor{zebraTint} 29.9 &\cellcolor{zebraTint} 40.8 &\cellcolor{zebraTint} 51.7 &\cellcolor{zebraTint} 61.8 &\cellcolor{zebraTint} 70.0 &\cellcolor{zebraTint} 76.0 &\cellcolor{zebraTint} 80.0 &\cellcolor{zebraTint} 82.1 &\cellcolor{zebraTint} 83.7 &\cellcolor{zebraTint} 85.6 &\cellcolor{zebraTint} 86.7 \\
 &  & AIME2025 & 23.4 & 29.3 & 34.6 & 40.2 & 46.2 & 52.1 & 58.0 & 64.1 & 69.5 & 73.6 & 76.7 \\
 &  &\cellcolor{zebraTint} AIME2026 &\cellcolor{zebraTint} 20.6 &\cellcolor{zebraTint} 28.6 &\cellcolor{zebraTint} 36.9 &\cellcolor{zebraTint} 45.0 &\cellcolor{zebraTint} 52.9 &\cellcolor{zebraTint} 60.8 &\cellcolor{zebraTint} 68.2 &\cellcolor{zebraTint} 74.4 &\cellcolor{zebraTint} 79.2 &\cellcolor{zebraTint} 82.9 &\cellcolor{zebraTint} 86.7 \\
\cmidrule(l){2-14}
 & \multirow{4}{*}{OPD} & AMC2023 & 87.6 & 92.9 & 94.9 & 95.6 & 96.2 & 97.1 & 98.4 & 99.5 & 100.0 & 100.0 & 100.0 \\
 &  &\cellcolor{zebraTint} AIME2024 &\cellcolor{zebraTint} 49.4 &\cellcolor{zebraTint} 59.5 &\cellcolor{zebraTint} 67.4 &\cellcolor{zebraTint} 73.3 &\cellcolor{zebraTint} 77.2 &\cellcolor{zebraTint} 79.3 &\cellcolor{zebraTint} 80.1 &\cellcolor{zebraTint} 80.4 &\cellcolor{zebraTint} 80.8 &\cellcolor{zebraTint} 81.7 &\cellcolor{zebraTint} 83.3 \\
 &  & AIME2025 & 34.7 & 40.6 & 46.4 & 52.2 & 57.3 & 61.2 & 64.5 & 67.5 & 69.8 & 71.6 & 73.3 \\
 &  &\cellcolor{zebraTint} AIME2026 &\cellcolor{zebraTint} 35.6 &\cellcolor{zebraTint} 44.8 &\cellcolor{zebraTint} 52.2 &\cellcolor{zebraTint} 58.5 &\cellcolor{zebraTint} 63.9 &\cellcolor{zebraTint} 69.3 &\cellcolor{zebraTint} 74.1 &\cellcolor{zebraTint} 77.5 &\cellcolor{zebraTint} 79.9 &\cellcolor{zebraTint} 82.1 &\cellcolor{zebraTint} 83.3 \\
\cmidrule(l){2-14}
 & \multirow{4}{*}{Teacher} & AMC2023 & 90.4 & 94.1 & 95.3 & 95.9 & 96.7 & 97.8 & 99.0 & 99.8 & 100.0 & 100.0 & 100.0 \\
 &  &\cellcolor{zebraTint} AIME2024 &\cellcolor{zebraTint} 52.1 &\cellcolor{zebraTint} 61.9 &\cellcolor{zebraTint} 70.1 &\cellcolor{zebraTint} 76.1 &\cellcolor{zebraTint} 79.3 &\cellcolor{zebraTint} 80.2 &\cellcolor{zebraTint} 80.4 &\cellcolor{zebraTint} 80.8 &\cellcolor{zebraTint} 81.7 &\cellcolor{zebraTint} 83.3 &\cellcolor{zebraTint} 86.7 \\
 &  & AIME2025 & 36.5 & 42.8 & 49.4 & 55.8 & 60.7 & 63.8 & 66.2 & 68.4 & 70.8 & 73.8 & 76.7 \\
 &  &\cellcolor{zebraTint} AIME2026 &\cellcolor{zebraTint} 38.0 &\cellcolor{zebraTint} 47.1 &\cellcolor{zebraTint} 54.2 &\cellcolor{zebraTint} 59.6 &\cellcolor{zebraTint} 64.4 &\cellcolor{zebraTint} 69.7 &\cellcolor{zebraTint} 74.2 &\cellcolor{zebraTint} 77.1 &\cellcolor{zebraTint} 79.0 &\cellcolor{zebraTint} 79.9 &\cellcolor{zebraTint} 80.0 \\

\midrule
\multirow{12}{*}{\shortstack[l]{Qwen3-1.7B\,$\leftarrow$\\Qwen3-4B-RL-Math}} & \multirow{4}{*}{Base} & AMC2023 & 46.1 & 59.2 & 70.6 & 79.2 & 85.2 & 89.6 & 93.2 & 96.1 & 98.1 & 99.5 & 100.0 \\
 &  &\cellcolor{zebraTint} AIME2024 &\cellcolor{zebraTint} 13.7 &\cellcolor{zebraTint} 19.3 &\cellcolor{zebraTint} 25.0 &\cellcolor{zebraTint} 31.0 &\cellcolor{zebraTint} 37.8 &\cellcolor{zebraTint} 45.0 &\cellcolor{zebraTint} 51.3 &\cellcolor{zebraTint} 56.4 &\cellcolor{zebraTint} 61.7 &\cellcolor{zebraTint} 68.2 &\cellcolor{zebraTint} 76.7 \\
 &  & AIME2025 & 10.1 & 14.1 & 18.0 & 22.3 & 27.7 & 34.3 & 41.6 & 48.5 & 55.7 & 63.2 & 70.0 \\
 &  &\cellcolor{zebraTint} AIME2026 &\cellcolor{zebraTint} 9.1 &\cellcolor{zebraTint} 13.3 &\cellcolor{zebraTint} 17.3 &\cellcolor{zebraTint} 21.1 &\cellcolor{zebraTint} 25.8 &\cellcolor{zebraTint} 31.1 &\cellcolor{zebraTint} 36.5 &\cellcolor{zebraTint} 42.3 &\cellcolor{zebraTint} 49.9 &\cellcolor{zebraTint} 59.8 &\cellcolor{zebraTint} 70.0 \\
\cmidrule(l){2-14}
 & \multirow{4}{*}{OPD} & AMC2023 & 82.5 & 88.9 & 92.6 & 94.6 & 95.5 & 96.0 & 96.6 & 97.2 & 97.5 & 97.5 & 97.5 \\
 &  &\cellcolor{zebraTint} AIME2024 &\cellcolor{zebraTint} 40.2 &\cellcolor{zebraTint} 48.7 &\cellcolor{zebraTint} 56.1 &\cellcolor{zebraTint} 62.2 &\cellcolor{zebraTint} 66.5 &\cellcolor{zebraTint} 69.9 &\cellcolor{zebraTint} 74.1 &\cellcolor{zebraTint} 78.6 &\cellcolor{zebraTint} 82.1 &\cellcolor{zebraTint} 84.6 &\cellcolor{zebraTint} 86.7 \\
 &  & AIME2025 & 33.1 & 39.8 & 46.7 & 52.5 & 57.3 & 62.2 & 67.2 & 70.8 & 73.0 & 74.8 & 76.7 \\
 &  &\cellcolor{zebraTint} AIME2026 &\cellcolor{zebraTint} 33.5 &\cellcolor{zebraTint} 41.6 &\cellcolor{zebraTint} 48.3 &\cellcolor{zebraTint} 53.8 &\cellcolor{zebraTint} 57.7 &\cellcolor{zebraTint} 60.5 &\cellcolor{zebraTint} 62.8 &\cellcolor{zebraTint} 65.1 &\cellcolor{zebraTint} 67.2 &\cellcolor{zebraTint} 69.8 &\cellcolor{zebraTint} 73.3 \\
\cmidrule(l){2-14}
 & \multirow{4}{*}{Teacher} & AMC2023 & 94.7 & 96.3 & 97.0 & 97.4 & 97.5 & 97.5 & 97.5 & 97.5 & 97.5 & 97.5 & 97.5 \\
 &  &\cellcolor{zebraTint} AIME2024 &\cellcolor{zebraTint} 61.2 &\cellcolor{zebraTint} 70.8 &\cellcolor{zebraTint} 77.1 &\cellcolor{zebraTint} 80.2 &\cellcolor{zebraTint} 82.0 &\cellcolor{zebraTint} 83.3 &\cellcolor{zebraTint} 84.4 &\cellcolor{zebraTint} 85.4 &\cellcolor{zebraTint} 86.7 &\cellcolor{zebraTint} 88.2 &\cellcolor{zebraTint} 90.0 \\
 &  & AIME2025 & 57.8 & 67.2 & 73.0 & 76.4 & 78.5 & 80.0 & 81.2 & 82.7 & 84.8 & 87.9 & 90.0 \\
 &  &\cellcolor{zebraTint} AIME2026 &\cellcolor{zebraTint} 57.8 &\cellcolor{zebraTint} 64.6 &\cellcolor{zebraTint} 70.2 &\cellcolor{zebraTint} 74.3 &\cellcolor{zebraTint} 76.8 &\cellcolor{zebraTint} 78.8 &\cellcolor{zebraTint} 80.7 &\cellcolor{zebraTint} 82.5 &\cellcolor{zebraTint} 84.7 &\cellcolor{zebraTint} 87.5 &\cellcolor{zebraTint} 90.0 \\
\bottomrule[1.6pt]
\end{tabular}
\end{table}

Table~\ref{tab:code-pass-numerical} gives the two Code settings on HumanEval+, MBPP+, and LiveCodeBench v5 and v6. It shows the Base--OPD comparison at each evaluated budget; HumanEval+ and MBPP+ reach $K=512$, while both LiveCodeBench versions reach $K=128$ in the reported evaluations.

\begin{table}[!htbp]
\centering
\caption{Code pass@$K$ (\%) for two student $\leftarrow$ teacher settings on HumanEval+, MBPP+, and LiveCodeBench v5 and v6. Rows compare Base, OPD, and Teacher; columns give $K$, with dashes marking entries that were not evaluated for the corresponding checkpoint and benchmark.}
\label{tab:code-pass-numerical}
\vspace{-8.5pt}
\fontsize{7.245}{8.05}\selectfont
\setlength{\tabcolsep}{(\textwidth-319.0pt)/26}
\renewcommand{\arraystretch}{0.88}
\setlength{\aboverulesep}{1pt}
\setlength{\belowrulesep}{1pt}
\begin{tabular}{@{}S{75.0pt}|S{35.9pt}S{56.9pt}*{8}{N{12.7pt}}*{3}{N{16.4pt}}@{}}
\toprule[1.6pt]
\textbf{Setting} & \textbf{Role} & \textbf{Dataset} & \textbf{1} & \textbf{2} & \textbf{4} & \textbf{8} & \textbf{16} & \textbf{32} & \textbf{64} & \textbf{128} & \textbf{256} & \textbf{512} & \textbf{1024} \\
\midrule
\multirow{12}{*}{\shortstack[l]{Qwen3-1.7B\,$\leftarrow$\\Qwen3-8B}} & \multirow{4}{*}{Base} & HumanEval+ & 56.2 & 60.3 & 63.6 & 66.6 & 69.2 & 71.1 & 72.3 & 72.9 & 73.4 & 73.8 & \textemdash \\
 &  &\cellcolor{zebraTint} MBPP+ &\cellcolor{zebraTint} 48.6 &\cellcolor{zebraTint} 51.4 &\cellcolor{zebraTint} 53.5 &\cellcolor{zebraTint} 55.3 &\cellcolor{zebraTint} 56.9 &\cellcolor{zebraTint} 58.1 &\cellcolor{zebraTint} 59.1 &\cellcolor{zebraTint} 59.9 &\cellcolor{zebraTint} 60.6 &\cellcolor{zebraTint} 61.4 &\cellcolor{zebraTint} \textemdash \\
 &  & LiveCodeBench v5 & 26.0 & 29.8 & 33.3 & 36.2 & 38.8 & 41.1 & 43.1 & 44.8 & \textemdash & \textemdash & \textemdash \\
 &  &\cellcolor{zebraTint} LiveCodeBench v6 &\cellcolor{zebraTint} 15.5 &\cellcolor{zebraTint} 17.7 &\cellcolor{zebraTint} 19.6 &\cellcolor{zebraTint} 21.4 &\cellcolor{zebraTint} 23.2 &\cellcolor{zebraTint} 25.1 &\cellcolor{zebraTint} 27.1 &\cellcolor{zebraTint} 29.1 &\cellcolor{zebraTint} \textemdash &\cellcolor{zebraTint} \textemdash &\cellcolor{zebraTint} \textemdash \\
\cmidrule(l){2-14}
 & \multirow{4}{*}{OPD} & HumanEval+ & 58.9 & 63.9 & 68.3 & 72.1 & 75.2 & 77.6 & 79.2 & 80.2 & 81.0 & 81.7 & \textemdash \\
 &  &\cellcolor{zebraTint} MBPP+ &\cellcolor{zebraTint} 49.6 &\cellcolor{zebraTint} 53.5 &\cellcolor{zebraTint} 56.6 &\cellcolor{zebraTint} 58.9 &\cellcolor{zebraTint} 61.1 &\cellcolor{zebraTint} 62.9 &\cellcolor{zebraTint} 64.4 &\cellcolor{zebraTint} 65.7 &\cellcolor{zebraTint} 66.9 &\cellcolor{zebraTint} 68.0 &\cellcolor{zebraTint} \textemdash \\
 &  & LiveCodeBench v5 & 27.9 & 32.7 & 37.1 & 41.1 & 44.8 & 48.2 & 51.4 & 54.1 & \textemdash & \textemdash & \textemdash \\
 &  &\cellcolor{zebraTint} LiveCodeBench v6 &\cellcolor{zebraTint} 18.9 &\cellcolor{zebraTint} 22.0 &\cellcolor{zebraTint} 24.7 &\cellcolor{zebraTint} 27.2 &\cellcolor{zebraTint} 29.2 &\cellcolor{zebraTint} 30.9 &\cellcolor{zebraTint} 32.6 &\cellcolor{zebraTint} 34.3 &\cellcolor{zebraTint} \textemdash &\cellcolor{zebraTint} \textemdash &\cellcolor{zebraTint} \textemdash \\
\cmidrule(l){2-14}
 & \multirow{4}{*}{Teacher} & HumanEval+ & 78.3 & 80.0 & 81.3 & 82.4 & 83.6 & 84.8 & 85.8 & 86.6 & 87.4 & 87.8 & \textemdash \\
 &  &\cellcolor{zebraTint} MBPP+ &\cellcolor{zebraTint} 68.0 &\cellcolor{zebraTint} 69.7 &\cellcolor{zebraTint} 71.2 &\cellcolor{zebraTint} 72.6 &\cellcolor{zebraTint} 73.8 &\cellcolor{zebraTint} 74.7 &\cellcolor{zebraTint} 75.3 &\cellcolor{zebraTint} 75.6 &\cellcolor{zebraTint} 75.8 &\cellcolor{zebraTint} 75.9 &\cellcolor{zebraTint} \textemdash \\
 &  & LiveCodeBench v5 & 43.2 & 48.4 & 52.9 & 56.6 & 59.8 & 62.6 & 64.9 & 67.0 & \textemdash & \textemdash & \textemdash \\
 &  &\cellcolor{zebraTint} LiveCodeBench v6 &\cellcolor{zebraTint} 25.8 &\cellcolor{zebraTint} 28.6 &\cellcolor{zebraTint} 31.2 &\cellcolor{zebraTint} 33.2 &\cellcolor{zebraTint} 34.8 &\cellcolor{zebraTint} 36.3 &\cellcolor{zebraTint} 38.0 &\cellcolor{zebraTint} 39.4 &\cellcolor{zebraTint} \textemdash &\cellcolor{zebraTint} \textemdash &\cellcolor{zebraTint} \textemdash \\
\midrule
\multirow{12}{*}{\shortstack[l]{Qwen3-1.7B\,$\leftarrow$\\Qwen3-4B-RL-Code}} & \multirow{4}{*}{Base} & HumanEval+ & 56.2 & 60.3 & 63.6 & 66.6 & 69.2 & 71.1 & 72.3 & 72.9 & 73.4 & 73.8 & \textemdash \\
 &  &\cellcolor{zebraTint} MBPP+ &\cellcolor{zebraTint} 48.6 &\cellcolor{zebraTint} 51.4 &\cellcolor{zebraTint} 53.5 &\cellcolor{zebraTint} 55.3 &\cellcolor{zebraTint} 56.9 &\cellcolor{zebraTint} 58.1 &\cellcolor{zebraTint} 59.1 &\cellcolor{zebraTint} 59.9 &\cellcolor{zebraTint} 60.6 &\cellcolor{zebraTint} 61.4 &\cellcolor{zebraTint} \textemdash \\
 &  & LiveCodeBench v5 & 26.0 & 29.8 & 33.3 & 36.2 & 38.8 & 41.1 & 43.1 & 44.8 & \textemdash & \textemdash & \textemdash \\
 &  &\cellcolor{zebraTint} LiveCodeBench v6 &\cellcolor{zebraTint} 15.5 &\cellcolor{zebraTint} 17.7 &\cellcolor{zebraTint} 19.6 &\cellcolor{zebraTint} 21.4 &\cellcolor{zebraTint} 23.2 &\cellcolor{zebraTint} 25.1 &\cellcolor{zebraTint} 27.1 &\cellcolor{zebraTint} 29.1 &\cellcolor{zebraTint} \textemdash &\cellcolor{zebraTint} \textemdash &\cellcolor{zebraTint} \textemdash \\
\cmidrule(l){2-14}
 & \multirow{4}{*}{OPD} & HumanEval+ & 62.0 & 70.5 & 77.3 & 82.5 & 86.1 & 88.4 & 89.9 & 90.9 & 91.6 & 92.1 & \textemdash \\
 &  &\cellcolor{zebraTint} MBPP+ &\cellcolor{zebraTint} 50.8 &\cellcolor{zebraTint} 56.5 &\cellcolor{zebraTint} 61.2 &\cellcolor{zebraTint} 65.1 &\cellcolor{zebraTint} 68.4 &\cellcolor{zebraTint} 71.3 &\cellcolor{zebraTint} 74.0 &\cellcolor{zebraTint} 76.5 &\cellcolor{zebraTint} 78.6 &\cellcolor{zebraTint} 80.2 &\cellcolor{zebraTint} \textemdash \\
 &  & LiveCodeBench v5 & 36.1 & 43.9 & 50.4 & 55.6 & 59.6 & 63.0 & 65.8 & 68.1 & \textemdash & \textemdash & \textemdash \\
 &  &\cellcolor{zebraTint} LiveCodeBench v6 &\cellcolor{zebraTint} 21.7 &\cellcolor{zebraTint} 26.0 &\cellcolor{zebraTint} 30.0 &\cellcolor{zebraTint} 33.2 &\cellcolor{zebraTint} 35.5 &\cellcolor{zebraTint} 37.2 &\cellcolor{zebraTint} 38.6 &\cellcolor{zebraTint} 40.0 &\cellcolor{zebraTint} \textemdash &\cellcolor{zebraTint} \textemdash &\cellcolor{zebraTint} \textemdash \\
\cmidrule(l){2-14}
 & \multirow{4}{*}{Teacher} & HumanEval+ & 79.6 & 82.7 & 85.1 & 87.2 & 88.9 & 90.1 & 90.9 & 91.6 & 92.0 & 92.1 & \textemdash \\
 &  &\cellcolor{zebraTint} MBPP+ &\cellcolor{zebraTint} 65.5 &\cellcolor{zebraTint} 68.5 &\cellcolor{zebraTint} 70.9 &\cellcolor{zebraTint} 72.7 &\cellcolor{zebraTint} 74.2 &\cellcolor{zebraTint} 75.3 &\cellcolor{zebraTint} 76.0 &\cellcolor{zebraTint} 76.5 &\cellcolor{zebraTint} 77.0 &\cellcolor{zebraTint} 77.5 &\cellcolor{zebraTint} \textemdash \\
 &  & LiveCodeBench v5 & 53.3 & 61.5 & 67.6 & 71.8 & 74.7 & 76.8 & 78.6 & 80.1 & \textemdash & \textemdash & \textemdash \\
 &  &\cellcolor{zebraTint} LiveCodeBench v6 &\cellcolor{zebraTint} 32.1 &\cellcolor{zebraTint} 36.5 &\cellcolor{zebraTint} 39.7 &\cellcolor{zebraTint} 42.1 &\cellcolor{zebraTint} 44.5 &\cellcolor{zebraTint} 46.7 &\cellcolor{zebraTint} 48.8 &\cellcolor{zebraTint} 50.9 &\cellcolor{zebraTint} \textemdash &\cellcolor{zebraTint} \textemdash &\cellcolor{zebraTint} \textemdash \\
\bottomrule[1.6pt]
\end{tabular}
\end{table}

Table~\ref{tab:fact-pass-numerical} reports the two Fact settings on HotpotQA, TriviaQA, PopQA, and 2WikiMultiHopQA. In both settings, the OPD-trained model remains above its pre-OPD base model at pass@$1024$ on all four benchmarks; the large-budget comparison therefore favors the OPD-trained model.

\begin{table}[!htbp]
\centering
\caption{Fact pass@$K$ (\%) for two student $\leftarrow$ teacher settings on HotpotQA, TriviaQA, PopQA, and 2WikiMultiHopQA. Rows compare Base, OPD, and Teacher; columns give the sampling budget $K$.}
\label{tab:fact-pass-numerical}
\vspace{-8.5pt}
\fontsize{7.245}{8.05}\selectfont
\setlength{\tabcolsep}{(\textwidth-319.0pt)/26}
\renewcommand{\arraystretch}{0.88}
\setlength{\aboverulesep}{1pt}
\setlength{\belowrulesep}{1pt}
\begin{tabular}{@{}S{75.0pt}|S{35.9pt}S{56.9pt}*{8}{N{12.7pt}}*{3}{N{16.4pt}}@{}}
\toprule[1.6pt]
\textbf{Setting} & \textbf{Role} & \textbf{Dataset} & \textbf{1} & \textbf{2} & \textbf{4} & \textbf{8} & \textbf{16} & \textbf{32} & \textbf{64} & \textbf{128} & \textbf{256} & \textbf{512} & \textbf{1024} \\
\midrule
\multirow{12}{*}{\shortstack[l]{Qwen3-1.7B\,$\leftarrow$\\Qwen3-8B-RL-RAR}} & \multirow{4}{*}{Base} & HotpotQA & 5.6 & 7.5 & 9.6 & 11.8 & 13.9 & 15.8 & 17.8 & 19.7 & 21.5 & 23.4 & 25.0 \\
 &  &\cellcolor{zebraTint} TriviaQA &\cellcolor{zebraTint} 21.1 &\cellcolor{zebraTint} 24.8 &\cellcolor{zebraTint} 28.2 &\cellcolor{zebraTint} 31.4 &\cellcolor{zebraTint} 34.7 &\cellcolor{zebraTint} 38.1 &\cellcolor{zebraTint} 41.4 &\cellcolor{zebraTint} 44.5 &\cellcolor{zebraTint} 47.1 &\cellcolor{zebraTint} 49.2 &\cellcolor{zebraTint} 51.0 \\
 &  & PopQA & 8.3 & 9.9 & 11.5 & 13.3 & 15.0 & 16.8 & 18.6 & 20.3 & 22.1 & 23.6 & 24.5 \\
 &  &\cellcolor{zebraTint} 2WikiMultiHopQA &\cellcolor{zebraTint} 3.0 &\cellcolor{zebraTint} 4.4 &\cellcolor{zebraTint} 5.9 &\cellcolor{zebraTint} 7.3 &\cellcolor{zebraTint} 8.7 &\cellcolor{zebraTint} 10.0 &\cellcolor{zebraTint} 11.2 &\cellcolor{zebraTint} 12.3 &\cellcolor{zebraTint} 13.3 &\cellcolor{zebraTint} 14.5 &\cellcolor{zebraTint} 16.0 \\
\cmidrule(l){2-14}
 & \multirow{4}{*}{OPD} & HotpotQA & 5.2 & 7.7 & 10.5 & 13.4 & 16.3 & 19.3 & 22.2 & 24.8 & 27.1 & 29.3 & 32.0 \\
 &  &\cellcolor{zebraTint} TriviaQA &\cellcolor{zebraTint} 22.4 &\cellcolor{zebraTint} 27.4 &\cellcolor{zebraTint} 32.0 &\cellcolor{zebraTint} 36.2 &\cellcolor{zebraTint} 40.4 &\cellcolor{zebraTint} 44.2 &\cellcolor{zebraTint} 47.4 &\cellcolor{zebraTint} 50.0 &\cellcolor{zebraTint} 52.5 &\cellcolor{zebraTint} 55.4 &\cellcolor{zebraTint} 58.5 \\
 &  & PopQA & 10.1 & 12.1 & 14.2 & 16.6 & 19.3 & 22.1 & 25.0 & 27.9 & 30.6 & 32.8 & 34.5 \\
 &  &\cellcolor{zebraTint} 2WikiMultiHopQA &\cellcolor{zebraTint} 4.4 &\cellcolor{zebraTint} 6.1 &\cellcolor{zebraTint} 7.9 &\cellcolor{zebraTint} 9.9 &\cellcolor{zebraTint} 11.9 &\cellcolor{zebraTint} 14.0 &\cellcolor{zebraTint} 16.3 &\cellcolor{zebraTint} 18.9 &\cellcolor{zebraTint} 21.6 &\cellcolor{zebraTint} 24.5 &\cellcolor{zebraTint} 27.0 \\
\cmidrule(l){2-14}
 & \multirow{4}{*}{Teacher} & HotpotQA & 10.5 & 13.1 & 16.3 & 19.9 & 23.4 & 26.7 & 29.9 & 33.0 & 36.1 & 38.9 & 41.0 \\
 &  &\cellcolor{zebraTint} TriviaQA &\cellcolor{zebraTint} 41.9 &\cellcolor{zebraTint} 47.6 &\cellcolor{zebraTint} 52.3 &\cellcolor{zebraTint} 56.4 &\cellcolor{zebraTint} 59.9 &\cellcolor{zebraTint} 63.2 &\cellcolor{zebraTint} 66.2 &\cellcolor{zebraTint} 68.9 &\cellcolor{zebraTint} 71.2 &\cellcolor{zebraTint} 73.2 &\cellcolor{zebraTint} 75.0 \\
 &  & PopQA & 14.1 & 17.3 & 20.5 & 23.3 & 25.9 & 28.4 & 30.8 & 33.1 & 35.4 & 37.7 & 40.5 \\
 &  &\cellcolor{zebraTint} 2WikiMultiHopQA &\cellcolor{zebraTint} 5.0 &\cellcolor{zebraTint} 7.0 &\cellcolor{zebraTint} 9.3 &\cellcolor{zebraTint} 11.9 &\cellcolor{zebraTint} 14.3 &\cellcolor{zebraTint} 16.7 &\cellcolor{zebraTint} 19.3 &\cellcolor{zebraTint} 22.0 &\cellcolor{zebraTint} 24.7 &\cellcolor{zebraTint} 27.7 &\cellcolor{zebraTint} 30.5 \\
\midrule
\multirow{12}{*}{\shortstack[l]{Qwen3-1.7B\,$\leftarrow$\\SESA-8B-search}} & \multirow{4}{*}{Base} & HotpotQA & 5.6 & 7.5 & 9.6 & 11.8 & 13.9 & 15.8 & 17.8 & 19.7 & 21.5 & 23.4 & 25.0 \\
 &  &\cellcolor{zebraTint} TriviaQA &\cellcolor{zebraTint} 21.1 &\cellcolor{zebraTint} 24.8 &\cellcolor{zebraTint} 28.2 &\cellcolor{zebraTint} 31.4 &\cellcolor{zebraTint} 34.7 &\cellcolor{zebraTint} 38.1 &\cellcolor{zebraTint} 41.4 &\cellcolor{zebraTint} 44.5 &\cellcolor{zebraTint} 47.1 &\cellcolor{zebraTint} 49.2 &\cellcolor{zebraTint} 51.0 \\
 &  & PopQA & 8.3 & 9.9 & 11.5 & 13.3 & 15.0 & 16.8 & 18.6 & 20.3 & 22.1 & 23.6 & 24.5 \\
 &  &\cellcolor{zebraTint} 2WikiMultiHopQA &\cellcolor{zebraTint} 3.0 &\cellcolor{zebraTint} 4.4 &\cellcolor{zebraTint} 5.9 &\cellcolor{zebraTint} 7.3 &\cellcolor{zebraTint} 8.7 &\cellcolor{zebraTint} 10.0 &\cellcolor{zebraTint} 11.2 &\cellcolor{zebraTint} 12.3 &\cellcolor{zebraTint} 13.3 &\cellcolor{zebraTint} 14.5 &\cellcolor{zebraTint} 16.0 \\
\cmidrule(l){2-14}
 & \multirow{4}{*}{OPD} & HotpotQA & 5.6 & 8.0 & 10.7 & 13.5 & 16.2 & 18.8 & 21.6 & 24.6 & 27.5 & 30.4 & 33.0 \\
 &  &\cellcolor{zebraTint} TriviaQA &\cellcolor{zebraTint} 22.4 &\cellcolor{zebraTint} 27.0 &\cellcolor{zebraTint} 31.0 &\cellcolor{zebraTint} 34.8 &\cellcolor{zebraTint} 38.6 &\cellcolor{zebraTint} 42.3 &\cellcolor{zebraTint} 45.5 &\cellcolor{zebraTint} 48.6 &\cellcolor{zebraTint} 51.5 &\cellcolor{zebraTint} 54.3 &\cellcolor{zebraTint} 56.5 \\
 &  & PopQA & 11.7 & 14.0 & 16.5 & 19.1 & 21.8 & 24.5 & 27.1 & 29.7 & 32.2 & 34.4 & 36.5 \\
 &  &\cellcolor{zebraTint} 2WikiMultiHopQA &\cellcolor{zebraTint} 3.3 &\cellcolor{zebraTint} 5.0 &\cellcolor{zebraTint} 6.9 &\cellcolor{zebraTint} 8.9 &\cellcolor{zebraTint} 10.9 &\cellcolor{zebraTint} 12.8 &\cellcolor{zebraTint} 14.8 &\cellcolor{zebraTint} 17.1 &\cellcolor{zebraTint} 19.7 &\cellcolor{zebraTint} 22.3 &\cellcolor{zebraTint} 24.5 \\
\cmidrule(l){2-14}
 & \multirow{4}{*}{Teacher} & HotpotQA & 12.4 & 15.5 & 18.8 & 22.2 & 25.6 & 28.8 & 32.0 & 35.4 & 39.2 & 42.9 & 46.0 \\
 &  &\cellcolor{zebraTint} TriviaQA &\cellcolor{zebraTint} 43.4 &\cellcolor{zebraTint} 48.4 &\cellcolor{zebraTint} 52.7 &\cellcolor{zebraTint} 56.2 &\cellcolor{zebraTint} 58.9 &\cellcolor{zebraTint} 61.2 &\cellcolor{zebraTint} 63.6 &\cellcolor{zebraTint} 65.9 &\cellcolor{zebraTint} 67.8 &\cellcolor{zebraTint} 69.7 &\cellcolor{zebraTint} 72.0 \\
 &  & PopQA & 16.5 & 19.7 & 22.7 & 25.6 & 28.5 & 31.4 & 34.5 & 37.4 & 40.2 & 43.0 & 46.5 \\
 &  &\cellcolor{zebraTint} 2WikiMultiHopQA &\cellcolor{zebraTint} 5.7 &\cellcolor{zebraTint} 8.0 &\cellcolor{zebraTint} 10.6 &\cellcolor{zebraTint} 13.2 &\cellcolor{zebraTint} 15.6 &\cellcolor{zebraTint} 18.0 &\cellcolor{zebraTint} 20.6 &\cellcolor{zebraTint} 23.3 &\cellcolor{zebraTint} 25.9 &\cellcolor{zebraTint} 28.1 &\cellcolor{zebraTint} 30.0 \\
\bottomrule[1.6pt]
\end{tabular}
\end{table}

\begin{revisionblock}
\subsection{Dataset 2 Results}
\label{app:training-data-results}
Table~\ref{tab:pass1-overview-dataset2} gives the Dataset~2 counterpart of Table~\ref{tab:pass1-overview}. Tables~\ref{tab:additional-math-results}, \ref{tab:additional-code-results}, and~\ref{tab:additional-fact-results} follow the full numerical format of Tables~\ref{tab:primary-pass-numerical}, \ref{tab:code-pass-numerical}, and~\ref{tab:fact-pass-numerical}. Each table includes the Base and Teacher results shared across the two training datasets, together with the Dataset~2 OPD-trained results on the full $K$ grid. Dashes mark unavailable evaluations. These tables accompany the training-data comparisons in Figures~\ref{fig:code-settings}, \ref{fig:fact-settings}, and~\ref{fig:math-settings}.
\end{revisionblock}
\begin{table}[!htbp]
{
\centering
\setlength{\tabcolsep}{0.85pt}
\renewcommand{\arraystretch}{1}

\caption{\rev{\textbf{Pass@$1$ and full-budget pass@$K$ (\%) with Dataset~2.} This is the Dataset~2 counterpart of Table~\ref{tab:pass1-overview}, using the same settings, columns, and benchmark averaging.}}
\label{tab:pass1-overview-dataset2}
\vspace{-8.5pt}
\fontsize{6.93}{7.7}\selectfont
\begin{tabular}{@{}l|ll|cc|cc@{}}
\toprule[1.01pt]
\textbf{Domain} & \textbf{Student} & \textbf{Teacher} & \textbf{Pass@$1$ (Base $\to$ OPD)} & $\boldsymbol{\Delta}$ \textbf{(pp)} & \textbf{Pass@$K$ (Base $\to$ OPD)} & $\boldsymbol{\Delta}$ \textbf{(pp)} \\
\midrule[0.32pt]
\multirow{4}{*}{\textbf{Math}} & Qwen3-1.7B-Base & Qwen3-4B-Base-GRPO & 10.6 $\longrightarrow$ 18.6 & \dcell{8.0}{1}{{\textbf{+8.0}}} & 73.3 $\longrightarrow$ 68.5 & \dcell{4.8}{-1}{{\textbf{\textminus 4.8}}} \\
~ & DS-Distill-Qwen-1.5B & Skywork-OR1-Math-7B & 36.5 $\longrightarrow$ 40.0 & \dcell{3.5}{1}{{\textbf{+3.5}}} & 87.5 $\longrightarrow$ 84.4 & \dcell{3.1}{-1}{{\textbf{\textminus 3.1}}} \\
~ & DS-Distill-Qwen-1.5B & JustRL-DS-1.5B & 36.5 $\longrightarrow$ 50.9 & \dcell{14.4}{1}{{\textbf{+14.4}}} & 87.5 $\longrightarrow$ 85.0 & \dcell{2.5}{-1}{{\textbf{\textminus 2.5}}} \\
~ & Qwen3-1.7B & Qwen3-4B-RL-Math & 19.7 $\longrightarrow$ 47.4 & \dcell{27.7}{1}{{\textbf{+27.7}}} & 79.2 $\longrightarrow$ 83.5 & \dcell{4.4}{1}{{\textbf{+4.4}}} \\
\midrule[0.32pt]
\multirow{2}{*}{\textbf{Code}} & Qwen3-1.7B & Qwen3-8B & 36.6 $\longrightarrow$ 38.0 & \dcell{1.4}{1}{{\textbf{+1.4}}} & 52.3 $\longrightarrow$ 57.3 & \dcell{5.1}{1}{{\textbf{+5.1}}} \\
~ & Qwen3-1.7B & Qwen3-4B-RL-Code & 36.6 $\longrightarrow$ 41.9 & \dcell{5.3}{1}{{\textbf{+5.3}}} & 52.3 $\longrightarrow$ 68.7 & \dcell{16.4}{1}{{\textbf{+16.4}}} \\
\midrule[0.32pt]
\multirow{2}{*}{\textbf{Fact}} & Qwen3-1.7B & Qwen3-8B-RL-RAR & 9.5 $\longrightarrow$ 11.1 & \dcell{1.6}{1}{{\textbf{+1.6}}} & 29.1 $\longrightarrow$ 38.8 & \dcell{9.6}{1}{{\textbf{+9.6}}} \\
~ & Qwen3-1.7B & SESA-8B-search & 9.5 $\longrightarrow$ 11.0 & \dcell{1.4}{1}{{\textbf{+1.4}}} & 29.1 $\longrightarrow$ 37.2 & \dcell{8.1}{1}{{\textbf{+8.1}}} \\
\bottomrule[1.01pt]
\end{tabular}
\par
}
\end{table}

\begin{table}[!htbp]
\centering

\caption{Math pass@$K$ (\%) with Dataset~2 (Hendrycks MATH). Rows compare Base, OPD, and Teacher for four student-teacher settings on AMC2023 and AIME2024/2025/2026. Columns give $K$, with all OPD-trained models evaluated through $K=1024$.}
\label{tab:additional-math-results}
\vspace{-8.5pt}
\fontsize{7.245}{8.05}\selectfont
\setlength{\tabcolsep}{(\textwidth-319.0pt)/26}
\renewcommand{\arraystretch}{0.88}
\setlength{\aboverulesep}{1pt}
\setlength{\belowrulesep}{1pt}
\begin{tabular}{@{}S{75.0pt}|S{35.9pt}S{56.9pt}*{8}{N{12.7pt}}*{3}{N{16.4pt}}@{}}
\toprule[1.6pt]
\textbf{Setting} & \textbf{Role} & \textbf{Dataset} & \textbf{1} & \textbf{2} & \textbf{4} & \textbf{8} & \textbf{16} & \textbf{32} & \textbf{64} & \textbf{128} & \textbf{256} & \textbf{512} & \textbf{1024} \\
\midrule

\multirow{12}{*}{\shortstack[l]{Qwen3-1.7B-Base\,$\leftarrow$\\Qwen3-4B-Base-GRPO}} & \multirow{4}{*}{Base} & AMC2023 & 32.3 & 44.4 & 56.3 & 66.8 & 75.8 & 83.5 & 89.7 & 94.3 & 97.7 & 99.6 & 100.0 \\
 &  &\cellcolor{zebraTint} AIME2024 &\cellcolor{zebraTint} 4.2 &\cellcolor{zebraTint} 7.0 &\cellcolor{zebraTint} 10.9 &\cellcolor{zebraTint} 16.0 &\cellcolor{zebraTint} 22.0 &\cellcolor{zebraTint} 28.2 &\cellcolor{zebraTint} 35.0 &\cellcolor{zebraTint} 42.9 &\cellcolor{zebraTint} 51.6 &\cellcolor{zebraTint} 61.0 &\cellcolor{zebraTint} 70.0 \\
 &  & AIME2025 & 3.1 & 5.3 & 8.3 & 12.4 & 17.7 & 23.5 & 29.6 & 37.0 & 45.7 & 55.1 & 66.7 \\
 &  &\cellcolor{zebraTint} AIME2026 &\cellcolor{zebraTint} 2.8 &\cellcolor{zebraTint} 4.9 &\cellcolor{zebraTint} 7.8 &\cellcolor{zebraTint} 10.9 &\cellcolor{zebraTint} 13.9 &\cellcolor{zebraTint} 17.7 &\cellcolor{zebraTint} 23.6 &\cellcolor{zebraTint} 31.8 &\cellcolor{zebraTint} 41.4 &\cellcolor{zebraTint} 50.4 &\cellcolor{zebraTint} 56.7 \\
\cmidrule(l){2-14}
 & \multirow{4}{*}{OPD} & AMC2023 & 45.8 & 55.7 & 64.0 & 71.7 & 78.7 & 83.8 & 87.6 & 90.8 & 93.5 & 95.8 & 97.5 \\
 &  &\cellcolor{zebraTint} AIME2024 & \cellcolor{zebraTint} 12.6 & \cellcolor{zebraTint} 17.7 & \cellcolor{zebraTint} 22.2 & \cellcolor{zebraTint} 25.7 & \cellcolor{zebraTint} 28.9 & \cellcolor{zebraTint} 31.8 & \cellcolor{zebraTint} 34.3 & \cellcolor{zebraTint} 37.3 & \cellcolor{zebraTint} 41.7 & \cellcolor{zebraTint} 48.8 & \cellcolor{zebraTint} 60.0 \\
 &  & AIME2025 & 8.9 & 13.1 & 18.2 & 23.2 & 27.6 & 31.7 & 36.0 & 41.3 & 48.1 & 55.7 & 63.3 \\
 &  &\cellcolor{zebraTint} AIME2026 & \cellcolor{zebraTint} 7.0 & \cellcolor{zebraTint} 10.9 & \cellcolor{zebraTint} 15.0 & \cellcolor{zebraTint} 18.7 & \cellcolor{zebraTint} 22.3 & \cellcolor{zebraTint} 26.1 & \cellcolor{zebraTint} 30.4 & \cellcolor{zebraTint} 35.1 & \cellcolor{zebraTint} 40.0 & \cellcolor{zebraTint} 45.6 & \cellcolor{zebraTint} 53.3 \\
\cmidrule(l){2-14}
 & \multirow{4}{*}{Teacher} & AMC2023 & 65.6 & 76.9 & 85.1 & 89.7 & 92.2 & 94.0 & 95.7 & 97.0 & 97.9 & 98.7 & 100.0 \\
 &  &\cellcolor{zebraTint} AIME2024 &\cellcolor{zebraTint} 24.2 &\cellcolor{zebraTint} 30.7 &\cellcolor{zebraTint} 37.2 &\cellcolor{zebraTint} 44.4 &\cellcolor{zebraTint} 51.8 &\cellcolor{zebraTint} 58.2 &\cellcolor{zebraTint} 63.1 &\cellcolor{zebraTint} 67.1 &\cellcolor{zebraTint} 70.4 &\cellcolor{zebraTint} 73.7 &\cellcolor{zebraTint} 76.7 \\
 &  & AIME2025 & 20.6 & 26.0 & 31.1 & 36.9 & 43.8 & 50.9 & 58.0 & 65.3 & 72.2 & 77.6 & 80.0 \\
 &  &\cellcolor{zebraTint} AIME2026 &\cellcolor{zebraTint} 18.7 &\cellcolor{zebraTint} 24.6 &\cellcolor{zebraTint} 30.1 &\cellcolor{zebraTint} 34.8 &\cellcolor{zebraTint} 39.8 &\cellcolor{zebraTint} 45.2 &\cellcolor{zebraTint} 51.0 &\cellcolor{zebraTint} 57.7 &\cellcolor{zebraTint} 65.1 &\cellcolor{zebraTint} 72.0 &\cellcolor{zebraTint} 76.7 \\
\midrule
\multirow{12}{*}{\shortstack[l]{DS-Distill-Qwen-1.5B\,$\leftarrow$\\Skywork-OR1-Math-7B}} & \multirow{4}{*}{Base} & AMC2023 & 72.2 & 82.5 & 89.8 & 93.6 & 95.3 & 96.2 & 97.2 & 98.5 & 99.6 & 100.0 & 100.0 \\
 &  &\cellcolor{zebraTint} AIME2024 &\cellcolor{zebraTint} 29.9 &\cellcolor{zebraTint} 40.8 &\cellcolor{zebraTint} 51.7 &\cellcolor{zebraTint} 61.8 &\cellcolor{zebraTint} 70.0 &\cellcolor{zebraTint} 76.0 &\cellcolor{zebraTint} 80.0 &\cellcolor{zebraTint} 82.1 &\cellcolor{zebraTint} 83.7 &\cellcolor{zebraTint} 85.6 &\cellcolor{zebraTint} 86.7 \\
 &  & AIME2025 & 23.4 & 29.3 & 34.6 & 40.2 & 46.2 & 52.1 & 58.0 & 64.1 & 69.5 & 73.6 & 76.7 \\
 &  &\cellcolor{zebraTint} AIME2026 &\cellcolor{zebraTint} 20.6 &\cellcolor{zebraTint} 28.6 &\cellcolor{zebraTint} 36.9 &\cellcolor{zebraTint} 45.0 &\cellcolor{zebraTint} 52.9 &\cellcolor{zebraTint} 60.8 &\cellcolor{zebraTint} 68.2 &\cellcolor{zebraTint} 74.4 &\cellcolor{zebraTint} 79.2 &\cellcolor{zebraTint} 82.9 &\cellcolor{zebraTint} 86.7 \\
\cmidrule(l){2-14}
 & \multirow{4}{*}{OPD} & AMC2023 & 73.8 & 83.3 & 90.5 & 94.0 & 95.0 & 95.2 & 95.3 & 95.6 & 96.1 & 96.9 & 97.5 \\
 &  &\cellcolor{zebraTint} AIME2024 & \cellcolor{zebraTint} 35.0 & \cellcolor{zebraTint} 45.6 & \cellcolor{zebraTint} 56.1 & \cellcolor{zebraTint} 64.7 & \cellcolor{zebraTint} 70.5 & \cellcolor{zebraTint} 73.9 & \cellcolor{zebraTint} 75.9 & \cellcolor{zebraTint} 77.6 & \cellcolor{zebraTint} 79.5 & \cellcolor{zebraTint} 81.7 & \cellcolor{zebraTint} 83.3 \\
 &  & AIME2025 & 27.3 & 32.6 & 37.1 & 41.6 & 46.1 & 50.7 & 55.4 & 60.3 & 65.1 & 69.4 & 73.3 \\
 &  &\cellcolor{zebraTint} AIME2026 & \cellcolor{zebraTint} 24.1 & \cellcolor{zebraTint} 32.3 & \cellcolor{zebraTint} 40.1 & \cellcolor{zebraTint} 47.6 & \cellcolor{zebraTint} 55.5 & \cellcolor{zebraTint} 62.6 & \cellcolor{zebraTint} 67.9 & \cellcolor{zebraTint} 72.2 & \cellcolor{zebraTint} 76.6 & \cellcolor{zebraTint} 80.4 & \cellcolor{zebraTint} 83.3 \\
\cmidrule(l){2-14}
 & \multirow{4}{*}{Teacher} & AMC2023 & 93.8 & 95.2 & 95.7 & 96.4 & 97.3 & 98.5 & 99.5 & 99.9 & 100.0 & 100.0 & 100.0 \\
 &  &\cellcolor{zebraTint} AIME2024 &\cellcolor{zebraTint} 67.0 &\cellcolor{zebraTint} 75.4 &\cellcolor{zebraTint} 80.2 &\cellcolor{zebraTint} 82.6 &\cellcolor{zebraTint} 84.1 &\cellcolor{zebraTint} 85.5 &\cellcolor{zebraTint} 86.7 &\cellcolor{zebraTint} 87.9 &\cellcolor{zebraTint} 88.9 &\cellcolor{zebraTint} 89.8 &\cellcolor{zebraTint} 90.0 \\
 &  & AIME2025 & 51.4 & 59.1 & 64.9 & 69.7 & 74.0 & 77.9 & 81.3 & 84.7 & 88.2 & 91.1 & 93.3 \\
 &  &\cellcolor{zebraTint} AIME2026 &\cellcolor{zebraTint} 60.9 &\cellcolor{zebraTint} 68.9 &\cellcolor{zebraTint} 74.9 &\cellcolor{zebraTint} 79.2 &\cellcolor{zebraTint} 82.1 &\cellcolor{zebraTint} 84.5 &\cellcolor{zebraTint} 86.6 &\cellcolor{zebraTint} 88.5 &\cellcolor{zebraTint} 89.7 &\cellcolor{zebraTint} 90.0 &\cellcolor{zebraTint} 90.0 \\
\midrule
\multirow{12}{*}{\shortstack[l]{DS-Distill-Qwen-1.5B\,$\leftarrow$\\JustRL-DS-1.5B}} & \multirow{4}{*}{Base} & AMC2023 & 72.2 & 82.5 & 89.8 & 93.6 & 95.3 & 96.2 & 97.2 & 98.5 & 99.6 & 100.0 & 100.0 \\
 &  &\cellcolor{zebraTint} AIME2024 &\cellcolor{zebraTint} 29.9 &\cellcolor{zebraTint} 40.8 &\cellcolor{zebraTint} 51.7 &\cellcolor{zebraTint} 61.8 &\cellcolor{zebraTint} 70.0 &\cellcolor{zebraTint} 76.0 &\cellcolor{zebraTint} 80.0 &\cellcolor{zebraTint} 82.1 &\cellcolor{zebraTint} 83.7 &\cellcolor{zebraTint} 85.6 &\cellcolor{zebraTint} 86.7 \\
 &  & AIME2025 & 23.4 & 29.3 & 34.6 & 40.2 & 46.2 & 52.1 & 58.0 & 64.1 & 69.5 & 73.6 & 76.7 \\
 &  &\cellcolor{zebraTint} AIME2026 &\cellcolor{zebraTint} 20.6 &\cellcolor{zebraTint} 28.6 &\cellcolor{zebraTint} 36.9 &\cellcolor{zebraTint} 45.0 &\cellcolor{zebraTint} 52.9 &\cellcolor{zebraTint} 60.8 &\cellcolor{zebraTint} 68.2 &\cellcolor{zebraTint} 74.4 &\cellcolor{zebraTint} 79.2 &\cellcolor{zebraTint} 82.9 &\cellcolor{zebraTint} 86.7 \\
\cmidrule(l){2-14}
 & \multirow{4}{*}{OPD} & AMC2023 & 86.8 & 92.6 & 94.9 & 95.6 & 96.2 & 97.1 & 98.4 & 99.5 & 100.0 & 100.0 & 100.0 \\
 &  &\cellcolor{zebraTint} AIME2024 & \cellcolor{zebraTint} 47.8 & \cellcolor{zebraTint} 58.5 & \cellcolor{zebraTint} 67.0 & \cellcolor{zebraTint} 73.0 & \cellcolor{zebraTint} 77.0 & \cellcolor{zebraTint} 79.5 & \cellcolor{zebraTint} 80.6 & \cellcolor{zebraTint} 81.4 & \cellcolor{zebraTint} 82.3 & \cellcolor{zebraTint} 83.1 & \cellcolor{zebraTint} 83.3 \\
 &  & AIME2025 & 34.9 & 40.8 & 46.8 & 52.8 & 57.7 & 61.2 & 64.1 & 66.8 & 69.2 & 71.2 & 73.3 \\
 &  &\cellcolor{zebraTint} AIME2026 & \cellcolor{zebraTint} 34.2 & \cellcolor{zebraTint} 43.4 & \cellcolor{zebraTint} 51.1 & \cellcolor{zebraTint} 57.9 & \cellcolor{zebraTint} 63.7 & \cellcolor{zebraTint} 69.0 & \cellcolor{zebraTint} 73.7 & \cellcolor{zebraTint} 77.2 & \cellcolor{zebraTint} 80.1 & \cellcolor{zebraTint} 82.3 & \cellcolor{zebraTint} 83.3 \\
\cmidrule(l){2-14}
 & \multirow{4}{*}{Teacher} & AMC2023 & 90.4 & 94.1 & 95.3 & 95.9 & 96.7 & 97.8 & 99.0 & 99.8 & 100.0 & 100.0 & 100.0 \\
 &  &\cellcolor{zebraTint} AIME2024 &\cellcolor{zebraTint} 52.1 &\cellcolor{zebraTint} 61.9 &\cellcolor{zebraTint} 70.1 &\cellcolor{zebraTint} 76.1 &\cellcolor{zebraTint} 79.3 &\cellcolor{zebraTint} 80.2 &\cellcolor{zebraTint} 80.4 &\cellcolor{zebraTint} 80.8 &\cellcolor{zebraTint} 81.7 &\cellcolor{zebraTint} 83.3 &\cellcolor{zebraTint} 86.7 \\
 &  & AIME2025 & 36.5 & 42.8 & 49.4 & 55.8 & 60.7 & 63.8 & 66.2 & 68.4 & 70.8 & 73.8 & 76.7 \\
 &  &\cellcolor{zebraTint} AIME2026 &\cellcolor{zebraTint} 38.0 &\cellcolor{zebraTint} 47.1 &\cellcolor{zebraTint} 54.2 &\cellcolor{zebraTint} 59.6 &\cellcolor{zebraTint} 64.4 &\cellcolor{zebraTint} 69.7 &\cellcolor{zebraTint} 74.2 &\cellcolor{zebraTint} 77.1 &\cellcolor{zebraTint} 79.0 &\cellcolor{zebraTint} 79.9 &\cellcolor{zebraTint} 80.0 \\

\midrule
\multirow{12}{*}{\shortstack[l]{Qwen3-1.7B\,$\leftarrow$\\Qwen3-4B-RL-Math}} & \multirow{4}{*}{Base} & AMC2023 & 46.1 & 59.2 & 70.6 & 79.2 & 85.2 & 89.6 & 93.2 & 96.1 & 98.1 & 99.5 & 100.0 \\
 &  &\cellcolor{zebraTint} AIME2024 &\cellcolor{zebraTint} 13.7 &\cellcolor{zebraTint} 19.3 &\cellcolor{zebraTint} 25.0 &\cellcolor{zebraTint} 31.0 &\cellcolor{zebraTint} 37.8 &\cellcolor{zebraTint} 45.0 &\cellcolor{zebraTint} 51.3 &\cellcolor{zebraTint} 56.4 &\cellcolor{zebraTint} 61.7 &\cellcolor{zebraTint} 68.2 &\cellcolor{zebraTint} 76.7 \\
 &  & AIME2025 & 10.1 & 14.1 & 18.0 & 22.3 & 27.7 & 34.3 & 41.6 & 48.5 & 55.7 & 63.2 & 70.0 \\
 &  &\cellcolor{zebraTint} AIME2026 &\cellcolor{zebraTint} 9.1 &\cellcolor{zebraTint} 13.3 &\cellcolor{zebraTint} 17.3 &\cellcolor{zebraTint} 21.1 &\cellcolor{zebraTint} 25.8 &\cellcolor{zebraTint} 31.1 &\cellcolor{zebraTint} 36.5 &\cellcolor{zebraTint} 42.3 &\cellcolor{zebraTint} 49.9 &\cellcolor{zebraTint} 59.8 &\cellcolor{zebraTint} 70.0 \\
\cmidrule(l){2-14}
 & \multirow{4}{*}{OPD} & AMC2023 & 81.8 & 88.6 & 92.5 & 94.6 & 95.5 & 96.1 & 96.8 & 97.3 & 97.5 & 97.5 & 97.5 \\
 &  &\cellcolor{zebraTint} AIME2024 & \cellcolor{zebraTint} 40.1 & \cellcolor{zebraTint} 48.9 & \cellcolor{zebraTint} 56.2 & \cellcolor{zebraTint} 61.7 & \cellcolor{zebraTint} 65.7 & \cellcolor{zebraTint} 68.9 & \cellcolor{zebraTint} 72.3 & \cellcolor{zebraTint} 76.1 & \cellcolor{zebraTint} 80.1 & \cellcolor{zebraTint} 82.8 & \cellcolor{zebraTint} 83.3 \\
 &  & AIME2025 & 33.3 & 40.4 & 47.3 & 53.2 & 58.1 & 62.4 & 66.5 & 70.1 & 72.3 & 73.2 & 73.3 \\
 &  &\cellcolor{zebraTint} AIME2026 & \cellcolor{zebraTint} 34.4 & \cellcolor{zebraTint} 43.0 & \cellcolor{zebraTint} 50.0 & \cellcolor{zebraTint} 55.5 & \cellcolor{zebraTint} 59.5 & \cellcolor{zebraTint} 62.3 & \cellcolor{zebraTint} 64.9 & \cellcolor{zebraTint} 67.9 & \cellcolor{zebraTint} 71.5 & \cellcolor{zebraTint} 75.6 & \cellcolor{zebraTint} 80.0 \\
\cmidrule(l){2-14}
 & \multirow{4}{*}{Teacher} & AMC2023 & 94.7 & 96.3 & 97.0 & 97.4 & 97.5 & 97.5 & 97.5 & 97.5 & 97.5 & 97.5 & 97.5 \\
 &  &\cellcolor{zebraTint} AIME2024 &\cellcolor{zebraTint} 61.2 &\cellcolor{zebraTint} 70.8 &\cellcolor{zebraTint} 77.1 &\cellcolor{zebraTint} 80.2 &\cellcolor{zebraTint} 82.0 &\cellcolor{zebraTint} 83.3 &\cellcolor{zebraTint} 84.4 &\cellcolor{zebraTint} 85.4 &\cellcolor{zebraTint} 86.7 &\cellcolor{zebraTint} 88.2 &\cellcolor{zebraTint} 90.0 \\
 &  & AIME2025 & 57.8 & 67.2 & 73.0 & 76.4 & 78.5 & 80.0 & 81.2 & 82.7 & 84.8 & 87.9 & 90.0 \\
 &  &\cellcolor{zebraTint} AIME2026 &\cellcolor{zebraTint} 57.8 &\cellcolor{zebraTint} 64.6 &\cellcolor{zebraTint} 70.2 &\cellcolor{zebraTint} 74.3 &\cellcolor{zebraTint} 76.8 &\cellcolor{zebraTint} 78.8 &\cellcolor{zebraTint} 80.7 &\cellcolor{zebraTint} 82.5 &\cellcolor{zebraTint} 84.7 &\cellcolor{zebraTint} 87.5 &\cellcolor{zebraTint} 90.0 \\
\bottomrule[1.6pt]
\end{tabular}
\end{table}
\begin{table}[!htbp]
\centering

\caption{Code pass@$K$ (\%) with Dataset~2 (APPS). Rows compare Base, OPD, and Teacher for both student-teacher settings. Columns give $K$; dashes mark entries not evaluated for the corresponding checkpoint and benchmark.}
\label{tab:additional-code-results}
\vspace{-8.5pt}
\fontsize{7.245}{8.05}\selectfont
\setlength{\tabcolsep}{(\textwidth-319.0pt)/26}
\renewcommand{\arraystretch}{0.88}
\setlength{\aboverulesep}{1pt}
\setlength{\belowrulesep}{1pt}
\begin{tabular}{@{}S{75.0pt}|S{35.9pt}S{56.9pt}*{8}{N{12.7pt}}*{3}{N{16.4pt}}@{}}
\toprule[1.6pt]
\textbf{Setting} & \textbf{Role} & \textbf{Dataset} & \textbf{1} & \textbf{2} & \textbf{4} & \textbf{8} & \textbf{16} & \textbf{32} & \textbf{64} & \textbf{128} & \textbf{256} & \textbf{512} & \textbf{1024} \\
\midrule
\multirow{12}{*}{\shortstack[l]{Qwen3-1.7B\,$\leftarrow$\\Qwen3-8B}} & \multirow{4}{*}{Base} & HumanEval+ & 56.2 & 60.3 & 63.6 & 66.6 & 69.2 & 71.1 & 72.3 & 72.9 & 73.4 & 73.8 & \textemdash \\
 &  &\cellcolor{zebraTint} MBPP+ &\cellcolor{zebraTint} 48.6 &\cellcolor{zebraTint} 51.4 &\cellcolor{zebraTint} 53.5 &\cellcolor{zebraTint} 55.3 &\cellcolor{zebraTint} 56.9 &\cellcolor{zebraTint} 58.1 &\cellcolor{zebraTint} 59.1 &\cellcolor{zebraTint} 59.9 &\cellcolor{zebraTint} 60.6 &\cellcolor{zebraTint} 61.4 &\cellcolor{zebraTint} \textemdash \\
 &  & LiveCodeBench v5 & 26.0 & 29.8 & 33.3 & 36.2 & 38.8 & 41.1 & 43.1 & 44.8 & \textemdash & \textemdash & \textemdash \\
 &  &\cellcolor{zebraTint} LiveCodeBench v6 &\cellcolor{zebraTint} 15.5 &\cellcolor{zebraTint} 17.7 &\cellcolor{zebraTint} 19.6 &\cellcolor{zebraTint} 21.4 &\cellcolor{zebraTint} 23.2 &\cellcolor{zebraTint} 25.1 &\cellcolor{zebraTint} 27.1 &\cellcolor{zebraTint} 29.1 &\cellcolor{zebraTint} \textemdash &\cellcolor{zebraTint} \textemdash &\cellcolor{zebraTint} \textemdash \\
\cmidrule(l){2-14}
 & \multirow{4}{*}{OPD} & HumanEval+ & 57.8 & 61.6 & 65.0 & 68.3 & 71.3 & 73.7 & 75.7 & 77.4 & 78.6 & 79.3 & \textemdash \\
 &  &\cellcolor{zebraTint} MBPP+ & \cellcolor{zebraTint} 48.8 & \cellcolor{zebraTint} 52.4 & \cellcolor{zebraTint} 55.4 & \cellcolor{zebraTint} 57.6 & \cellcolor{zebraTint} 59.5 & \cellcolor{zebraTint} 61.1 & \cellcolor{zebraTint} 62.3 & \cellcolor{zebraTint} 63.2 & \cellcolor{zebraTint} 64.0 & \cellcolor{zebraTint} 65.1 & \cellcolor{zebraTint} \textemdash \\
 &  & LiveCodeBench v5 & 27.7 & 32.3 & 36.1 & 39.4 & 42.8 & 46.2 & 49.6 & 52.4 & \textemdash & \textemdash & \textemdash \\
 &  &\cellcolor{zebraTint} LiveCodeBench v6 & \cellcolor{zebraTint} 17.7 & \cellcolor{zebraTint} 20.2 & \cellcolor{zebraTint} 22.7 & \cellcolor{zebraTint} 25.0 & \cellcolor{zebraTint} 27.0 & \cellcolor{zebraTint} 28.8 & \cellcolor{zebraTint} 30.7 & \cellcolor{zebraTint} 32.6 & \cellcolor{zebraTint} \textemdash & \cellcolor{zebraTint} \textemdash & \cellcolor{zebraTint} \textemdash \\
\cmidrule(l){2-14}
 & \multirow{4}{*}{Teacher} & HumanEval+ & 78.3 & 80.0 & 81.3 & 82.4 & 83.6 & 84.8 & 85.8 & 86.6 & 87.4 & 87.8 & \textemdash \\
 &  &\cellcolor{zebraTint} MBPP+ &\cellcolor{zebraTint} 68.0 &\cellcolor{zebraTint} 69.7 &\cellcolor{zebraTint} 71.2 &\cellcolor{zebraTint} 72.6 &\cellcolor{zebraTint} 73.8 &\cellcolor{zebraTint} 74.7 &\cellcolor{zebraTint} 75.3 &\cellcolor{zebraTint} 75.6 &\cellcolor{zebraTint} 75.8 &\cellcolor{zebraTint} 75.9 &\cellcolor{zebraTint} \textemdash \\
 &  & LiveCodeBench v5 & 43.2 & 48.4 & 52.9 & 56.6 & 59.8 & 62.6 & 64.9 & 67.0 & \textemdash & \textemdash & \textemdash \\
 &  &\cellcolor{zebraTint} LiveCodeBench v6 &\cellcolor{zebraTint} 25.8 &\cellcolor{zebraTint} 28.6 &\cellcolor{zebraTint} 31.2 &\cellcolor{zebraTint} 33.2 &\cellcolor{zebraTint} 34.8 &\cellcolor{zebraTint} 36.3 &\cellcolor{zebraTint} 38.0 &\cellcolor{zebraTint} 39.4 &\cellcolor{zebraTint} \textemdash &\cellcolor{zebraTint} \textemdash &\cellcolor{zebraTint} \textemdash \\
\midrule
\multirow{12}{*}{\shortstack[l]{Qwen3-1.7B\,$\leftarrow$\\Qwen3-4B-RL-Code}} & \multirow{4}{*}{Base} & HumanEval+ & 56.2 & 60.3 & 63.6 & 66.6 & 69.2 & 71.1 & 72.3 & 72.9 & 73.4 & 73.8 & \textemdash \\
 &  &\cellcolor{zebraTint} MBPP+ &\cellcolor{zebraTint} 48.6 &\cellcolor{zebraTint} 51.4 &\cellcolor{zebraTint} 53.5 &\cellcolor{zebraTint} 55.3 &\cellcolor{zebraTint} 56.9 &\cellcolor{zebraTint} 58.1 &\cellcolor{zebraTint} 59.1 &\cellcolor{zebraTint} 59.9 &\cellcolor{zebraTint} 60.6 &\cellcolor{zebraTint} 61.4 &\cellcolor{zebraTint} \textemdash \\
 &  & LiveCodeBench v5 & 26.0 & 29.8 & 33.3 & 36.2 & 38.8 & 41.1 & 43.1 & 44.8 & \textemdash & \textemdash & \textemdash \\
 &  &\cellcolor{zebraTint} LiveCodeBench v6 &\cellcolor{zebraTint} 15.5 &\cellcolor{zebraTint} 17.7 &\cellcolor{zebraTint} 19.6 &\cellcolor{zebraTint} 21.4 &\cellcolor{zebraTint} 23.2 &\cellcolor{zebraTint} 25.1 &\cellcolor{zebraTint} 27.1 &\cellcolor{zebraTint} 29.1 &\cellcolor{zebraTint} \textemdash &\cellcolor{zebraTint} \textemdash &\cellcolor{zebraTint} \textemdash \\
\cmidrule(l){2-14}
 & \multirow{4}{*}{OPD} & HumanEval+ & 62.3 & 70.6 & 77.4 & 82.3 & 85.4 & 87.3 & 88.6 & 89.5 & 90.2 & 90.9 & \textemdash \\
 &  &\cellcolor{zebraTint} MBPP+ & \cellcolor{zebraTint} 49.5 & \cellcolor{zebraTint} 54.9 & \cellcolor{zebraTint} 59.4 & \cellcolor{zebraTint} 63.2 & \cellcolor{zebraTint} 66.2 & \cellcolor{zebraTint} 68.8 & \cellcolor{zebraTint} 71.3 & \cellcolor{zebraTint} 73.7 & \cellcolor{zebraTint} 75.9 & \cellcolor{zebraTint} 77.5 & \cellcolor{zebraTint} \textemdash \\
 &  & LiveCodeBench v5 & 34.5 & 41.6 & 47.7 & 52.9 & 57.4 & 61.1 & 64.2 & 66.9 & \textemdash & \textemdash & \textemdash \\
 &  &\cellcolor{zebraTint} LiveCodeBench v6 & \cellcolor{zebraTint} 21.3 & \cellcolor{zebraTint} 24.8 & \cellcolor{zebraTint} 28.1 & \cellcolor{zebraTint} 31.1 & \cellcolor{zebraTint} 33.6 & \cellcolor{zebraTint} 35.6 & \cellcolor{zebraTint} 37.3 & \cellcolor{zebraTint} 39.4 & \cellcolor{zebraTint} \textemdash & \cellcolor{zebraTint} \textemdash & \cellcolor{zebraTint} \textemdash \\
\cmidrule(l){2-14}
 & \multirow{4}{*}{Teacher} & HumanEval+ & 79.6 & 82.7 & 85.1 & 87.2 & 88.9 & 90.1 & 90.9 & 91.6 & 92.0 & 92.1 & \textemdash \\
 &  &\cellcolor{zebraTint} MBPP+ &\cellcolor{zebraTint} 65.5 &\cellcolor{zebraTint} 68.5 &\cellcolor{zebraTint} 70.9 &\cellcolor{zebraTint} 72.7 &\cellcolor{zebraTint} 74.2 &\cellcolor{zebraTint} 75.3 &\cellcolor{zebraTint} 76.0 &\cellcolor{zebraTint} 76.5 &\cellcolor{zebraTint} 77.0 &\cellcolor{zebraTint} 77.5 &\cellcolor{zebraTint} \textemdash \\
 &  & LiveCodeBench v5 & 53.3 & 61.5 & 67.6 & 71.8 & 74.7 & 76.8 & 78.6 & 80.1 & \textemdash & \textemdash & \textemdash \\
 &  &\cellcolor{zebraTint} LiveCodeBench v6 &\cellcolor{zebraTint} 32.1 &\cellcolor{zebraTint} 36.5 &\cellcolor{zebraTint} 39.7 &\cellcolor{zebraTint} 42.1 &\cellcolor{zebraTint} 44.5 &\cellcolor{zebraTint} 46.7 &\cellcolor{zebraTint} 48.8 &\cellcolor{zebraTint} 50.9 &\cellcolor{zebraTint} \textemdash &\cellcolor{zebraTint} \textemdash &\cellcolor{zebraTint} \textemdash \\
\bottomrule[1.6pt]
\end{tabular}
\end{table}
\begin{table}[!htbp]
\centering

\caption{Fact pass@$K$ (\%) with Dataset~2 (TriviaQA). Rows compare Base, OPD, and Teacher for both student-teacher settings across all four benchmarks. Columns give the sampling budget $K$.}
\label{tab:additional-fact-results}
\vspace{-8.5pt}
\fontsize{7.245}{8.05}\selectfont
\setlength{\tabcolsep}{(\textwidth-319.0pt)/26}
\renewcommand{\arraystretch}{0.88}
\setlength{\aboverulesep}{1pt}
\setlength{\belowrulesep}{1pt}
\begin{tabular}{@{}S{75.0pt}|S{35.9pt}S{56.9pt}*{8}{N{12.7pt}}*{3}{N{16.4pt}}@{}}
\toprule[1.6pt]
\textbf{Setting} & \textbf{Role} & \textbf{Dataset} & \textbf{1} & \textbf{2} & \textbf{4} & \textbf{8} & \textbf{16} & \textbf{32} & \textbf{64} & \textbf{128} & \textbf{256} & \textbf{512} & \textbf{1024} \\
\midrule
\multirow{12}{*}{\shortstack[l]{Qwen3-1.7B\,$\leftarrow$\\Qwen3-8B-RL-RAR}} & \multirow{4}{*}{Base} & HotpotQA & 5.6 & 7.5 & 9.6 & 11.8 & 13.9 & 15.8 & 17.8 & 19.7 & 21.5 & 23.4 & 25.0 \\
 &  &\cellcolor{zebraTint} TriviaQA &\cellcolor{zebraTint} 21.1 &\cellcolor{zebraTint} 24.8 &\cellcolor{zebraTint} 28.2 &\cellcolor{zebraTint} 31.4 &\cellcolor{zebraTint} 34.7 &\cellcolor{zebraTint} 38.1 &\cellcolor{zebraTint} 41.4 &\cellcolor{zebraTint} 44.5 &\cellcolor{zebraTint} 47.1 &\cellcolor{zebraTint} 49.2 &\cellcolor{zebraTint} 51.0 \\
 &  & PopQA & 8.3 & 9.9 & 11.5 & 13.3 & 15.0 & 16.8 & 18.6 & 20.3 & 22.1 & 23.6 & 24.5 \\
 &  &\cellcolor{zebraTint} 2WikiMultiHopQA &\cellcolor{zebraTint} 3.0 &\cellcolor{zebraTint} 4.4 &\cellcolor{zebraTint} 5.9 &\cellcolor{zebraTint} 7.3 &\cellcolor{zebraTint} 8.7 &\cellcolor{zebraTint} 10.0 &\cellcolor{zebraTint} 11.2 &\cellcolor{zebraTint} 12.3 &\cellcolor{zebraTint} 13.3 &\cellcolor{zebraTint} 14.5 &\cellcolor{zebraTint} 16.0 \\
\cmidrule(l){2-14}
 & \multirow{4}{*}{OPD} & HotpotQA & 6.0 & 8.3 & 10.9 & 13.6 & 16.4 & 18.9 & 21.4 & 23.9 & 26.4 & 28.7 & 31.0 \\
 &  &\cellcolor{zebraTint} TriviaQA & \cellcolor{zebraTint} 24.4 & \cellcolor{zebraTint} 29.5 & \cellcolor{zebraTint} 34.2 & \cellcolor{zebraTint} 38.1 & \cellcolor{zebraTint} 41.3 & \cellcolor{zebraTint} 44.3 & \cellcolor{zebraTint} 47.3 & \cellcolor{zebraTint} 50.7 & \cellcolor{zebraTint} 54.3 & \cellcolor{zebraTint} 57.9 & \cellcolor{zebraTint} 61.0 \\
 &  & PopQA & 9.5 & 11.5 & 13.5 & 15.8 & 18.4 & 21.2 & 24.1 & 27.2 & 30.3 & 33.0 & 36.0 \\
 &  &\cellcolor{zebraTint} 2WikiMultiHopQA & \cellcolor{zebraTint} 4.5 & \cellcolor{zebraTint} 6.4 & \cellcolor{zebraTint} 8.3 & \cellcolor{zebraTint} 10.2 & \cellcolor{zebraTint} 12.3 & \cellcolor{zebraTint} 14.6 & \cellcolor{zebraTint} 17.1 & \cellcolor{zebraTint} 19.6 & \cellcolor{zebraTint} 22.1 & \cellcolor{zebraTint} 24.6 & \cellcolor{zebraTint} 27.0 \\
\cmidrule(l){2-14}
 & \multirow{4}{*}{Teacher} & HotpotQA & 10.5 & 13.1 & 16.3 & 19.9 & 23.4 & 26.7 & 29.9 & 33.0 & 36.1 & 38.9 & 41.0 \\
 &  &\cellcolor{zebraTint} TriviaQA &\cellcolor{zebraTint} 41.9 &\cellcolor{zebraTint} 47.6 &\cellcolor{zebraTint} 52.3 &\cellcolor{zebraTint} 56.4 &\cellcolor{zebraTint} 59.9 &\cellcolor{zebraTint} 63.2 &\cellcolor{zebraTint} 66.2 &\cellcolor{zebraTint} 68.9 &\cellcolor{zebraTint} 71.2 &\cellcolor{zebraTint} 73.2 &\cellcolor{zebraTint} 75.0 \\
 &  & PopQA & 14.1 & 17.3 & 20.5 & 23.3 & 25.9 & 28.4 & 30.8 & 33.1 & 35.4 & 37.7 & 40.5 \\
 &  &\cellcolor{zebraTint} 2WikiMultiHopQA &\cellcolor{zebraTint} 5.0 &\cellcolor{zebraTint} 7.0 &\cellcolor{zebraTint} 9.3 &\cellcolor{zebraTint} 11.9 &\cellcolor{zebraTint} 14.3 &\cellcolor{zebraTint} 16.7 &\cellcolor{zebraTint} 19.3 &\cellcolor{zebraTint} 22.0 &\cellcolor{zebraTint} 24.7 &\cellcolor{zebraTint} 27.7 &\cellcolor{zebraTint} 30.5 \\
\midrule
\multirow{12}{*}{\shortstack[l]{Qwen3-1.7B\,$\leftarrow$\\SESA-8B-search}} & \multirow{4}{*}{Base} & HotpotQA & 5.6 & 7.5 & 9.6 & 11.8 & 13.9 & 15.8 & 17.8 & 19.7 & 21.5 & 23.4 & 25.0 \\
 &  &\cellcolor{zebraTint} TriviaQA &\cellcolor{zebraTint} 21.1 &\cellcolor{zebraTint} 24.8 &\cellcolor{zebraTint} 28.2 &\cellcolor{zebraTint} 31.4 &\cellcolor{zebraTint} 34.7 &\cellcolor{zebraTint} 38.1 &\cellcolor{zebraTint} 41.4 &\cellcolor{zebraTint} 44.5 &\cellcolor{zebraTint} 47.1 &\cellcolor{zebraTint} 49.2 &\cellcolor{zebraTint} 51.0 \\
 &  & PopQA & 8.3 & 9.9 & 11.5 & 13.3 & 15.0 & 16.8 & 18.6 & 20.3 & 22.1 & 23.6 & 24.5 \\
 &  &\cellcolor{zebraTint} 2WikiMultiHopQA &\cellcolor{zebraTint} 3.0 &\cellcolor{zebraTint} 4.4 &\cellcolor{zebraTint} 5.9 &\cellcolor{zebraTint} 7.3 &\cellcolor{zebraTint} 8.7 &\cellcolor{zebraTint} 10.0 &\cellcolor{zebraTint} 11.2 &\cellcolor{zebraTint} 12.3 &\cellcolor{zebraTint} 13.3 &\cellcolor{zebraTint} 14.5 &\cellcolor{zebraTint} 16.0 \\
\cmidrule(l){2-14}
 & \multirow{4}{*}{OPD} & HotpotQA & 5.7 & 7.9 & 10.5 & 13.0 & 15.4 & 17.8 & 20.3 & 22.8 & 25.1 & 27.6 & 30.0 \\
 &  &\cellcolor{zebraTint} TriviaQA & \cellcolor{zebraTint} 24.2 & \cellcolor{zebraTint} 29.2 & \cellcolor{zebraTint} 33.8 & \cellcolor{zebraTint} 38.1 & \cellcolor{zebraTint} 42.1 & \cellcolor{zebraTint} 45.6 & \cellcolor{zebraTint} 48.4 & \cellcolor{zebraTint} 50.9 & \cellcolor{zebraTint} 53.3 & \cellcolor{zebraTint} 55.9 & \cellcolor{zebraTint} 58.0 \\
 &  & PopQA & 10.9 & 12.8 & 14.5 & 16.5 & 19.1 & 22.2 & 25.6 & 28.9 & 31.9 & 34.2 & 36.0 \\
 &  &\cellcolor{zebraTint} 2WikiMultiHopQA & \cellcolor{zebraTint} 3.0 & \cellcolor{zebraTint} 4.7 & \cellcolor{zebraTint} 6.8 & \cellcolor{zebraTint} 8.9 & \cellcolor{zebraTint} 10.9 & \cellcolor{zebraTint} 12.8 & \cellcolor{zebraTint} 14.7 & \cellcolor{zebraTint} 17.0 & \cellcolor{zebraTint} 19.6 & \cellcolor{zebraTint} 22.3 & \cellcolor{zebraTint} 25.0 \\
\cmidrule(l){2-14}
 & \multirow{4}{*}{Teacher} & HotpotQA & 12.4 & 15.5 & 18.8 & 22.2 & 25.6 & 28.8 & 32.0 & 35.4 & 39.2 & 42.9 & 46.0 \\
 &  &\cellcolor{zebraTint} TriviaQA &\cellcolor{zebraTint} 43.4 &\cellcolor{zebraTint} 48.4 &\cellcolor{zebraTint} 52.7 &\cellcolor{zebraTint} 56.2 &\cellcolor{zebraTint} 58.9 &\cellcolor{zebraTint} 61.2 &\cellcolor{zebraTint} 63.6 &\cellcolor{zebraTint} 65.9 &\cellcolor{zebraTint} 67.8 &\cellcolor{zebraTint} 69.7 &\cellcolor{zebraTint} 72.0 \\
 &  & PopQA & 16.5 & 19.7 & 22.7 & 25.6 & 28.5 & 31.4 & 34.5 & 37.4 & 40.2 & 43.0 & 46.5 \\
 &  &\cellcolor{zebraTint} 2WikiMultiHopQA &\cellcolor{zebraTint} 5.7 &\cellcolor{zebraTint} 8.0 &\cellcolor{zebraTint} 10.6 &\cellcolor{zebraTint} 13.2 &\cellcolor{zebraTint} 15.6 &\cellcolor{zebraTint} 18.0 &\cellcolor{zebraTint} 20.6 &\cellcolor{zebraTint} 23.3 &\cellcolor{zebraTint} 25.9 &\cellcolor{zebraTint} 28.1 &\cellcolor{zebraTint} 30.0 \\
\bottomrule[1.6pt]
\end{tabular}
\end{table}

\section{Additional \texorpdfstring{avg@$K$}{avg@K} Curves}
\label{app:avg-at-k}

\begin{figure}[!htbp]
    \centering
    \includegraphics[width=0.95\textwidth]{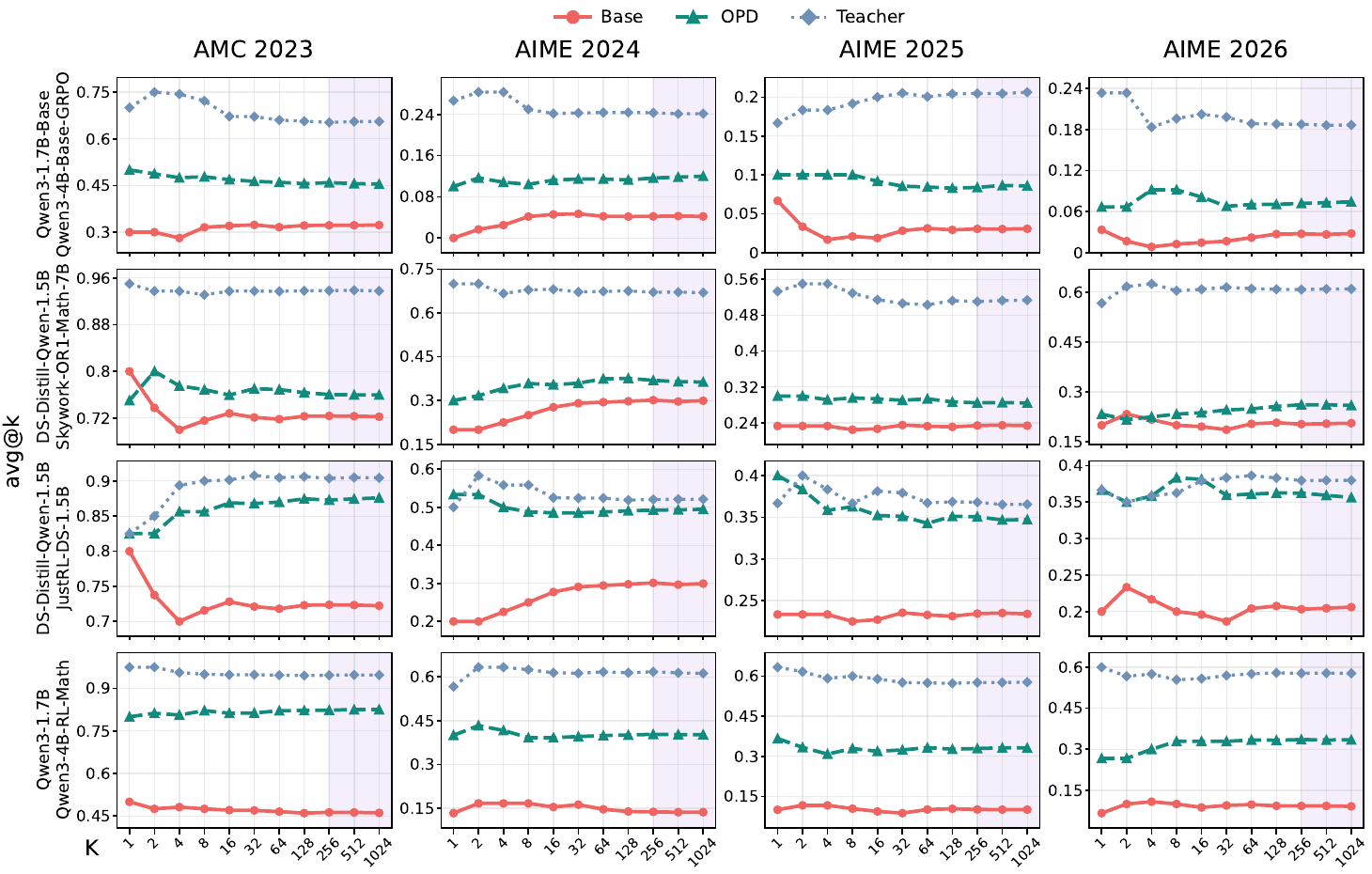}
    \vspace{-7.7pt}
    \captionsetup{width=0.95\textwidth}
    \caption{\textbf{Avg@$K$ in four Math settings.} Each point averages the correct-response fraction in the first $K$ samples per problem. Rows identify student--teacher settings; columns show AMC2023 and AIME2024--2026. Curves compare Base, OPD, and Teacher up to $K=1024$.}
    \label{fig:avgk-math}
\end{figure}

\begin{figure}[!htbp]
    \centering
    \includegraphics[width=0.95\textwidth]{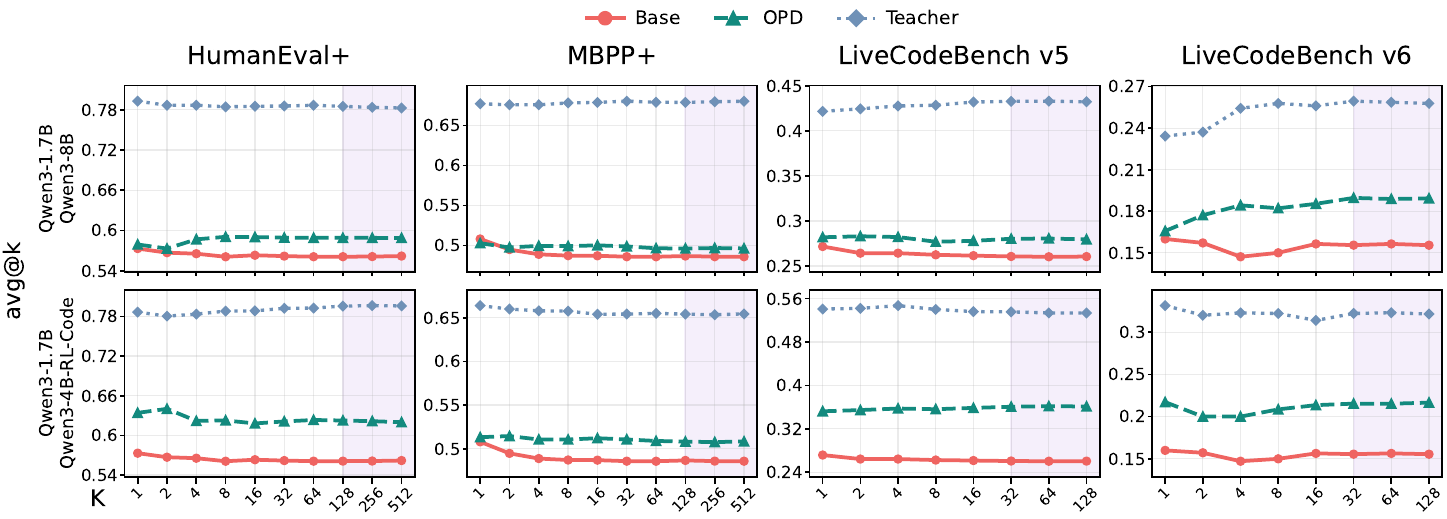}
    \vspace{-7.7pt}
    \captionsetup{width=0.95\textwidth}
    \caption{\textbf{Avg@$K$ in two Code settings.} Rows identify student--teacher settings, and columns show HumanEval+, MBPP+, and LiveCodeBench v5 and v6. Curves compare Base, OPD, and Teacher; the maximum $K$ is 512 for HumanEval+ and MBPP+ and 128 for LiveCodeBench.}
    \label{fig:avgk-code}
\end{figure}

\begin{figure}[!htbp]
    \centering
    \includegraphics[width=0.95\textwidth]{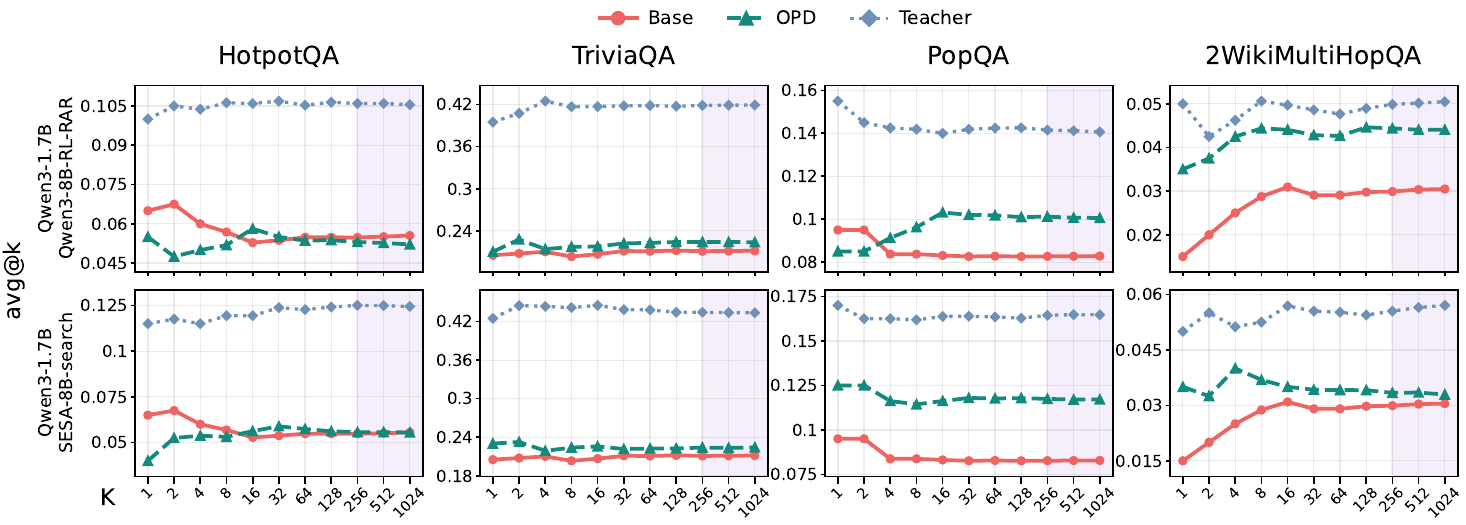}
    \vspace{-7.7pt}
    \captionsetup{width=0.95\textwidth}
    \caption{\textbf{Avg@$K$ in two Fact settings.} Rows use Qwen3-8B-RL-RAR and SESA-8B-search as teachers for Qwen3-1.7B; columns show HotpotQA, TriviaQA, PopQA, and 2WikiMultiHopQA. Curves compare Base, OPD, and Teacher up to $K=1024$. On HotpotQA with the RAR teacher, OPD has slightly lower avg@$1024$ than Base, although its pass@$1024$ is higher, so the two metrics order the two models differently at this sampling budget (Table~\ref{tab:fact-pass-numerical}).}
    \label{fig:avgk-fact}
\end{figure}

\begin{revisionblock}
Figs.~\ref{fig:avgk-math}, \ref{fig:avgk-code}, and~\ref{fig:avgk-fact} complement the Dataset~1 pass@$K$ curves in Figs.~\ref{fig:math-settings}, \ref{fig:code-settings}, and~\ref{fig:fact-settings}, using the same evaluation runs. For each problem, avg@$K$ is the fraction of correct responses among its first $K$ samples. We then average over problems. It measures sampled-response correctness, whereas pass@$K$ measures whether at least one of $K$ responses is correct. At each panel's maximum budget $K=n$, avg@$n$ equals the finite-sample pass@$1$ estimate computed from all $n$ responses. Avg@$1$ uses the first response per problem, whereas the pass@$1$ estimate uses all responses.
\end{revisionblock}

\section{Random-Trajectory Perplexity Analysis}
\label{app:random-trajectory-ppl}

To construct Fig.~\ref{fig:skywork-retained-alignment}, we use evaluation
runs for the pre-OPD base, OPD-trained, and teacher models in the DS-Distill-Qwen-1.5B $\leftarrow$ Skywork-OR1-Math-7B setting on AMC2023,
AIME2024, AIME2025, and AIME2026. With seed~0, we independently sample four problems from
each benchmark, yielding 16 problems in total. For every selected problem,
we sample 32 trajectories from each source model, resulting in
$16\times 32=512$ trajectories per source and 1,536 trajectories overall.

{Under the teacher scorer, trajectories generated by the OPD-trained model have lower perplexity than those generated by the pre-OPD base model. Under the pre-OPD base scorer, trajectories generated by the OPD-trained model have lower perplexity than those generated by either the pre-OPD base model or the teacher. The sampled OPD-generated trajectories thus combine lower teacher-scored perplexity with high likelihood under the pre-OPD base model across the sampled trajectories.}

\section{Numerical Results for Training Dynamics}

Tables~\ref{tab:training-dynamics-endpoints} and~\ref{tab:training-dynamics-code-numerical} give benchmark-level pass@$K$ (\%) for the checkpoints shown in Fig.~\ref{fig:opd-training-dynamics} and~\ref{fig:opd-training-dynamics-code}, respectively. Checkpoint blocks list Base and each evaluated OPD training step. Math uses 1,024 responses per problem; Code sampling depth varies by benchmark and checkpoint, so dashes mark budgets that were not evaluated for the corresponding checkpoint and benchmark.

Table~\ref{tab:training-dynamics-endpoints} expands the DS-Distill-Qwen-1.5B $\leftarrow$ Skywork-OR1-Math-7B training trajectory into pass@$K$ values for each benchmark and evaluated training step. At Step~260, pass@$1$ exceeds that of the pre-OPD base model on all four benchmarks, whereas pass@$1024$ becomes lower on AMC2023, AIME2025, and AIME2026, and matches the pre-OPD base model on AIME2024. By Step~80, pass@$1024$ has already dropped below the pre-OPD base model on three benchmarks, including a $10.0$~pp decrease on AIME2026; however, the subsequent checkpoints do not exhibit a monotonic decline. These results show that a pass@$1$ gain can coexist with lower pass@$1024$, and that the decrease can appear early in OPD training, well before the final training checkpoint.

\begin{table}[!htbp]
\centering
\caption{Math training dynamics: pass@$K$ (\%) for Base and Steps~20--260 on AMC2023 and AIME2024--2026, corresponding to Fig.~\ref{fig:opd-training-dynamics}. Columns give the sampling budget $K$.}
\label{tab:training-dynamics-endpoints}
\vspace{-8.5pt}
\fontsize{7.245}{8.05}\selectfont
\setlength{\tabcolsep}{(\textwidth-319.0pt)/26}
\renewcommand{\arraystretch}{0.88}
\setlength{\aboverulesep}{1pt}
\setlength{\belowrulesep}{1pt}
\begin{tabular}{@{}S{75.0pt}|S{35.9pt}S{56.9pt}*{8}{N{12.7pt}}*{3}{N{16.4pt}}@{}}
\toprule[1.6pt]
\textbf{Setting} & \textbf{Checkpoint} & \textbf{Dataset} & \textbf{1} & \textbf{2} & \textbf{4} & \textbf{8} & \textbf{16} & \textbf{32} & \textbf{64} & \textbf{128} & \textbf{256} & \textbf{512} & \textbf{1024} \\
\midrule
\multirow{24}{*}{\shortstack[l]{DS-Distill-Qwen-1.5B\,$\leftarrow$\\Skywork-OR1-Math-7B}} & \multirow{4}{*}{Base} & AMC2023 & 72.2 & 82.5 & 89.8 & 93.6 & 95.3 & 96.2 & 97.2 & 98.5 & 99.6 & 100.0 & 100.0 \\
 &  &\cellcolor{zebraTint} AIME2024 &\cellcolor{zebraTint} 29.9 &\cellcolor{zebraTint} 40.8 &\cellcolor{zebraTint} 51.7 &\cellcolor{zebraTint} 61.8 &\cellcolor{zebraTint} 70.0 &\cellcolor{zebraTint} 76.0 &\cellcolor{zebraTint} 80.0 &\cellcolor{zebraTint} 82.1 &\cellcolor{zebraTint} 83.7 &\cellcolor{zebraTint} 85.6 &\cellcolor{zebraTint} 86.7 \\
 &  & AIME2025 & 23.4 & 29.3 & 34.6 & 40.2 & 46.2 & 52.1 & 58.0 & 64.1 & 69.5 & 73.6 & 76.7 \\
 &  &\cellcolor{zebraTint} AIME2026 &\cellcolor{zebraTint} 20.6 &\cellcolor{zebraTint} 28.6 &\cellcolor{zebraTint} 36.9 &\cellcolor{zebraTint} 45.0 &\cellcolor{zebraTint} 52.9 &\cellcolor{zebraTint} 60.8 &\cellcolor{zebraTint} 68.2 &\cellcolor{zebraTint} 74.4 &\cellcolor{zebraTint} 79.2 &\cellcolor{zebraTint} 82.9 &\cellcolor{zebraTint} 86.7 \\
\cmidrule(l){2-14}
 & \multirow{4}{*}{Step 20} & AMC2023 & 73.3 & 83.3 & 90.3 & 94.0 & 95.5 & 96.3 & 97.3 & 98.6 & 99.7 & 100.0 & 100.0 \\
 &  &\cellcolor{zebraTint} AIME2024 &\cellcolor{zebraTint} 31.8 &\cellcolor{zebraTint} 43.0 &\cellcolor{zebraTint} 54.1 &\cellcolor{zebraTint} 63.9 &\cellcolor{zebraTint} 71.5 &\cellcolor{zebraTint} 76.7 &\cellcolor{zebraTint} 79.7 &\cellcolor{zebraTint} 81.3 &\cellcolor{zebraTint} 82.3 &\cellcolor{zebraTint} 83.1 &\cellcolor{zebraTint} 83.3 \\
 &  & AIME2025 & 24.7 & 30.5 & 36.0 & 41.7 & 47.8 & 53.4 & 58.6 & 63.7 & 69.1 & 73.6 & 76.7 \\
 &  &\cellcolor{zebraTint} AIME2026 &\cellcolor{zebraTint} 22.6 &\cellcolor{zebraTint} 31.1 &\cellcolor{zebraTint} 39.7 &\cellcolor{zebraTint} 47.9 &\cellcolor{zebraTint} 55.9 &\cellcolor{zebraTint} 63.6 &\cellcolor{zebraTint} 70.3 &\cellcolor{zebraTint} 75.6 &\cellcolor{zebraTint} 79.9 &\cellcolor{zebraTint} 83.3 &\cellcolor{zebraTint} 86.7 \\
\cmidrule(l){2-14}
 & \multirow{4}{*}{Step 80} & AMC2023 & 68.3 & 77.4 & 85.3 & 90.6 & 92.9 & 94.0 & 94.8 & 95.5 & 96.1 & 96.9 & 97.5 \\
 &  &\cellcolor{zebraTint} AIME2024 &\cellcolor{zebraTint} 32.1 &\cellcolor{zebraTint} 42.5 &\cellcolor{zebraTint} 52.8 &\cellcolor{zebraTint} 62.0 &\cellcolor{zebraTint} 69.5 &\cellcolor{zebraTint} 74.9 &\cellcolor{zebraTint} 78.5 &\cellcolor{zebraTint} 81.2 &\cellcolor{zebraTint} 83.9 &\cellcolor{zebraTint} 86.1 &\cellcolor{zebraTint} 86.7 \\
 &  & AIME2025 & 26.2 & 31.8 & 36.4 & 40.7 & 45.3 & 49.9 & 54.5 & 59.1 & 63.9 & 67.7 & 70.0 \\
 &  &\cellcolor{zebraTint} AIME2026 &\cellcolor{zebraTint} 23.7 &\cellcolor{zebraTint} 31.5 &\cellcolor{zebraTint} 38.8 &\cellcolor{zebraTint} 45.6 &\cellcolor{zebraTint} 53.2 &\cellcolor{zebraTint} 60.7 &\cellcolor{zebraTint} 66.3 &\cellcolor{zebraTint} 70.7 &\cellcolor{zebraTint} 74.4 &\cellcolor{zebraTint} 76.2 &\cellcolor{zebraTint} 76.7 \\
\cmidrule(l){2-14}
 & \multirow{4}{*}{Step 140} & AMC2023 & 72.9 & 82.4 & 89.9 & 93.7 & 94.9 & 95.1 & 95.2 & 95.3 & 95.6 & 96.2 & 97.5 \\
 &  &\cellcolor{zebraTint} AIME2024 &\cellcolor{zebraTint} 34.9 &\cellcolor{zebraTint} 45.5 &\cellcolor{zebraTint} 56.1 &\cellcolor{zebraTint} 65.1 &\cellcolor{zebraTint} 71.5 &\cellcolor{zebraTint} 75.3 &\cellcolor{zebraTint} 77.4 &\cellcolor{zebraTint} 78.8 &\cellcolor{zebraTint} 80.5 &\cellcolor{zebraTint} 82.5 &\cellcolor{zebraTint} 83.3 \\
 &  & AIME2025 & 27.6 & 32.8 & 37.2 & 41.5 & 46.0 & 50.4 & 54.6 & 58.9 & 63.0 & 66.6 & 70.0 \\
 &  &\cellcolor{zebraTint} AIME2026 &\cellcolor{zebraTint} 24.7 &\cellcolor{zebraTint} 33.1 &\cellcolor{zebraTint} 41.4 &\cellcolor{zebraTint} 49.5 &\cellcolor{zebraTint} 57.7 &\cellcolor{zebraTint} 64.4 &\cellcolor{zebraTint} 68.6 &\cellcolor{zebraTint} 71.8 &\cellcolor{zebraTint} 74.5 &\cellcolor{zebraTint} 76.7 &\cellcolor{zebraTint} 80.0 \\
\cmidrule(l){2-14}
 & \multirow{4}{*}{Step 200} & AMC2023 & 74.1 & 83.4 & 90.5 & 94.0 & 95.0 & 95.1 & 95.2 & 95.3 & 95.6 & 96.2 & 97.5 \\
 &  &\cellcolor{zebraTint} AIME2024 &\cellcolor{zebraTint} 35.3 &\cellcolor{zebraTint} 46.3 &\cellcolor{zebraTint} 57.2 &\cellcolor{zebraTint} 66.1 &\cellcolor{zebraTint} 71.6 &\cellcolor{zebraTint} 74.7 &\cellcolor{zebraTint} 76.7 &\cellcolor{zebraTint} 78.8 &\cellcolor{zebraTint} 81.2 &\cellcolor{zebraTint} 84.0 &\cellcolor{zebraTint} 86.7 \\
 &  & AIME2025 & 27.7 & 32.6 & 36.8 & 41.0 & 45.4 & 50.1 & 55.1 & 60.1 & 64.8 & 68.5 & 70.0 \\
 &  &\cellcolor{zebraTint} AIME2026 &\cellcolor{zebraTint} 24.0 &\cellcolor{zebraTint} 32.3 &\cellcolor{zebraTint} 40.5 &\cellcolor{zebraTint} 48.7 &\cellcolor{zebraTint} 57.0 &\cellcolor{zebraTint} 63.7 &\cellcolor{zebraTint} 68.0 &\cellcolor{zebraTint} 71.4 &\cellcolor{zebraTint} 75.0 &\cellcolor{zebraTint} 79.6 &\cellcolor{zebraTint} 86.7 \\
\cmidrule(l){2-14}
 & \multirow{4}{*}{Step 260} & AMC2023 & 76.0 & 85.2 & 91.6 & 94.3 & 95.1 & 95.4 & 95.8 & 96.4 & 97.1 & 97.5 & 97.5 \\
 &  &\cellcolor{zebraTint} AIME2024 &\cellcolor{zebraTint} 36.4 &\cellcolor{zebraTint} 47.2 &\cellcolor{zebraTint} 57.9 &\cellcolor{zebraTint} 66.6 &\cellcolor{zebraTint} 72.3 &\cellcolor{zebraTint} 75.6 &\cellcolor{zebraTint} 77.5 &\cellcolor{zebraTint} 78.8 &\cellcolor{zebraTint} 80.6 &\cellcolor{zebraTint} 83.1 &\cellcolor{zebraTint} 86.7 \\
 &  & AIME2025 & 28.5 & 33.4 & 37.7 & 42.0 & 46.4 & 50.8 & 55.1 & 60.0 & 64.7 & 67.9 & 70.0 \\
 &  &\cellcolor{zebraTint} AIME2026 &\cellcolor{zebraTint} 25.9 &\cellcolor{zebraTint} 34.5 &\cellcolor{zebraTint} 42.5 &\cellcolor{zebraTint} 50.4 &\cellcolor{zebraTint} 58.2 &\cellcolor{zebraTint} 64.3 &\cellcolor{zebraTint} 68.0 &\cellcolor{zebraTint} 71.4 &\cellcolor{zebraTint} 74.8 &\cellcolor{zebraTint} 77.5 &\cellcolor{zebraTint} 80.0 \\
\bottomrule[1.6pt]
\end{tabular}
\end{table}

Table~\ref{tab:training-dynamics-code-numerical} expands the Qwen3-1.7B $\leftarrow$ Qwen3-4B-RL-Code training trajectory into pass@$K$ values for HumanEval+, MBPP+, and LiveCodeBench v5 and v6. Already at Step~100, pass@$1$ and the largest evaluated budget for each benchmark exceed the corresponding Base values. The available sampling depth varies across benchmark and checkpoint rows, so not every budget is available.

\begin{table}[!htbp]
\centering
\caption{Code training dynamics: pass@$K$ (\%) for Base and Steps~100--500 on HumanEval+, MBPP+, and LiveCodeBench v5 and v6, corresponding to Fig.~\ref{fig:opd-training-dynamics-code}. Columns give $K$; dashes mark budgets that were not evaluated for the corresponding checkpoint and benchmark.}
\label{tab:training-dynamics-code-numerical}
\vspace{-8.5pt}
\fontsize{7.245}{8.05}\selectfont
\setlength{\tabcolsep}{(\textwidth-319.0pt)/26}
\renewcommand{\arraystretch}{0.88}
\setlength{\aboverulesep}{1pt}
\setlength{\belowrulesep}{1pt}
\begin{tabular}{@{}S{75.0pt}|S{35.9pt}S{56.9pt}*{8}{N{12.7pt}}*{3}{N{16.4pt}}@{}}
\toprule[1.6pt]
\textbf{Setting} & \textbf{Checkpoint} & \textbf{Dataset} & \textbf{1} & \textbf{2} & \textbf{4} & \textbf{8} & \textbf{16} & \textbf{32} & \textbf{64} & \textbf{128} & \textbf{256} & \textbf{512} & \textbf{1024} \\
\midrule
\multirow{24}{*}{\shortstack[l]{Qwen3-1.7B\,$\leftarrow$\\Qwen3-4B-RL-Code}} & \multirow{4}{*}{Base} & HumanEval+ & 56.2 & 60.3 & 63.6 & 66.6 & 69.2 & 71.1 & 72.3 & 72.9 & 73.4 & 73.8 & \textemdash \\
 &  &\cellcolor{zebraTint} MBPP+ &\cellcolor{zebraTint} 48.6 &\cellcolor{zebraTint} 51.4 &\cellcolor{zebraTint} 53.5 &\cellcolor{zebraTint} 55.3 &\cellcolor{zebraTint} 56.9 &\cellcolor{zebraTint} 58.1 &\cellcolor{zebraTint} 59.1 &\cellcolor{zebraTint} 59.9 &\cellcolor{zebraTint} 60.6 &\cellcolor{zebraTint} 61.4 &\cellcolor{zebraTint} \textemdash \\
 &  & LiveCodeBench v5 & 26.0 & 29.8 & 33.3 & 36.2 & 38.8 & 41.1 & 43.1 & 44.8 & \textemdash & \textemdash & \textemdash \\
 &  &\cellcolor{zebraTint} LiveCodeBench v6 &\cellcolor{zebraTint} 15.5 &\cellcolor{zebraTint} 17.7 &\cellcolor{zebraTint} 19.6 &\cellcolor{zebraTint} 21.4 &\cellcolor{zebraTint} 23.2 &\cellcolor{zebraTint} 25.1 &\cellcolor{zebraTint} 27.1 &\cellcolor{zebraTint} 29.1 &\cellcolor{zebraTint} \textemdash &\cellcolor{zebraTint} \textemdash &\cellcolor{zebraTint} \textemdash \\
\cmidrule(l){2-14}
 & \multirow{4}{*}{Step 100} & HumanEval+ & 59.5 & 67.0 & 73.4 & 78.9 & 83.2 & 85.9 & 87.5 & 88.5 & 89.6 & \textemdash & \textemdash \\
 &  &\cellcolor{zebraTint} MBPP+ &\cellcolor{zebraTint} 50.8 &\cellcolor{zebraTint} 56.0 &\cellcolor{zebraTint} 60.5 &\cellcolor{zebraTint} 64.4 &\cellcolor{zebraTint} 67.8 &\cellcolor{zebraTint} 70.8 &\cellcolor{zebraTint} 73.6 &\cellcolor{zebraTint} 76.1 &\cellcolor{zebraTint} 78.0 &\cellcolor{zebraTint} \textemdash &\cellcolor{zebraTint} \textemdash \\
 &  & LiveCodeBench v5 & 34.7 & 41.9 & 48.2 & 53.4 & 57.7 & 61.1 & 63.9 & 66.0 & \textemdash & \textemdash & \textemdash \\
 &  &\cellcolor{zebraTint} LiveCodeBench v6 &\cellcolor{zebraTint} 21.5 &\cellcolor{zebraTint} 25.2 &\cellcolor{zebraTint} 28.6 &\cellcolor{zebraTint} 31.6 &\cellcolor{zebraTint} 34.1 &\cellcolor{zebraTint} 36.4 &\cellcolor{zebraTint} 38.5 &\cellcolor{zebraTint} 40.6 &\cellcolor{zebraTint} \textemdash &\cellcolor{zebraTint} \textemdash &\cellcolor{zebraTint} \textemdash \\
\cmidrule(l){2-14}
 & \multirow{4}{*}{Step 200} & HumanEval+ & 61.1 & 68.9 & 75.7 & 81.3 & 85.1 & 87.3 & 88.7 & 89.8 & 90.9 & \textemdash & \textemdash \\
 &  &\cellcolor{zebraTint} MBPP+ &\cellcolor{zebraTint} 50.4 &\cellcolor{zebraTint} 55.7 &\cellcolor{zebraTint} 60.3 &\cellcolor{zebraTint} 64.3 &\cellcolor{zebraTint} 67.8 &\cellcolor{zebraTint} 70.8 &\cellcolor{zebraTint} 73.4 &\cellcolor{zebraTint} 75.4 &\cellcolor{zebraTint} 76.7 &\cellcolor{zebraTint} \textemdash &\cellcolor{zebraTint} \textemdash \\
 &  & LiveCodeBench v5 & 35.4 & 43.0 & 49.4 & 54.7 & 58.9 & 62.4 & 65.5 & 68.0 & \textemdash & \textemdash & \textemdash \\
 &  &\cellcolor{zebraTint} LiveCodeBench v6 &\cellcolor{zebraTint} 21.2 &\cellcolor{zebraTint} 25.4 &\cellcolor{zebraTint} 29.3 &\cellcolor{zebraTint} 32.4 &\cellcolor{zebraTint} 34.7 &\cellcolor{zebraTint} 36.5 &\cellcolor{zebraTint} 38.2 &\cellcolor{zebraTint} 39.4 &\cellcolor{zebraTint} \textemdash &\cellcolor{zebraTint} \textemdash &\cellcolor{zebraTint} \textemdash \\
\cmidrule(l){2-14}
 & \multirow{4}{*}{Step 300} & HumanEval+ & 61.7 & 69.9 & 76.7 & 82.0 & 85.6 & 88.0 & 89.8 & 91.0 & 91.5 & \textemdash & \textemdash \\
 &  &\cellcolor{zebraTint} MBPP+ &\cellcolor{zebraTint} 50.3 &\cellcolor{zebraTint} 56.0 &\cellcolor{zebraTint} 60.8 &\cellcolor{zebraTint} 64.8 &\cellcolor{zebraTint} 68.0 &\cellcolor{zebraTint} 70.8 &\cellcolor{zebraTint} 73.5 &\cellcolor{zebraTint} 75.9 &\cellcolor{zebraTint} 77.8 &\cellcolor{zebraTint} \textemdash &\cellcolor{zebraTint} \textemdash \\
 &  & LiveCodeBench v5 & 35.8 & 43.6 & 50.3 & 55.7 & 60.0 & 63.6 & 66.7 & 69.2 & \textemdash & \textemdash & \textemdash \\
 &  &\cellcolor{zebraTint} LiveCodeBench v6 &\cellcolor{zebraTint} 21.1 &\cellcolor{zebraTint} 25.6 &\cellcolor{zebraTint} 29.7 &\cellcolor{zebraTint} 32.9 &\cellcolor{zebraTint} 35.1 &\cellcolor{zebraTint} 36.7 &\cellcolor{zebraTint} 38.0 &\cellcolor{zebraTint} 38.9 &\cellcolor{zebraTint} \textemdash &\cellcolor{zebraTint} \textemdash &\cellcolor{zebraTint} \textemdash \\
\cmidrule(l){2-14}
 & \multirow{4}{*}{Step 400} & HumanEval+ & 62.0 & 70.5 & 77.3 & 82.5 & 86.1 & 88.4 & 89.9 & 90.9 & 91.6 & 92.1 & \textemdash \\
 &  &\cellcolor{zebraTint} MBPP+ &\cellcolor{zebraTint} 50.8 &\cellcolor{zebraTint} 56.5 &\cellcolor{zebraTint} 61.2 &\cellcolor{zebraTint} 65.1 &\cellcolor{zebraTint} 68.4 &\cellcolor{zebraTint} 71.3 &\cellcolor{zebraTint} 74.0 &\cellcolor{zebraTint} 76.5 &\cellcolor{zebraTint} 78.6 &\cellcolor{zebraTint} 80.2 &\cellcolor{zebraTint} \textemdash \\
 &  & LiveCodeBench v5 & 36.1 & 43.9 & 50.4 & 55.6 & 59.6 & 63.0 & 65.8 & 68.1 & \textemdash & \textemdash & \textemdash \\
 &  &\cellcolor{zebraTint} LiveCodeBench v6 &\cellcolor{zebraTint} 21.7 &\cellcolor{zebraTint} 26.0 &\cellcolor{zebraTint} 30.0 &\cellcolor{zebraTint} 33.2 &\cellcolor{zebraTint} 35.5 &\cellcolor{zebraTint} 37.2 &\cellcolor{zebraTint} 38.6 &\cellcolor{zebraTint} 40.0 &\cellcolor{zebraTint} \textemdash &\cellcolor{zebraTint} \textemdash &\cellcolor{zebraTint} \textemdash \\
\cmidrule(l){2-14}
 & \multirow{4}{*}{Step 500} & HumanEval+ & 62.6 & 71.5 & 78.3 & 83.3 & 86.8 & 89.3 & 90.7 & 91.8 & 92.7 & \textemdash & \textemdash \\
 &  &\cellcolor{zebraTint} MBPP+ &\cellcolor{zebraTint} 50.4 &\cellcolor{zebraTint} 56.1 &\cellcolor{zebraTint} 60.8 &\cellcolor{zebraTint} 64.7 &\cellcolor{zebraTint} 67.8 &\cellcolor{zebraTint} 70.4 &\cellcolor{zebraTint} 73.1 &\cellcolor{zebraTint} 75.9 &\cellcolor{zebraTint} 78.6 &\cellcolor{zebraTint} \textemdash &\cellcolor{zebraTint} \textemdash \\
 &  & LiveCodeBench v5 & 37.0 & 44.8 & 51.2 & 56.4 & 60.4 & 63.7 & 66.3 & 68.4 & \textemdash & \textemdash & \textemdash \\
 &  &\cellcolor{zebraTint} LiveCodeBench v6 &\cellcolor{zebraTint} 22.2 &\cellcolor{zebraTint} 26.6 &\cellcolor{zebraTint} 30.5 &\cellcolor{zebraTint} 33.6 &\cellcolor{zebraTint} 35.9 &\cellcolor{zebraTint} 37.6 &\cellcolor{zebraTint} 39.0 &\cellcolor{zebraTint} 40.0 &\cellcolor{zebraTint} \textemdash &\cellcolor{zebraTint} \textemdash &\cellcolor{zebraTint} \textemdash \\
\bottomrule[1.6pt]
\end{tabular}
\end{table}

\begin{revisionblock}
\section{Additional Teacher-Selection Analyses}
\label{app:teacher-selection-analyses}
\subsection{Selection Agreement with Confidence Intervals}
\label{app:selection-uncertainty}
Table~\ref{tab:selection-ci} reports selection agreement with 95\% CIs from 10,000 paired test-problem bootstrap replicates. We resample problems within each benchmark, share draws across budgets and methods, and recompute student pass@$K$ to assess the fixed teacher choices. Overall pools the original 177 comparisons. Student ties are non-hits. Random reports its 50\% expectation. These intervals quantify uncertainty from the evaluated problems rather than from retraining.
\end{revisionblock}
\begin{table}[!htbp]
\centering
\renewcommand{\arraystretch}{1}
\setlength{\aboverulesep}{1pt}
\setlength{\belowrulesep}{1pt}

\footnotesize
\setlength{\tabcolsep}{6pt}

\caption{Selection agreement (\%) [95\% CI] for the original teacher-selection comparisons.}
\label{tab:selection-ci}
\vspace{-8.5pt}
\begin{tabular}{@{}lccccc@{}}\toprule[1.01pt]
\textbf{Teacher pair} & \textbf{Random} & \textbf{Teacher pass@$1$} & \textbf{Teacher pass@$K$} & \textbf{UTAS@$K$} & \textbf{TAS@$K$} \\ \midrule[0.32pt]
Qwen-Math & 50.0 & \shortstack{90.9\\{}[81.8, 97.7]} & \shortstack{95.5\\{}[84.1, 100.0]} & \shortstack{90.9\\{}[81.8, 97.7]} & \shortstack{90.9\\{}[81.8, 97.7]} \\
Qwen-Code & 50.0 & \shortstack{72.2\\{}[69.4, 75.0]} & \shortstack{91.7\\{}[88.9, 94.4]} & \shortstack{72.2\\{}[69.4, 72.2]} & \shortstack{77.8\\{}[77.8, 80.6]} \\
DeepSeek-Math & 50.0 & \shortstack{3.8\\{}[0.0, 20.8]} & \shortstack{4.7\\{}[0.0, 21.7]} & \shortstack{96.2\\{}[75.5, 98.1]} & \shortstack{90.6\\{}[73.6, 94.3]} \\
Qwen-Fact & 50.0 & \shortstack{43.2\\{}[25.0, 63.6]} & \shortstack{63.6\\{}[38.6, 75.0]} & \shortstack{61.4\\{}[43.2, 84.1]} & \shortstack{72.7\\{}[54.5, 86.4]} \\
Overall & 50.0 & \shortstack{49.2\\{}[44.1, 57.1]} & \shortstack{59.6\\{}[52.3, 64.7]} & \shortstack{81.4\\{}[73.4, 85.9]} & \shortstack{83.6\\{}[75.7, 86.4]} \\
\bottomrule[1.01pt]\end{tabular}
\end{table}
\begin{revisionblock}
\subsection{Changes in Teacher Preference Across Budgets}
\label{app:budget-switches}
Table~\ref{tab:budget-switches} expands the five teacher-benchmark combinations where the preferred teacher changes with $K$. Teacher pass@$1$ and UTAS@$K$ have no budget-dependent weights and therefore make a fixed choice across budgets. TAS@$K$ and teacher pass@$K$ can change their choices. On TriviaQA, TAS@$K$ identifies the preference for teacher $A$ at $K=4$, earlier than teacher pass@$K$ at $K=8$. Across these 55 comparisons, TAS@$K$ achieves 39 correct selections and the lowest mean loss (0.378 pp). The methods differ in which switches they capture.

Teacher labels $A/B$ denote Qwen3-8B / RL-Math in Qwen-Math, Skywork / JustRL in DeepSeek-Math, and Binary-RAR / SESA in Qwen-Fact. These labels apply at every budget.
\end{revisionblock}
\begin{table}[!htbp]
\centering
\renewcommand{\arraystretch}{1}
\setlength{\aboverulesep}{1pt}
\setlength{\belowrulesep}{1pt}
\footnotesize\setlength{\tabcolsep}{4pt}

\caption{Teacher choices for the five combinations with changes in student ordering. Agreement counts correct selections, and mean loss (pp) averages all 11 budgets per benchmark.}
\label{tab:budget-switches}
\vspace{-8.5pt}
\begin{tabular}{@{}llrcccc@{}}\toprule[1.01pt]
\textbf{Teacher pair} & \textbf{Benchmark} & $\boldsymbol{K}$ & \shortstack{\textbf{Preferred}\\\textbf{teacher}} & \textbf{Teacher pass@$K$} & \textbf{UTAS@$K$} & \textbf{TAS@$K$} \\
\midrule[0.32pt]
Qwen-Math & AMC2023 & 1-64 & B & B & B & B \\
 &  & 128 & B & A & B & B \\
 &  & 256-1024 & A & A & B & B \\
 & & & Agreement & 10/11 & 8/11 & 8/11 \\
 & & & Mean loss (pp) & 0.012 & 0.527 & 0.527 \\
\midrule[0.32pt]
Qwen-Math & AIME2026 & 1-512 & B & B & B & B \\
 &  & 1024 & A & B & B & B \\
 & & & Agreement & 10/11 & 10/11 & 10/11 \\
 & & & Mean loss (pp) & 0.303 & 0.303 & 0.303 \\
\midrule[0.32pt]
DeepSeek-Math & AIME2024 & 1-256 & B & A & B & B \\
 &  & 512-1024 & A & A & B & B \\
 & & & Agreement & 2/11 & 9/11 & 9/11 \\
 & & & Mean loss (pp) & 4.965 & 0.436 & 0.436 \\
\midrule[0.32pt]
Qwen-Fact & HotpotQA & 1-8 & B & B & A & A \\
 &  & 16-128 & A & B & A & A \\
 &  & 256-512 & B & B & A & A \\
 &  & 1024 & B & B & A & B \\
 & & & Agreement & 7/11 & 4/11 & 5/11 \\
 & & & Mean loss (pp) & 0.130 & 0.308 & 0.217 \\
\midrule[0.32pt]
Qwen-Fact & TriviaQA & 1 & B & B & B & B \\
 &  & 2 & A & B & B & B \\
 &  & 4 & A & B & B & A \\
 &  & 8-128 & A & A & B & A \\
 &  & 256-1024 & A & A & B & B \\
 & & & Agreement & 9/11 & 1/11 & 7/11 \\
 & & & Mean loss (pp) & 0.122 & 1.247 & 0.406 \\
\midrule[0.32pt]
\multicolumn{4}{@{}l}{\textbf{Overall agreement}} & \textbf{38/55} & \textbf{32/55} & \textbf{39/55} \\
\multicolumn{4}{@{}l}{\textbf{Overall mean loss (pp)}} & \textbf{1.106} & \textbf{0.564} & \textbf{0.378} \\
\bottomrule[1.01pt]\end{tabular}
\end{table}
\begin{revisionblock}
\subsection{Cross-Benchmark Selection Protocols}
\label{app:cross-benchmark-protocol}
The leave-one-benchmark-out analysis in Section~\ref{sec:cross-benchmark-selection} uses four benchmarks per pair. For each held-out benchmark and target $K$, we equally average TAS@$K$ scores across the other three benchmarks and select by the sign of this average. Math uses AMC2023 and AIME2024/2025/2026; Code uses HumanEval+, MBPP+, and LiveCodeBench v5/v6; Fact uses HotpotQA, TriviaQA, PopQA, and 2WikiMultiHopQA. Math and Fact use $K\in\{1,2,4,\ldots,1024\}$, while Code uses the common grid through 128. Excluding MATH500 and the higher-budget EvalPlus points gives 164 comparisons, rather than the 177 comparisons in the target-benchmark analysis.

For the additional test benchmarks, the calibration set comprises AMC2023 and AIME2024/2025/2026 (130 problems). We equally average benchmark-level TAS@$K$ scores to choose between Skywork and JustRL at each budget. We then evaluate the two DAPO-trained DeepSeek students on BRUMO2025, CMIMC2025, and HMMT-Feb2025, which contain 100 test problems in total and are disjoint from the calibration set. The 33 comparisons use 11 budgets per test benchmark. CIs in Table~\ref{tab:cross-benchmark-selection} use the paired test-problem bootstrap described in Appendix~\ref{app:selection-uncertainty}, while retaining the teacher choices computed from the calibration benchmarks.

\subsection{Repeated Evaluation}
\label{app:repeated-evaluation}
We evaluate the DeepSeek-Math pre-OPD base model and both OPD-trained models on AIME2024 with sampling seeds 0, 1, 2, and 3. Table~\ref{tab:repeated-passk} gives each evaluation, the mean, and the sample standard deviation. The checkpoints are fixed, so this tests evaluation sampling rather than training-seed variation. Both OPD-trained models improve pass@$1$ in all four evaluations, while both fall below the base model at $K=512$. Table~\ref{tab:repeated-selection} evaluates the TAS@$K$ teacher choices against each evaluation's student ordering: agreement is 81.8\% with seed 0 and 90.9\% with seeds 1, 2, and 3.
\end{revisionblock}
\begin{table}[!htbp]
\centering
\renewcommand{\arraystretch}{1}
\setlength{\aboverulesep}{1pt}
\setlength{\belowrulesep}{1pt}

\scriptsize
\setlength{\tabcolsep}{2.25pt}

\caption{AIME2024 pass@$K$ (\%) across four evaluation configurations. SD is in percentage points. Skywork and JustRL label their corresponding OPD-trained models.}
\label{tab:repeated-passk}
\vspace{-8.5pt}
\begin{tabular}{@{}llrrrrrrrrrrr@{}}\toprule[1.01pt]
\textbf{Model} & & \multicolumn{11}{c}{\textbf{Sampling budget $K$}} \\
\cmidrule(lr){3-13}
& \textbf{Seed / stat.} & \textbf{1} & \textbf{2} & \textbf{4} & \textbf{8} & \textbf{16} & \textbf{32} & \textbf{64} & \textbf{128} & \textbf{256} & \textbf{512} & \textbf{1024} \\ \midrule[0.32pt]
\multirow{6}{*}{Base} & 0 & 29.93 & 40.75 & 51.74 & 61.82 & 70.01 & 76.05 & 79.97 & 82.06 & 83.74 & 85.63 & 86.67 \\
 & 1 & 29.75 & 40.56 & 51.47 & 61.54 & 69.88 & 76.03 & 80.00 & 82.25 & 84.00 & 85.73 & 86.67 \\
 & 2 & 29.75 & 40.59 & 51.52 & 61.51 & 69.79 & 76.09 & 80.31 & 82.50 & 83.84 & 84.99 & 86.67 \\
 & 3 & 29.29 & 40.19 & 51.22 & 61.30 & 69.57 & 75.77 & 79.78 & 81.78 & 83.39 & 85.42 & 86.67 \\
\cmidrule(lr){2-13}
 & \textbf{Mean} & 29.68 & 40.53 & 51.49 & 61.54 & 69.81 & 75.99 & 80.01 & 82.15 & 83.74 & 85.44 & 86.67 \\
 & SD & 0.27 & 0.24 & 0.22 & 0.21 & 0.19 & 0.15 & 0.22 & 0.30 & 0.26 & 0.33 & 0.00 \\
\midrule[0.32pt]
\multirow{6}{*}{Skywork} & 0 & 36.36 & 47.24 & 57.90 & 66.56 & 72.30 & 75.63 & 77.46 & 78.85 & 80.61 & 83.13 & 86.67 \\
 & 1 & 36.28 & 47.10 & 57.47 & 65.71 & 71.33 & 74.85 & 76.94 & 78.79 & 81.03 & 84.17 & 86.67 \\
 & 2 & 36.35 & 47.31 & 58.11 & 66.82 & 72.63 & 76.09 & 78.25 & 80.17 & 82.32 & 84.56 & 86.67 \\
 & 3 & 36.54 & 47.46 & 57.95 & 66.20 & 71.70 & 75.16 & 77.11 & 78.48 & 80.05 & 82.09 & 83.33 \\
\cmidrule(lr){2-13}
 & \textbf{Mean} & 36.38 & 47.28 & 57.86 & 66.32 & 71.99 & 75.43 & 77.44 & 79.07 & 81.01 & 83.48 & 85.83 \\
 & SD & 0.11 & 0.15 & 0.27 & 0.48 & 0.58 & 0.55 & 0.58 & 0.75 & 0.96 & 1.11 & 1.67 \\
\midrule[0.32pt]
\multirow{6}{*}{JustRL} & 0 & 49.45 & 59.49 & 67.42 & 73.29 & 77.16 & 79.32 & 80.13 & 80.42 & 80.83 & 81.67 & 83.33 \\
 & 1 & 49.36 & 59.36 & 67.36 & 73.27 & 77.14 & 79.42 & 80.64 & 81.51 & 82.76 & 84.58 & 86.67 \\
 & 2 & 49.25 & 59.38 & 67.41 & 73.41 & 77.42 & 79.68 & 80.72 & 81.52 & 82.76 & 84.58 & 86.67 \\
 & 3 & 49.45 & 59.37 & 67.27 & 73.21 & 77.17 & 79.36 & 80.30 & 80.78 & 81.46 & 82.50 & 83.33 \\
\cmidrule(lr){2-13}
 & \textbf{Mean} & 49.38 & 59.40 & 67.37 & 73.30 & 77.22 & 79.45 & 80.45 & 81.06 & 81.95 & 83.33 & 85.00 \\
 & SD & 0.09 & 0.06 & 0.07 & 0.08 & 0.13 & 0.16 & 0.28 & 0.55 & 0.97 & 1.48 & 1.92 \\
\bottomrule[1.01pt]\end{tabular}
\end{table}
\begin{table}[!htbp]
\centering
\renewcommand{\arraystretch}{1}
\setlength{\aboverulesep}{1pt}
\setlength{\belowrulesep}{1pt}

\small
\setlength{\tabcolsep}{4pt}

\caption{TAS@$K$ agreement across the repeated AIME2024 evaluations. Overall pools the 44 comparisons from four evaluation configurations and 11 budgets.}
\label{tab:repeated-selection}
\vspace{-8.5pt}
\begin{tabular}{@{}lrc@{}}\toprule[1.01pt]
\textbf{Evaluation seed} & \textbf{Hits} & \textbf{TAS@$K$ (\%) [95\% CI]} \\ \midrule[0.32pt]
0 & 9/11 & 81.8 [36.4, 90.9] \\
1 & 10/11 & 90.9 [54.5, 90.9] \\
2 & 10/11 & 90.9 [45.5, 90.9] \\
3 & 10/11 & 90.9 [45.5, 100.0] \\
Overall & 39/44 & 88.6 [47.7, 93.2] \\
\bottomrule[1.01pt]\end{tabular}
\end{table}
\begin{revisionblock}
\subsection{Uncertainty-Based Student Orderings}
\label{app:unresolved-orderings}
We classify a student comparison as $A$ preferred if the 95\% CI of $S_A(K)-S_B(K)$ is above zero, $B$ preferred if it is below zero, and unresolved if it contains zero. Unresolved means that the evaluated problems do not clearly favor either student. It does not establish equal performance. Of the 177 comparisons, 102 have a resolved ordering and 75 are unresolved. We bootstrap problems jointly for the student pass@$K$ difference and each method's selection score, using 10,000 paired replicates, then apply the same sign rule to compare their labels. Qwen-Math teacher-score intervals use a refreshed Qwen3-8B response pool under the original evaluation settings. The other teacher and student pools remain fixed. Random assigns equal probability to the three labels.

Table~\ref{tab:unresolved-selection} shows that TAS@$K$ identifies the most unresolved student orderings and has the highest overall three-label agreement among the compared methods. Teacher pass@$K$ identifies the most resolved A/B orderings. This complements the binary point-estimate comparison.
\end{revisionblock}
\begin{table}[!htbp]
\centering
\renewcommand{\arraystretch}{1}
\setlength{\aboverulesep}{1pt}\setlength{\belowrulesep}{1pt}

\small\setlength{\tabcolsep}{5pt}
\caption{Three-label agreement with uncertainty-based student orderings. Columns separate match counts from agreement rates within each group. Best results in each group are bold.}
\label{tab:unresolved-selection}
\vspace{-8.5pt}
\begin{tabular}{@{}lrrrrrr@{}}\toprule[1.01pt]
& \multicolumn{2}{c}{\shortstack{Resolved A/B\\($n=102$)}} & \multicolumn{2}{c}{\shortstack{Unresolved\\($n=75$)}} & \multicolumn{2}{c}{\shortstack{Overall\\($n=177$)}} \\
\cmidrule(lr){2-3}\cmidrule(lr){4-5}\cmidrule(lr){6-7}
\textbf{Method} & \textbf{Matches} & \textbf{Rate (\%)} & \textbf{Matches} & \textbf{Rate (\%)} & \textbf{Matches} & \textbf{Rate (\%)} \\ \midrule[0.32pt]
Random & 34 & 33.3 & 25 & 33.3 & 59 & 33.3 \\
Teacher pass@$1$ & 55 & 53.9 & 22 & 29.3 & 77 & 43.5 \\
Teacher pass@$K$ & \textbf{61} & \textbf{59.8} & 36 & 48.0 & 97 & 54.8 \\
UTAS@$K$ & 52 & 51.0 & 44 & 58.7 & 96 & 54.2 \\
TAS@$K$ & 42 & 41.2 & \textbf{57} & \textbf{76.0} & \textbf{99} & \textbf{55.9} \\
\bottomrule[1.01pt]\end{tabular}
\end{table}

\begin{revisionblock}
\subsection{Teacher Performance versus Preference on Student Responses}
\label{app:teacher-signal-disagreement}
Teacher pass@$K$ scores a teacher on its own responses, whereas TAS@$K$, like OPD itself, scores it on the responses of the pre-OPD base model. We quantify this preference by the margin: a teacher's mean token-level log-likelihood on correct minus incorrect pre-OPD responses, averaged over problems with both correct and incorrect responses and then over benchmarks with equal weight.

\textbf{TAS@$K$ is most informative when a teacher's own performance and its preference on the student's responses disagree, which teacher pass@$K$ cannot detect.} DeepSeek-Math is the only pair in which the two signals favor different teachers: Skywork has higher own pass@$1$, whereas JustRL has the larger margin and yields the better OPD-trained model (Tables~\ref{tab:teacher-signal-agreement} and~\ref{tab:teacher-margin}). Teacher pass@$K$ therefore selects Skywork and agrees with the student ordering in 4.7\% of comparisons, compared with 90.6\% for TAS@$K$. In the other pairs, the two signals agree or are both weak.
\end{revisionblock}
\par\medskip\noindent\begin{minipage}{\textwidth}
\centering
\renewcommand{\arraystretch}{1}
\setlength{\aboverulesep}{1pt}\setlength{\belowrulesep}{1pt}

\small\setlength{\tabcolsep}{5pt}
\captionof{table}{Teacher favored by each signal and teacher of the better OPD-trained model. Mixed: no consistent winner across benchmarks. Bold marks the pair in which the signals disagree.}
\label{tab:teacher-signal-agreement}
\vspace{-8.5pt}
\begin{tabular}{@{}llll@{}}\toprule[1.01pt]
\textbf{Teacher pair} & \textbf{Higher own pass@$1$} & \textbf{Larger margin} & \textbf{Teacher of the better student} \\ \midrule[0.32pt]
Qwen-Math & RL-Math & RL-Math & RL-Math \\
\textbf{DeepSeek-Math} & \textbf{Skywork} & \textbf{JustRL} & \textbf{JustRL} \\
Qwen-Code & RL-Code & RL-Code & RL-Code \\
Qwen-Fact & SESA (slightly) & Mixed & Mixed \\
\bottomrule[1.01pt]\end{tabular}
\end{minipage}\par\medskip
\par\medskip\noindent\begin{minipage}{\textwidth}
\centering
\renewcommand{\arraystretch}{1}
\setlength{\aboverulesep}{1pt}\setlength{\belowrulesep}{1pt}

\small\setlength{\tabcolsep}{5pt}
\captionof{table}{Own pass@$1$ and margin per teacher. $\Delta$Margin: second teacher minus first, with 95\% CIs from 10,000 paired problem-bootstrap samples. Larger values within each pair are bold.}
\label{tab:teacher-margin}
\vspace{-8.5pt}
\begin{tabular}{@{}llrrr@{}}\toprule[1.01pt]
\textbf{Teacher pair} & \textbf{Teacher} & \textbf{Own pass@$1$ (\%)} & \textbf{Margin ($\times 10^{-2}$)} & \textbf{$\Delta$Margin [95\% CI]} \\ \midrule[0.32pt]
Qwen-Math & Qwen3-8B & 33.4 & 1.57 & \\
 & RL-Math & \textbf{67.9} & \textbf{3.03} & $+1.46$ [$+0.79$, $+2.15$] \\ \midrule[0.32pt]
DeepSeek-Math & Skywork & \textbf{73.8} & 1.18 & \\
 & JustRL & 61.8 & \textbf{2.01} & $+0.83$ [$+0.50$, $+1.15$] \\ \midrule[0.32pt]
Qwen-Code & Qwen3-8B & 53.8 & 22.48 & \\
 & RL-Code & \textbf{57.6} & \textbf{23.79} & $+1.31$ [$-1.17$, $+3.85$] \\ \midrule[0.32pt]
Qwen-Fact & Binary-RAR & 17.9 & 18.58 & \\
 & SESA & \textbf{19.5} & \textbf{19.45} & $+0.88$ [$-0.27$, $+2.03$] \\
\bottomrule[1.01pt]\end{tabular}
\end{minipage}\par\medskip

\begin{revisionblock}
\section{Response Diversity Before and After OPD}
\label{app:self-bleu}
We compare lexical diversity before and after OPD across three domains, six benchmarks, and 12 teacher-benchmark combinations using Self-BLEU-4. For each problem and model, we draw 16 responses and compute each response's BLEU-4 score against the other 15 responses as references. BLEU-4 measures overlap in sequences of one to four words. We lowercase and tokenize responses with a Unicode regular expression and apply Chen-Cherry method-1 smoothing. We average over responses, three paired draws (seeds 0, 1, and 2), and then problems. Changes are OPD minus the pre-OPD base model. Lower Self-BLEU indicates higher lexical diversity: negative $\Delta$Self-BLEU means diversity increased, while positive values mean it decreased. We estimate the 95\% confidence intervals by resampling paired problems 2,000 times for each teacher-benchmark combination.

{All three Math cases with small-budget gains and large-budget losses show reduced lexical diversity, whereas all eight Code and Fact cases with large-budget gains show increased mean lexical diversity. With JustRL on AMC2023, lexical diversity decreases while pass@$1024$ remains unchanged relative to the pre-OPD base model. Self-BLEU characterizes lexical concentration, and our pass@$K$ analysis connects changes in per-problem correctness to their budget-dependent aggregate effect. Together, they provide complementary views of the observed scaling patterns across sampling budgets.}
\end{revisionblock}
\begin{table}[!htbp]
\centering
\renewcommand{\arraystretch}{1}
\setlength{\aboverulesep}{1pt}
\setlength{\belowrulesep}{1pt}

\footnotesize
\setlength{\tabcolsep}{3pt}

\caption{Changes in pass@$K$ (pp) and Self-BLEU after OPD, with confidence intervals for the diversity changes. Math and Fact use $K=1024$, while Code uses $K=512$.}
\label{tab:self-bleu}
\vspace{-8.5pt}
\begin{tabular}{@{}lllrrr@{}}\toprule[1.01pt]
\textbf{Domain} & \textbf{Teacher} & \textbf{Benchmark} & \textbf{$\Delta$pass@$1$} & \textbf{$\Delta$pass@$K$} & \textbf{$\Delta$Self-BLEU [95\% CI]} \\ \midrule[0.32pt]
Math & Skywork & AMC2023 & +3.73 & -2.50 & +0.0494 [+0.0416, +0.0570] \\
Math & Skywork & AIME2025 & +5.06 & -6.67 & +0.0650 [+0.0517, +0.0763] \\
Math & JustRL & AMC2023 & +15.36 & +0.00 & +0.0509 [+0.0401, +0.0625] \\
Math & JustRL & AIME2025 & +11.30 & -3.33 & +0.0777 [+0.0653, +0.0895] \\
Code & Qwen3-8B & HumanEval+ & +2.70 & +7.93 & -0.0056 [-0.0101, -0.0012] \\
Code & Qwen3-8B & MBPP+ & +1.04 & +6.61 & -0.0357 [-0.0404, -0.0313] \\
Code & RL-Code & HumanEval+ & +5.80 & +18.29 & -0.1090 [-0.1170, -0.1009] \\
Code & RL-Code & MBPP+ & +2.26 & +18.78 & -0.1035 [-0.1091, -0.0977] \\
Fact & Binary-RAR & HotpotQA & -0.34 & +7.00 & -0.1093 [-0.1215, -0.0972] \\
Fact & Binary-RAR & TriviaQA & +1.22 & +7.50 & -0.1215 [-0.1334, -0.1096] \\
Fact & SESA & HotpotQA & -0.003 & +8.00 & -0.0905 [-0.1052, -0.0762] \\
Fact & SESA & TriviaQA & +1.27 & +5.50 & -0.0757 [-0.0883, -0.0638] \\
\bottomrule[1.01pt]\end{tabular}
\end{table}

\end{document}

%% file: math_commands.tex
\usepackage{amsmath,amsfonts,bm}

\def\eqref#1{equation~\ref{#1}}

\def\1{\bm{1}}

\DeclareMathAlphabet{\mathsfit}{\encodingdefault}{\sfdefault}{m}{sl}
\SetMathAlphabet{\mathsfit}{bold}{\encodingdefault}{\sfdefault}{bx}{n}

